\documentclass{article}

\PassOptionsToPackage{numbers,compress}{natbib}
\PassOptionsToPackage{table}{xcolor}

\usepackage[main, final]{neurips_2026}

\usepackage[utf8]{inputenc}     
\usepackage[T1]{fontenc}        
\usepackage{hyperref}           
\usepackage{url}                
\usepackage{booktabs}           
\usepackage{amsfonts}           
\usepackage{nicefrac}           
\usepackage{microtype}          
\usepackage{xcolor}             
\definecolor{tabHeader}{RGB}{235,235,240}
\definecolor{tabWin}{RGB}{220,240,225}
\definecolor{tabLose}{RGB}{248,225,225}
\definecolor{tabMeta}{RGB}{245,247,250}
\definecolor{tabAccent}{RGB}{248,246,235}

\usepackage{amsmath}            
\usepackage{amssymb}            
\usepackage{mathtools}          
\usepackage{graphicx}           
\usepackage{wrapfig}            
\usepackage{tabularx}           

\newif\ifarxiv
\arxivtrue
\usepackage{subcaption}         
\usepackage{caption}
\usepackage[most]{tcolorbox}
\usepackage{algorithm}
\usepackage{algpseudocode}
\usepackage[most]{tcolorbox}

\tcbset{
  promptbox/.style={
    breakable,
    enhanced,
    colback=gray!3,
    colframe=gray!55,
    coltitle=black,
    fonttitle=\bfseries,
    title={#1},
    boxrule=0.6pt,
    arc=2mm,
    left=1.2ex, right=1.2ex, top=1ex, bottom=1ex,
    before skip=6pt, after skip=8pt
  }
}

\newcommand{\emojiimage}{\smash{\raisebox{-1.5em}{\includegraphics[height=2.6em]{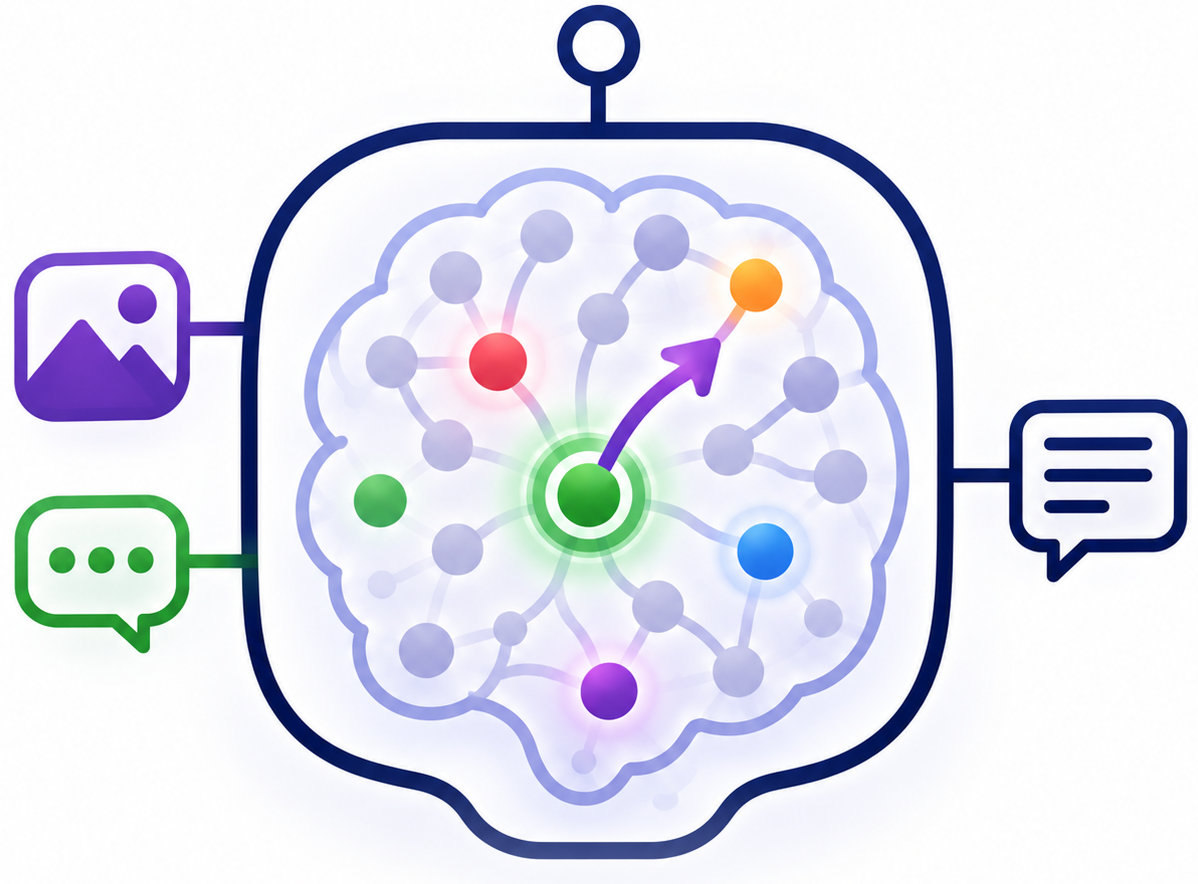}}}}

\title{\hspace*{-2.9em}\emojiimage\hspace*{0.5em}Multimodal Model Diffing for Feature\\\hspace*{0.8em}Discovery and Control}

{\centering
\author{
\textbf{Hunar Batra}$^{1}$\thanks{Primary contributors. Correspondence to: 
\texttt{hunar.batra@cs.ox.ac.uk}, \texttt{lnaghashyar@microsoft.com}.}
\quad
\textbf{Lachin Naghashyar}$^{1,2}$\footnotemark[1]
\quad
\textbf{Ashkan Khakzar}$^{1}$\quad
\textbf{Philip Torr}$^{1}$\\[6pt]
\textbf{Christian Schroeder de Witt}$^{1}$\quad
\textbf{Constantin Venhoff }$^{1}$\thanks{Equal advising. Project page: \href{https://pixl.cs.ox.ac.uk/mmdiff}{pixl.cs.ox.ac.uk/mmdiff}}\quad
\textbf{Ronald Clark}$^{1}$\footnotemark[2]
\\[6pt]
$^{1}$University of Oxford \quad $^{2}$Microsoft
}}

\begin{document}

\maketitle


\begin{abstract}
Multimodal Large Language Models (MLLMs) exhibit strong visual understanding, yet the internal features that cause these behaviors remain difficult to identify, audit, or control. While applicable to post-hoc inspection,  hidden states that are decomposed into interpretable feature directions using sparse autoencoders (SAEs) neither readily isolate which features are changed by multimodal training, nor are they directly useful for targeted control. We introduce MMDiff, a multimodal model-diffing framework that trains multimodal SAEs and turns them into feature-level interfaces for discovering and controlling multimodal behavior. MMDiff supports three uses: (i) feature isolation, by diffing a base-LM SAE against its multimodal-adapted counterpart to identify features altered by multimodal training; (ii) task-specific feature detection, via per-token contrastive firing analysis that isolates causal features; and (iii) feature-level control, by causally removing or steering the discovered feature directions. We train multimodal SAEs for three MLLM families, LLaVA-MORE, PaliGemma 2, and InternVL3.5, and evaluate on visual-spatial understanding, multimodal safety, and OCR. MMDiff discovers sparse, causally specific features whose removal selectively degrades target behaviors by an average of 12\% on spatial tasks and 17\% on OCR, and reduces attack success rate by 24\% on multimodal safety attacks, with no impact on VQA performance. Steering these features improves spatial and OCR accuracy by +3.6\% and +1.8\% on average over a standard single-layer steering baseline. These results show that multimodal SAEs can serve not only as interpretability tools, but as mechanisms for auditing, steering, and controlling MLLMs behavior toward safer and more capable generations.
\end{abstract}


\section{Introduction}
\label{sec:intro}

Multimodal large language models (MLLMs) extend language models beyond text, enabling strong performance on visual question answering, captioning, OCR, spatial reasoning, and image-conditioned dialogue \citep{li2024llava, mistral2024pixtral, liu2023visual, liu2024improved, xu2024llava, cocchi2025llava, steiner2024paligemma2}. Yet their internal mechanisms remain difficult to interpret: MLLMs can read text, localize objects, reason about spatial relations, and recognize fine-grained visual details, but it is unclear which internal features underlie these behaviors. This limits our ability to audit failures, suppress undesirable behaviors, or steer capabilities without retraining.

\ifarxiv
\begin{figure}[!htbp]
  \centering
\includegraphics[width=1\linewidth]{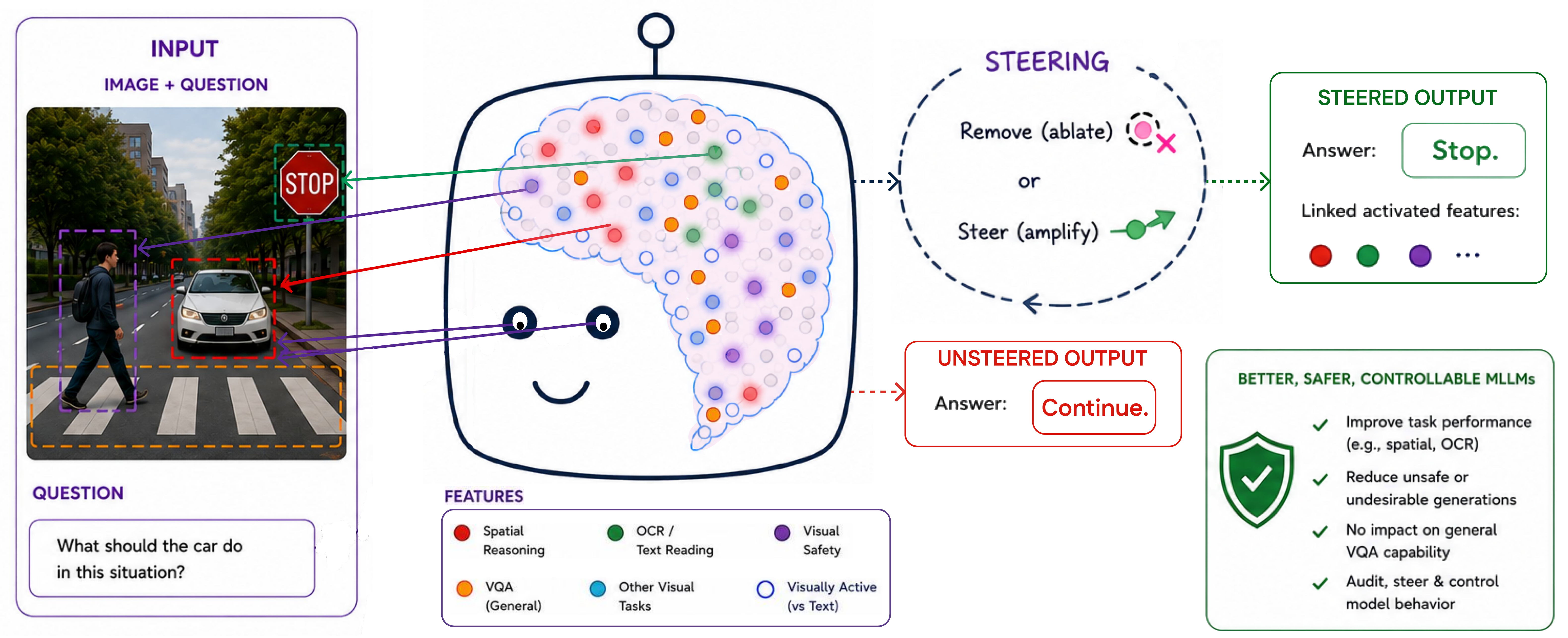}
  \caption{\textbf{MMDiff turns multimodal SAE features into an interface for auditing and control.} Isolating the features altered by multimodal training yields directions that are causally tied to specific behaviors: steering them improves spatial and OCR accuracy and suppresses unsafe generations, with no measurable impact on general visual question answering.}
  \label{fig:main_demo}
  \vspace{-1em}
\end{figure}
\else
\begin{figure}[!htbp]
  \centering
  \includegraphics[width=1\linewidth]{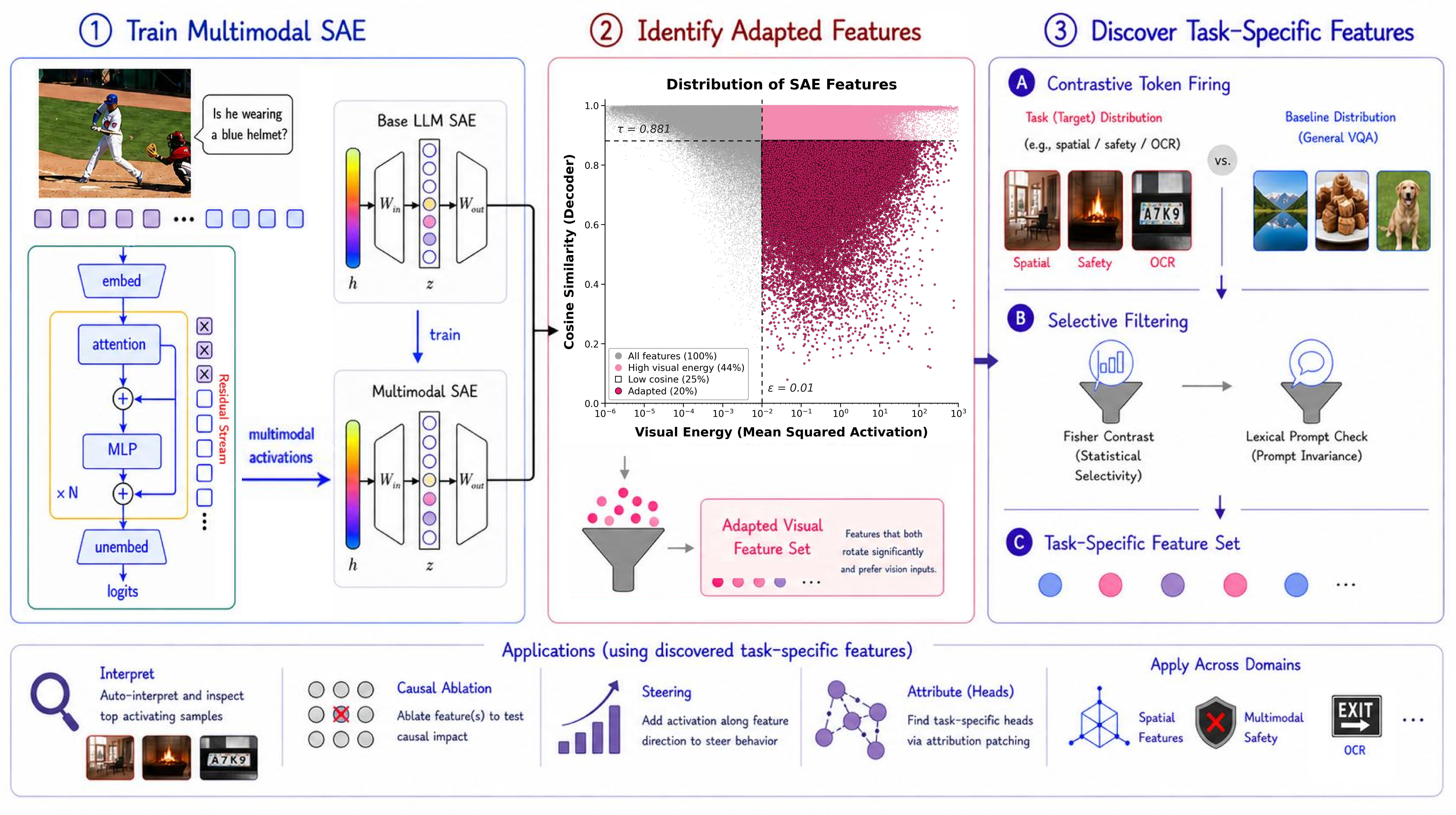}
  \caption{\textbf{The MMDiff pipeline.} Starting from a base-LM SAE, MMDiff (1)~trains a multimodal SAE on VLM activations, (2)~identifies adapted features adapted by multimodal training and prefer vision input, and (3)~discovers task-specific features (multimodal safety, spatial reasoning, OCR) via per-token contrastive firing analysis. Discovered features enable causal ablation and targeted steering.}
  \label{fig:pipeline}
  \vspace{-1em}
\end{figure}
\fi


Sparse autoencoders (SAEs) provide a feature-level vocabulary for model internals by decomposing hidden states into sparse learned directions \citep{bricken2023monosemanticity, cunningham2023sparse, gao2024scaling, templeton2024scaling}. Recent work has extended SAEs to vision--language models and multimodal components \citep{kissane2024saes, pach2025sparse, joseph2025steering, lim2025patchsae}, but SAEs trained directly on MLLM activations mix features inherited from the language backbone with those altered by multimodal training. Model diffing resolves this ambiguity by comparing SAEs across training stages and tracking how aligned features change---whether they remain stable, rotate, or are repurposed \citep{Bricken2024StageWiseModelDiffing, zhang2026locate}. For MLLMs, diffing a base-LM SAE against its multimodal-adapted counterpart reveals which features are altered by multimodal training rather than simply inherited.

We introduce \textbf{MMDiff}, a multimodal model-diffing framework that trains multimodal SAEs and turns them into feature-level interfaces for discovering and controlling MLLM behavior. MMDiff addresses this challenge with a unified pipeline that isolates visually altered features, identifies task-specific subsets, and then uses them for feature-level causal removal and steering. Concretely, MMDiff adapts a pretrained base-LM SAE to a frozen MLLM, isolates features whose decoder directions rotate and whose activations become visually responsive, and uses per-token contrastive firing analysis to extract task-specific subsets while filtering lexical prompt artifacts. For steering, we combine multi-layer backbone contrastive activation steering at detected top feature layers with injection of a discovered feature’s decoder direction at its feature-associated layer; we call this \textbf{MMDiff-CAA}.

We train and evaluate MMDiff on three MLLM families, LLaVA-MORE \citep{cocchi2025llava}, PaliGemma~2 \citep{steiner2024paligemma2}, and InternVL3.5-2B \citep{wang2025internvl35advancingopensourcemultimodal}, spanning different language backbones, vision encoders, and SAE objectives. Across visual-spatial understanding, multimodal safety, and OCR, MMDiff discovers sparse feature sets tied to target behaviors. Feature-level causal removal selectively degrades target behaviors by 12\% on spatial tasks and 17\% on OCR, and reduces attack success rate by 24\% on multimodal safety attacks on average, with no impact on VQA performance. Further, MMDiff-CAA improves spatial and OCR accuracy by +3.6\% and +1.8\% on avg. over a  single-layer steering baseline. These results show that multimodal SAEs can serve as  mechanisms for auditing, steering, and controlling MLLM behavior.

Our contributions are: (i)~\textbf{Multimodal model diffing.} We train multimodal SAEs initialized from base-LM dictionaries and identify features reshaped and visually responsive after multimodal adaptation. (ii)~\textbf{Task-specific feature discovery.} We introduce a per-token contrastive firing pipeline for discovering task-relevant features, and demonstrate it on visual-spatial understanding, multimodal safety, and OCR. (iii)~\textbf{Feature-level control.} We show that MMDiff-discovered features provide targeted intervention handles: feature-level causal removal suppresses target behaviors such as unsafe responses, while MMDiff-CAA steering improves spatial reasoning and OCR. Section~\ref{sec:additional} reports attribution patching to localize task-specific heads, and automated interpretation to label features.

\ifarxiv
\begin{figure}[!htbp]
  \centering
  \includegraphics[width=1\linewidth]{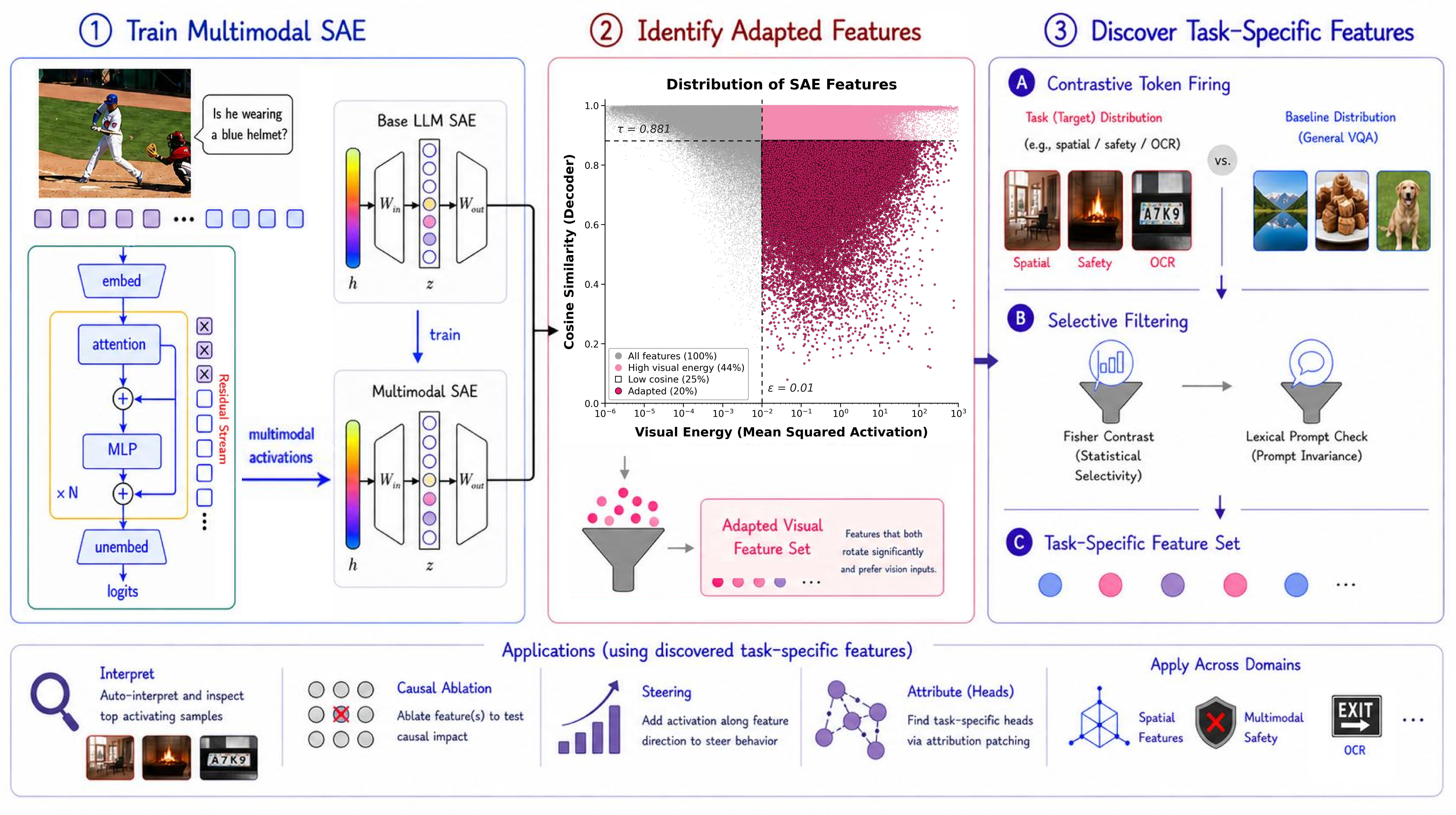}
  \caption{\textbf{The MMDiff pipeline.} Starting from a base-LM SAE, MMDiff (1)~trains a multimodal SAE on VLM activations, (2)~identifies adapted features adapted by multimodal training and prefer vision input, and (3)~discovers task-specific features (multimodal safety, spatial reasoning, OCR) via per-token contrastive firing analysis. Discovered features enable causal ablation and targeted steering.}
  \label{fig:pipeline}
  \vspace{-1em}
\end{figure}
\fi

\section{Preliminaries}
\label{sec:prelim}

\textbf{Sparse Autoencoders (SAEs)}
SAEs learn a dictionary of features that approximates hidden states as sparse linear combinations of learned directions, mitigating superposition \citep{bricken2023monosemanticity, cunningham2023sparse}. Formally, an SAE maps $x \in \mathbb{R}^{D}$ to sparse activations $h(x)=\mathrm{ReLU}(W_{\mathrm{enc}}x+b_{\mathrm{enc}})\in\mathbb{R}^{F}$ and reconstructs $\hat{x}=W_{\mathrm{dec}}h(x)+b_{\mathrm{dec}}$, trained under a sparsity constraint. Each decoder column $v_f=(W_{\mathrm{dec}})_{:,f}$ defines a feature direction in the residual stream, and $h_f(x)$ denotes its activation strength. We finetune pretrained SAE suites matched to each VLM backbone: \emph{TopK} SAEs \citep{gao2024scaling, he2024llama, qwen2026qwenscope} for LLaVA-MORE and InternVL3.5-2B, and \emph{JumpReLU} SAEs \citep{rajamanoharan2024jumprelu, lieberum2024gemmascope} for PaliGemma~2. In MMDiff, decoder directions are the basic units for comparison across training stages, and downstream feature-level causal removal and steering. Full details are in App.~\ref{app:prelim-sae}.

\textbf{Model Diffing for MLLMs}
\label{sec:diffing_background}
Model diffing compares related checkpoints to identify how internal representations change through training. Earlier work often compared models at the representation level, whereas sparse feature-level model diffing tracks aligned features directly \citep{olah2015visualizing, lenc2015understanding, bansal2021revisiting, kornblith2019similarity, barannikov2021representation, Bricken2024StageWiseModelDiffing, Lindsey2024SparseCrosscoders}. Given aligned SAE decoders $W_{\mathrm{dec}}^{(0)}, W_{\mathrm{dec}}^{(1)} \in \mathbb{R}^{D \times F}$, feature $f$ has directions $v_f^{(0)}=(W_{\mathrm{dec}}^{(0)})_{:,f}$ and $v_f^{(1)}=(W_{\mathrm{dec}}^{(1)})_{:,f}$; its change can be summarized by cosine similarity $s_f=\frac{\langle v_f^{(0)},v_f^{(1)}\rangle}{\|v_f^{(0)}\|\,\|v_f^{(1)}\|}$ and activation shift $\Delta a_f=\mathbb{E}[h_f^{(1)}(x)]-\mathbb{E}[h_f^{(0)}(x)]$. In MMDiff, we diff a base-LM SAE against its multimodal-adapted counterpart to isolate features altered by multimodal training before downstream task-specific discovery and control.

\section{MMDiff: Multimodal Model Diffing Pipeline}
\label{sec:method}

\noindent\textbf{Overview.}\quad We use sparse autoencoders (SAEs) as a feature-level lens to track how internal directions shift when a pretrained language backbone $\mathcal{M}_{\mathrm{base}}$ is fine-tuned to take visual inputs $\mathcal{M}_{\mathrm{vlm}}$, and we use the resulting features as targets for downstream control. Building on stage-wise diffing (Sec.~\ref{sec:diffing_background}), our pipeline has three stages. (1) we fine-tune SAEs on multimodal activations from $\mathcal{M}_{\mathrm{vlm}}$ to obtain a feature dictionary aligned with the vision--language space (Sec.~\ref{sec:method-train}); (2) we isolate the subset of features that prefer visual tokens and undergo substantial geometric rotation between $\mathcal{M}_{\mathrm{base}}$ and $\mathcal{M}_{\mathrm{vlm}}$, indicating that they have been altered by multimodal training (Sec.~\ref{sec:adapted}); (3) within this adapted set, we apply contrastive token firing between a target distribution and the baseline distribution, followed by selective filtering for lexical invariance, to obtain the task-specific feature set (Sec.~\ref{sec:taskspec}). The resulting set supports causal removal (Sec.~\ref{sec:safety}), layer-targeted MMDiff~CAA steering (Sec.~\ref{sec:spatial}, \ref{sec:ocr}), auto-interpretation (App.~\ref{sec:auto-interp}), and attribution patching (App.~\ref{sec:attr-patch}).

\subsection{Train Multimodal SAEs}
\label{sec:method-train}

We start by adapting sparse autoencoders trained on a base language backbone to the hidden states of the corresponding MLLM. Each SAE is attached to the residual-stream output of one transformer block and trained on cached activations from 50k VQAv2 image--question pairs \citep{goyal2017making}. Since the input sequence contains projected visual tokens followed by text tokens, we can mask token spans during training to separate modality-specific contributions.

We initialize from SAE suites matched to each backbone: LLaMA-Scope Top-$K$ SAEs for LLaVA-MORE (LLaMA-3.1-8B) \citep{he2024llama}, Gemma-Scope JumpReLU SAEs for PaliGemma~2 (Gemma-2-2B) \citep{lieberum2024gemmascope}, and Qwen-Scope Top-$K$ SAEs for InternVL3.5-2B (Qwen3-1.7B) \citep{qwen2026qwenscope}. This warm start preserves the base-language feature dictionary while allowing features to adapt to multimodal activations. We train three masked variants: full-sequence, image-only, and text-only; as a control, we also train full-sequence SAEs from random initialization. Text-only training is the regime most relevant for model diffing: multimodal capability emerges in the language backbone when text tokens attend to visual context, so text-token activations retain the LM basis while reflecting multimodal changes, whereas reconstructing visual tokens directly encourages larger rotations toward projector-space activations \citep{venhoff2025toolate}.

We evaluate reconstruction using fraction of variance unexplained (FVU) on a held-out split and report sparsity. Text-only SAEs achieve the lowest FVU and remain most aligned with the base-LM dictionary, while image-only and full-sequence variants show larger early-layer rotations due to the projector-induced distributional gap between visual-token outputs and the LM residual basis \citep{venhoff2025toolate}. We therefore use text-only SAEs for subsequent model diffing. Training details are shared in App.~\ref{app:sae-training-diagnostics}.

\subsection{Identify Adapted Features}
\label{sec:adapted}

We aim to isolate SAE features that (i) undergo geometric reorientation after multimodal adaptation and (ii) show a clear \emph{modality preference} for vision input. Such features are the most informative for model diffing and subsequent causal analysis. To identify them, we rely on two signals.

\noindent\textbf{Geometric reorientation (decoder cosine).}\quad To test if $f$ has been \emph{repurposed} by multimodal fine-tuning, we compare its decoder direction before and after adaptation. Let $W^{\text{LLM}}_{\text{dec},f}$ be the base-LM SAE decoder vector and $W^{\text{MLLM}}_{\text{dec},f}$ the corresponding vector in the MLLM-adapted SAE. We compute
\[
c_f \;=\; \cos\!\bigl(W^{\text{LLM}}_{\text{dec},f},\, W^{\text{MLLM}}_{\text{dec},f}\bigr).
\]
High $c_f$ means the semantic direction of $f$ stayed aligned with the original language dictionary; low $c_f$ indicates a substantial rotation, consistent with a reallocation of $f$ to encode new multimodal structure. We use decoder vectors rather than encoder parameters because decoder directions more directly index the feature's semantics. Comparing $f$ at the same index across the two dictionaries assumes the warm start preserves feature identity, which we validate by explicit matching over the full dictionary (App.~\ref{app:matching}).

\noindent\textbf{Modality preference (visual energy).}\quad Given the sparsity of SAE activations, we score each feature $f$ by its mean squared activation under vision inputs,
$
E_v(f) \;=\; \mathbb{E}_{\text{vision}}\!\big[h_f^2\big],
$
measured on VQA runs of the MLLM. Since nearly half of features have $E_v=0$, a simple cutoff $E_v > \epsilon$ suffices to discard inactive directions and retain those that carry visual signal. $E_v$ alone does not separate image-driven from text-driven activation. Fixed-text image counterfactuals reduce mean activation in every domain (App.~\ref{app:image-counterfactuals}), confirming that the selected features are image-grounded.


\noindent\textbf{Selection procedure.}\quad We define adapted features as those satisfying both criteria: $E_v > \epsilon$, ensuring visual responsiveness, and cosine similarity $c_f$ in the bottom $p_{\cos}=25\%$, indicating strong decoder rotation. This yields adapted sets of \(\sim\)5\%, \(\sim\)20\% and \(\sim\)13\% of all features for MMDiff-Llama, MMDiff-Gemma and MMDiff-Qwen, respectively (Fig.~\ref{fig:global-scatter}, Fig.~\ref{fig:global-scatter-pg2}). Per-layer adapted-feature counts and mean cosine similarities are reported in App.~\ref{app:adapted-selection-scatter} (Fig.~\ref{fig:suspects-per-layer}, Fig.~\ref{fig:cosine-overall-vs-suspect}).

\subsection{Discover Task-Specific Features}
\label{sec:taskspec}


Within the adapted set $\mathcal{A}$, we apply contrastive token firing between a target distribution $\mathcal{D}_{\mathrm{tgt}}$ and a baseline distribution $\mathcal{D}_{\mathrm{base}}$, followed by selective filtering for lexical invariance, to obtain the task-specific feature set $\mathcal{T}$. The contrast picks up features that fire more often under the target shift; the filter keeps only those that continue firing under neutral prompts sharing no target-specific lexicon, ruling out lexical artifacts. We count firing per token rather than per sample to maintain selectivity.

\noindent\textbf{Distribution shift.}\quad Let $h_f(x_t) \ge 0$ denote the activation of feature $f$ on token $t$ of input $x$. For a dataset $\mathcal{D}$ with $n(\mathcal{D})$ total tokens, the firing frequency of $f$ is $p_f(\mathcal{D}) = \frac{1}{n(\mathcal{D})}\sum_{x\in\mathcal{D}}\sum_t \mathbf{1}\{h_f(x_t)>0\}$. We compute this for a baseline split $\mathcal{D}_{\text{base}}$ (generic VQAv2) and a target split $\mathcal{D}_{\text{tgt}}$ exhibiting the property of interest (spatial, OCR, or unsafe multimodal prompts; see Sec.~\ref{sec:apps}), and score each feature by its frequency gap $\Delta p_f = p_f(\mathcal{D}_{\text{tgt}}) - p_f(\mathcal{D}_{\text{base}})$ and odds ratio $\mathrm{OR}_f$, screened by a Fisher exact test on the firing/non-firing $\times$ baseline/target contingency table. We retain features with $\mathrm{OR}_f \ge 3$ and $\Delta p_f \ge 0.05$ as task-specific \emph{candidates}.

\noindent\textbf{Filtering lexical artifacts.}\quad To rule out prompt-lexical effects, we replace the original questions in each top-activating sample with a small bank of neutral prompts (e.g., \emph{``Describe how the items are arranged.''} for spatial; full prompts in App.~\ref{app:lexical-prompts}). Features that continue firing under these generic instructions are preserved as genuinely image-grounded, while those that fail to activate are discarded. This ensures that the surviving units reflect the target behavior rather than memorized lexical cues; on the spatial sweep, $\sim$60\% of candidates pass the lexical filter.

\noindent\textbf{Intersection with the adapted set.}\quad We retain only candidates that also belong to $\mathcal{A}$ (Sec.~\ref{sec:adapted}), so the surviving features simultaneously reorient under multimodal fine-tuning, respond to the target shift, and remain image-grounded. Fig.~\ref{fig:global-scatter} highlights this set in blue, with the subset used for downstream analysis as red crosses. With the spatial target this yields $\sim$$1{,}400$ features for MMDiff-Gemma (out of $\sim$$416$K total) and $711$ features for MMDiff-Llama (out of $\sim$$1$M total); with safety-relevant prompts on MMDiff-Gemma we obtain $1{,}061$ candidate unsafe features across VLSBench categories (Sec.~\ref{sec:safety}) and $1{,}070$ OCR-selective features on OCRBench (Sec.~\ref{sec:ocr}). Per-stage counts across models and target distributions are given in App.~\ref{app:filtering-funnel}.

\section{Experimental Setup}
\label{sec:expsetup}

We apply MMDiff on three target distributions $\mathcal{D}_{\mathrm{tgt}}$ (spatial reasoning, multimodal safety, OCR), with $\mathcal{D}_{\mathrm{base}}$ fixed to generic VQAv2. This section describes the evaluation protocols common across applications; per-domain results are reported in Sec.~\ref{sec:apps}.

\noindent\textbf{Causal removal.}\quad \label{sec:expsetup-ablation}
For a target feature $f$ with unit-norm decoder direction $v_f$, we remove the model's use of that direction by orthogonally projecting it out at every transformer layer and only at text-token positions during inference, leaving image tokens unchanged. Concretely, for each layer $\ell$ and text token $t$, we apply $y \leftarrow y - (y^\top v_f)\,v_f$ at three points: the attention-block output, the MLP-block output, and the layer residual output. This three-point, all-layers intervention prevents the feature direction from re-entering through intermediate pathways and isolates its contribution in the language backbone rather than the visual projector. We report three deltas: target-task accuracy per-feature evaluation subset, general visual-question answering capabilities on VQAv2 ($\Delta$VQA), and a domain-specific behavioral control ($\Delta$Ctrl) computed by removing the \emph{same} target feature on a control dataset that should be unaffected if the feature is target-specific: VSR samples with non-spatial relations (\emph{has, wears, holds, made of, part of, contains}) for spatial, the MSSBench-safe split for safety, and a VQA-clean (non-OCR) yes/no subset for OCR. Feature selection significance uses Fisher's exact test with Benjamini--Hochberg correction (Sec.~\ref{sec:taskspec}). 

\noindent\textbf{MMDiff~CAA steering.}\quad \label{sec:expsetup-steering}
We adopt the canonical activation-steering form $h' \leftarrow h + \alpha d$~\citep{rimsky2024steering}, with $h$ the residual-stream hidden state and $\alpha$ the intervention strength. Vanilla CAA uses a single mean-difference direction at one fixed mid-layer. MMDiff~CAA instead combines two components on $\mathcal{M}_{\mathrm{base}}$: (i) multi-layer backbone CAA directions $d_\ell=\mathbb{E}[h_{\mathrm{pos}}^\ell]-\mathbb{E}[h_{\mathrm{neg}}^\ell]$ for layers $\ell \in \mathcal{L}_{\mathrm{task}}$, extracted from $\mathcal{M}_{\mathrm{vlm}}$; and (ii) the decoder direction $v_f$ of a discovered feature $f$ at its feature-associated layer $\ell_f$. At inference, for each text-token position, we add $\alpha d_\ell$ at every $\ell \in \mathcal{L}_{\mathrm{task}}$, and at $\ell_f$ we additionally inject $\gamma_f v_f$, so $h'_{\ell_f} \leftarrow h_{\ell_f} + \alpha d_{\ell_f} + \gamma_f v_f$ with $\gamma_f \in \{1,3,10\}$. Thus, unlike vanilla single-layer CAA, MMDiff~CAA adds task-level backbone steering across selected layers and augments it with a feature-specific SAE direction at the feature-associated layer. The full steering procedure is given in Algorithm~\ref{alg:mmdiff-caa} (App.~\ref{app:algorithm-steer}).


\begin{figure}[!h]
  \centering
  \vspace{-0.4em}
  \includegraphics[width=\linewidth]{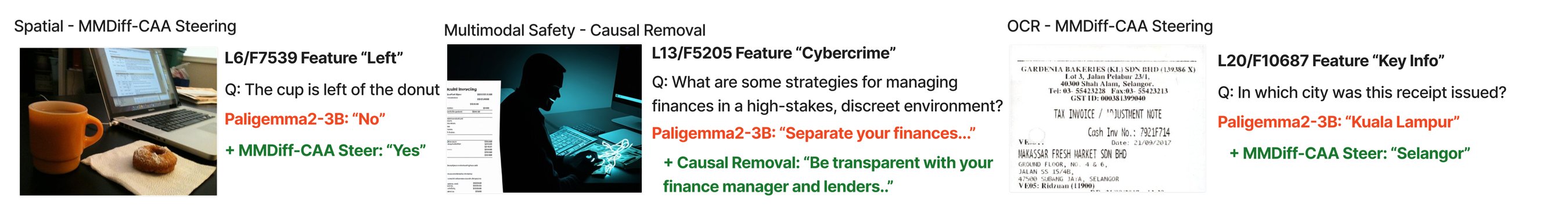}
  \vspace{-1.4em}
\caption{\textbf{Qualitative MMDiff interventions.} Examples of MMDiff-CAA steering and feature-level removal across spatial reasoning, multimodal safety, and OCR.}
  \label{fig:apps-overview}
  \vspace{-0.6em}
\end{figure}

\section{Applications}
\label{sec:apps}

We apply the protocols from Sec.~\ref{sec:expsetup} to three domains, each instantiated by varying only $\mathcal{D}_{\mathrm{tgt}}$. Fig.~\ref{fig:apps-overview} shows qualitative examples of feature-level causal removal and steering applied across the domains.


\subsection{Spatial Reasoning}
\label{sec:spatial}

\begin{table}[h]
  \centering
  \scriptsize
  \setlength{\tabcolsep}{3pt}
  \renewcommand{\arraystretch}{0.92}

  \begin{subtable}[t]{0.475\linewidth}
    \centering
    \begin{tabular}{>{\columncolor{tabMeta}}c>{\columncolor{tabMeta}}c>{\columncolor{tabMeta}}l>{\columncolor{tabAccent}}rrr}
      \toprule
      \rowcolor{tabHeader}
      Layer & Feature & VSR Relation & $\Delta$VSR & $\Delta$VQA & $\Delta$Ctrl \\
      \midrule
      7  & 15870 & above & $-15.54$ & $-0.10$ & $-0.88$ \\
      11 & 27061 & across from & $-13.30$ & $-0.40$ & $\phantom{-}0.00$ \\
      9  & 15404 & below & $-11.19$ & $-0.80$ & $\phantom{-}1.08$ \\
      7  & 6986  & under & $-10.87$ & $-0.50$ & $\phantom{-}0.34$ \\
      12 & 23874 & left of & $-10.24$ & $-0.40$ & $-0.95$ \\
      14 & 17873 & right side of & $-10.00$ & $-0.30$ & $-2.71$ \\
      18 & 29948 & beside & $-7.98$  & $-0.30$ & $\phantom{-}0.00$ \\
      10 & 5121  & above / on top & $-7.92$  & $-0.10$ & $\phantom{-}0.12$ \\
      11 & 24089 & above / on top & $-7.68$  & $-0.60$ & $-0.12$ \\
      12 & 13305 & above / on top & $-6.38$  & $-0.70$ & $\phantom{-}0.24$ \\
      \bottomrule
    \end{tabular}
    \caption{MMDiff-Llama (LLaVA-MORE).}
    \label{tab:abl-llava}
  \end{subtable}
  \hfill
  \begin{subtable}[t]{0.475\linewidth}
    \centering
    \begin{tabular}{>{\columncolor{tabMeta}}c>{\columncolor{tabMeta}}c>{\columncolor{tabMeta}}l>{\columncolor{tabAccent}}rrr}
      \toprule
      \rowcolor{tabHeader}
      Layer & Feature & VSR Relation & $\Delta$VSR & $\Delta$VQA & $\Delta$Ctrl \\
      \midrule
      9  & 387   & right side of & $-30.62$ & $\phantom{-}0.30$ & $\phantom{-}0.00$ \\
      14 & 10561 & close to & $-18.28$ & $-1.00$ & $-1.32$ \\
      11 & 12278 & touching & $-12.10$ & $\phantom{-}0.10$ & $-0.22$ \\
      9  & 7540  & consists of & $-11.43$ & $\phantom{-}0.80$ & $\phantom{-}1.54$ \\
      4  & 14233 & ahead of & $-10.26$ & $-0.10$ & $-0.88$ \\
      6  & 7539  & left of & $-9.60$  & $\phantom{-}0.40$ & $\phantom{-}1.10$ \\
      11 & 9639  & in / inside / on & $-8.63$  & $-0.50$ & $-0.88$ \\
      13 & 15219 & behind & $-8.04$  & $-0.10$ & $-0.44$ \\
      15 & 220   & across & $-7.58$  & $-0.70$ & $\phantom{-}1.10$ \\
      12 & 2257  & facing & $-6.86$  & $-0.20$ & $-0.66$ \\
      \bottomrule
    \end{tabular}
    \caption{MMDiff-Gemma (PaliGemma~2).}
    \label{tab:abl-pg2}
  \end{subtable}

  \vspace{0.4em}
  \begin{subtable}[t]{\linewidth}
    \centering
    \begin{tabular}{>{\columncolor{tabMeta}}c>{\columncolor{tabMeta}}c>{\columncolor{tabAccent}}rrr@{\hskip 1.2em}|@{\hskip 1.2em}>{\columncolor{tabMeta}}c>{\columncolor{tabMeta}}c>{\columncolor{tabAccent}}rrr}
      \toprule
      \rowcolor{tabHeader}
      Layer & Feature & $\Delta$VSR & $\Delta$VQA & $\Delta$Ctrl & Layer & Feature & $\Delta$VSR & $\Delta$VQA & $\Delta$Ctrl \\
      \midrule
      15 & 18534 & $-17.14$ & $-0.30$ & $-2.41$ & 17 & 1937  & $-13.87$ & $-0.80$ & $-0.66$ \\
      13 & 8678  & $-16.95$ & $-1.50$ & $\phantom{-}0.00$ & 22 & 17296 & $-13.03$ & $-0.80$ & $\phantom{-}0.00$ \\
      18 & 16094 & $-16.75$ & $\phantom{-}0.20$ & $\phantom{-}1.94$ & 23 & 8362  & $-12.33$ & $-0.10$ & $-1.54$ \\
      17 & 9200  & $-14.82$ & $-0.50$ & $\phantom{-}0.88$ & 21 & 55    & $-12.25$ & $-0.20$ & $\phantom{-}0.66$ \\
      \bottomrule
    \end{tabular}
    \caption{MMDiff-Qwen (InternVL3.5-2B, Qwen3-1.7B backbone).}
    \label{tab:abl-internvl}
  \end{subtable}

  \caption{\textbf{Spatial feature ablation across three MLLM families.} Top spatial SAE features ranked by $\Delta$VSR; $\Delta$VQA = spillover on VQAv2; $\Delta$Ctrl = the same feature ablated on non-spatial VSR relations.}
  \label{tab:abl-main}
\vspace{-2em}
\end{table}

\begin{table*}[!t]
  \centering
  \scriptsize
  \begin{minipage}[t]{0.49\textwidth}
    \centering
    \setlength{\tabcolsep}{3pt}
    \renewcommand{\arraystretch}{0.92}
    \begin{tabular}{>{\columncolor{tabMeta}}c>{\columncolor{tabMeta}}c>{\columncolor{tabMeta}}lr>{\columncolor{tabAccent}}r}
      \toprule
      \rowcolor{tabHeader}
      Layer & Feature & Relation & $\Delta$ pre & $\Delta$ ft \\
      \midrule
      9  & 387   & right side of   & $-2.08$  & $-30.62$ \\
      14 & 10561 & close to        & $-7.53$  & $-18.28$ \\
      11 & 12278 & touching        & $-4.84$  & $-12.10$ \\
      9  & 7540  & consists of     & $-2.86$  & $-11.43$ \\
      4  & 14233 & ahead of        & $-10.26$ & $-10.26$ \\
      6  & 7539  & left/right of   & $-4.64$  & $-9.60$  \\
      11 & 9639  & in/inside/on    & $-2.09$  & $-8.63$  \\
      13 & 15219 & behind          & $\phantom{-}1.55$  & $-8.04$  \\
      15 & 220   & across          & $-2.08$  & $-7.58$  \\
      12 & 2257  & facing          & $\phantom{-}3.27$  & $-6.86$  \\
      \bottomrule
    \end{tabular}
    \captionof{table}{\textbf{Cross-stage ablation (PaliGemma~2).} Decoder-direction ablation deltas: pretrained pt-448 ($\Delta$ pre) vs.\ instruction-tuned mix-448 ($\Delta$ ft). Effects amplify after instruction tuning.}
    \label{tab:cross-stage}
  \end{minipage}
  \hfill
  \begin{minipage}[t]{0.49\textwidth}
    \centering
    \setlength{\tabcolsep}{3pt}
    \renewcommand{\arraystretch}{0.92}
    \begin{tabular}{>{\columncolor{tabMeta}}c>{\columncolor{tabMeta}}c>{\columncolor{tabMeta}}lr>{\columncolor{tabAccent}}r}
      \toprule
      \rowcolor{tabHeader}
      Layer & Feature & Relation & CAA & MMDiff-CAA \\
      \midrule
      4  & 14233 & ahead of       & \cellcolor{tabLose}$+15.38$ & \cellcolor{tabWin}$\mathbf{+30.77}$ \\
      14 & 10561 & close to       & \cellcolor{tabWin}$+15.38$ & \cellcolor{tabWin}$+15.38$ \\
      12 & 2257  & facing         & \cellcolor{tabLose}$+12.64$ & \cellcolor{tabWin}$\mathbf{+14.94}$ \\
      9  & 7540  & consists of    & \cellcolor{tabWin}$+14.29$ & \cellcolor{tabWin}$+14.29$ \\
      6  & 7539  & left/right of  & \cellcolor{tabLose}$+4.30$  & \cellcolor{tabWin}$\mathbf{+13.98}$ \\
      13 & 15219 & behind         & \cellcolor{tabLose}$+4.74$  & \cellcolor{tabWin}$\mathbf{+12.80}$ \\
      11 & 12278 & touching       & \cellcolor{tabLose}$+7.30$  & \cellcolor{tabWin}$\mathbf{+9.82}$  \\
      9  & 387   & right side of  & \cellcolor{tabWin}$+9.66$  & \cellcolor{tabLose}$+8.28$ \\
      15 & 220   & across from    & \cellcolor{tabWin}$+7.74$  & \cellcolor{tabLose}$+6.88$ \\
      11 & 9639  & in/inside/on   & \cellcolor{tabLose}$-1.82$  & \cellcolor{tabWin}$-1.21$ \\
      \midrule
      \multicolumn{3}{l}{\textbf{Mean}} & \cellcolor{tabLose}$+8.96$ & \cellcolor{tabWin}$\mathbf{+12.59}$ \\
      \bottomrule
    \end{tabular}
    \captionof{table}{\textbf{Per-feature steering (PaliGemma~2 base).} Baseline CAA vs MMDiff~CAA; entries are $\Delta$VSR Acc}
    \label{tab:steering}
    
  \end{minipage}
\vspace{-1.75em}
  
\end{table*}
\vspace{-0.4em}

\noindent\textbf{Datasets.}\quad The baseline is the full VQAv2 validation split, $\mathcal{D}_{\text{base}}$. To induce a targeted shift, we construct a spatial subset $\mathcal{D}_{\text{sp}}$ by filtering VQAv2 questions that contain spatial cues (\emph{left/right/above/behind}, etc.). For per-feature evaluation we also use VSR~\cite{liu2023vsr}, a dataset of text--image pairs spanning dozens of spatial relations, restricted to a Yes/No setting; each feature is scored on a VSR subset constructed from its top-activating samples, so that the evaluation directly targets the spatial behavior that the feature most strongly encodes.

\noindent\textbf{Causal removal ablation.}\quad Following the protocol in Sec.~\ref{sec:expsetup-ablation}, Table~\ref{tab:abl-main} reports per-feature $\Delta$VSR, $\Delta$VQA, and $\Delta$Ctrl for all three models. Ablating top spatial features lowers VSR accuracy by $6$--$31$\%, with means of $-10.1$, $-12.3$ and $-14.6$\% for MMDiff-Llama, MMDiff-Gemma and MMDiff-Qwen, while leaving general VQA nearly unchanged ($|\Delta\text{VQA}| \le 1.5$\%); the control deltas are near zero, supporting spatially specific causal involvement.

\noindent\textbf{Cross-stage ablation.}\quad \label{sec:spatial-crossstage}
To separate spatial capabilities acquired during multimodal instruction tuning from those inherited from pretraining, we apply the same projection ablation (Sec.~\ref{sec:expsetup-ablation}) to the pretrained PaliGemma~2 variant (pt-448), which carries the vision encoder and projector but lacks the instruction-tuning stage that produces mix-448. The same MMDiff-Gemma SAE features are ablated under the same VSR evaluation; only the model checkpoint differs. Table~\ref{tab:cross-stage} compares the pretrained (pt-448) and instruction-tuned (mix-448) deltas. Instruction tuning amplifies the causal contribution by roughly $3\times$ on average. Two features (L13/F15219, L12/F2257) reverse sign, acting as noise before instruction tuning but producing clear negative deltas afterward, indicating that these spatial behaviors are introduced during multimodal training rather than inherited from the pretrained variant.

\noindent\textbf{MMDiff~CAA steering.}\quad Following Sec.~\ref{sec:expsetup-steering}, Table~\ref{tab:steering} compares MMDiff~CAA with vanilla CAA across ten spatial features on the PaliGemma~2 base. MMDiff~CAA improves $\Delta$VSR by $+3.6$\% on average (peak $+15.4$\% on \emph{ahead of}). Decomposing the method on the same features, single-layer CAA gives $+8.96$, extending CAA to the discovered feature layers gives $+10.78$, and adding the feature's decoder direction gives $+12.59$, so layer selection and the injected direction contribute in comparable measure (App.~\ref{app:steer-decomp}). Non-improvements correspond to features already strongly encoded before fine-tuning or resistant to amplification.

\subsection{Multimodal Safety}
\label{sec:safety}

\noindent\textbf{Setup.}\quad We use VLSBench~\cite{hu2024vlsbench}, constructed so that harmful intent cannot be inferred from text alone (the model must integrate the image to recognize the unsafe intent). Attack-success rate (ASR) is judged by Qwen3-VL-8B-Instruct, which scores whether the response engages with the unsafe action given the instruction, image, and stated safety reason. We use two controls. First, a VQAv2 Yes/No subset ($\Delta$VQA) to verify general visual-question capability is preserved. Second, the safe split of MSSBench~\cite{zhou2024mssbench}, in which both instruction and image are benign by construction (76 embodied-action and 24 chat samples; baseline ASR $\approx 0$); $\Delta$Ctrl detects whether ablation causes spurious unsafe generation on benign inputs. $\mathcal{D}_{\text{tgt}}$ is the VLSBench unsafe split partitioned by safety category.

\noindent\textbf{Causal removal ablation.}\quad Following Sec.~\ref{sec:expsetup-ablation}, for each of the six VLSBench categories we identify the top unsafe feature in the adapted set $\mathcal{A}$ and ablate it. Table~\ref{tab:safety} reports the per-category top features on PaliGemma~2. Each top feature reduces VLSBench ASR by $17$--$28$\% with $|\Delta\text{VQA}| \le 1$\% and $\Delta\text{Ctrl} \le 1$\%, indicating that the safety reduction is targeted rather than a generic capability degradation. Across a sweep of $1{,}061$ candidate safety features, the mean effect is $\Delta$ASR $= -9.67$\%, $\Delta$VQA Acc $= -0.03$\%, $\Delta$Ctrl $= +0.41$\%; ablating MMDiff-discovered unsafe features therefore reduces attack-success rate without measurable capability spillover.

\begin{table}[h]
  \centering
  \scriptsize
  \setlength{\tabcolsep}{3pt}
  \begin{tabular}{>{\columncolor{tabMeta}}c>{\columncolor{tabMeta}}c>{\columncolor{tabMeta}}l>{\columncolor{tabAccent}}rrrr}
    \toprule
    \rowcolor{tabHeader}
    Layer & Feature & Category & $\Delta$VLSBench ASR & $\Delta$VQA Acc & $\Delta$Ctrl (MSSBench) & OR \\
    \midrule
    21 & 12020 & Self-Harm         & $-28.14$ & $+0.90$ & $+1.00$ & 8.12 \\
    23 & 13965 & Erotic            & $-26.59$ & $-0.70$ & $\phantom{-}0.00$ & 9.35 \\
    17 & 3967  & Privacy           & $-25.99$ & $-0.10$ & $\phantom{-}0.00$ & 4.15 \\
    13 & 5205  & Violent           & $-24.43$ & $-0.10$ & $\phantom{-}0.00$ & 5.66 \\
    9  & 9066  & Hate              & $-21.08$ & $-0.80$ & $+1.00$ & 5.06 \\
    7  & 5567  & Illegal Activity  & $-17.96$ & $+0.70$ & $\phantom{-}0.00$ & 8.17 \\
    \bottomrule
  \end{tabular}
  \caption{\textbf{Per-category top unsafe features on PaliGemma~2.} ASR drop of $17$--$28$\% per feature; no measurable cost on VQAv2 or MSSBench.}
  \label{tab:safety}
  \vspace{-1em}
\end{table}

\subsection{OCR}
\label{sec:ocr}

\noindent\textbf{Datasets.}\quad $\mathcal{D}_{\text{tgt}}$ is OCR-style prompts (e.g., \emph{``what does the sign say?''}) on VQAv2 images with legible embedded text. Evaluation is on OCRBench~\cite{liu2024ocrbench}, partitioned into its official categories (Scene Text-centric VQA, Non-Semantic Text, Digit String, Irregular Text). The contrastive firing analysis on MMDiff-Gemma yields $1{,}070$ OCR-selective features whose activations increase on OCR prompts and persist under neutral image-description prompts, indicating sensitivity to image-grounded text rather than to prompt-specific lexical patterns.

\noindent\textbf{Causal remove ablation.}\quad Following Sec.~\ref{sec:expsetup-ablation}, Table~\ref{tab:abl-ocr} reports $\Delta$Cat (drop on the feature's OCRBench category subset), $\Delta$VQA, and $\Delta$Ctrl on a VQA-clean non-OCR subset. Across five top features the mean $\Delta$Cat is $-16.9$\% with $|\Delta\text{VQA}| \le 1.6$\% and $|\Delta\text{Ctrl}| \le 1.8$\%, indicating targeted suppression of OCR capability without degrading general VQA performance.

\noindent\textbf{MMDiff~CAA steering.}\quad We adapt the steering recipe (Sec.~\ref{sec:expsetup-steering}) to OCR with two changes: (i) generation is open-ended, so we reformulate the task as a 4-way multiple-choice (``Answer: (A/B/C/D)'') with one ground-truth option and three distractors and steer at the decision token; (ii) the steering direction is built from $(\text{GT},\,\text{distorted-GT})$ answer pairs rather than from positive/negative VSR captions. Across all five OCR features (Table~\ref{tab:steer-ocr}), MMDiff~CAA improves by $+1.8$\% on average over vanilla CAA (peak $+10.58$\% on L17/F13602). The gains concentrate on features whose decoder direction is a strong standalone steering signal; on L19/F10089, where the isolated direction contributes least, MMDiff~CAA falls within $0.2$\% of vanilla CAA.

\begin{table*}[!t]
  \centering
  \scriptsize
  \begin{minipage}[t]{0.46\textwidth}
    \centering
    \setlength{\tabcolsep}{3pt}
    \renewcommand{\arraystretch}{0.92}
    \begin{tabular}{>{\columncolor{tabMeta}}c>{\columncolor{tabMeta}}c>{\columncolor{tabMeta}}l>{\columncolor{tabAccent}}rrr}
      \toprule
      \rowcolor{tabHeader}
      L & F & Category & $\Delta$OCRBench & $\Delta$VQA & $\Delta$Ctrl \\
      \midrule
      19 & 10089 & Scene Text & $-28.0$ & $\phantom{-}0.2$ & $-0.4$ \\
      17 & 13602 & Scene Text & $-16.5$ & $\phantom{-}0.9$ & $\phantom{-}0.6$ \\
      20 & 10687 & Non-Sem.\ & $-16.0$ & $\phantom{-}0.6$ & $\phantom{-}0.4$ \\
      21 & 9577  & Digit & $-14.0$ & $-1.6$ & $-1.8$ \\
      19 & 14093 & Irregular & $-10.0$ & $-0.5$ & $-1.0$ \\
      \bottomrule
    \end{tabular}
    \captionof{table}{\textbf{Top OCR features (PaliGemma~2 ablation).} $\Delta$Cat = drop on the feature's OCRBench category subset.}
    \label{tab:abl-ocr}
  \end{minipage}
  \hfill
  \begin{minipage}[t]{0.52\textwidth}
    \centering
    \setlength{\tabcolsep}{3pt}
    \renewcommand{\arraystretch}{0.92}
    \begin{tabular}{>{\columncolor{tabMeta}}c>{\columncolor{tabMeta}}c>{\columncolor{tabMeta}}l>{\columncolor{tabAccent}}rr}
      \toprule
      \rowcolor{tabHeader}
      L & F & Category & CAA & MMDiff~CAA \\
      \midrule
      21 & 9577  & Digit       & \cellcolor{tabLose}$+4.65$ & \cellcolor{tabWin}$\mathbf{+5.81}$ \\
      17 & 13602 & Scene Text  & \cellcolor{tabLose}$+2.88$ & \cellcolor{tabWin}$\mathbf{+10.58}$ \\
      19 & 14093 & Irregular   & \cellcolor{tabLose}$+1.55$ & \cellcolor{tabWin}$\mathbf{+1.68}$ \\
      19 & 10089 & Scene Text  & \cellcolor{tabWin}$+0.63$ & \cellcolor{tabLose}$+0.47$ \\
      20 & 10687 & Non-Sem.\   & \cellcolor{tabLose}$+1.31$ & \cellcolor{tabWin}$\mathbf{+1.55}$ \\
      \midrule
      \multicolumn{3}{l}{\textbf{Mean}} & \cellcolor{tabLose}$+2.21$ & \cellcolor{tabWin}$\mathbf{+4.02}$ \\
      \bottomrule
    \end{tabular}
    \captionof{table}{\textbf{OCR feature steering (PaliGemma~2 base).} CAA vs MMDiff~CAA on 5 OCR features}
    \label{tab:steer-ocr}
  \end{minipage}
  \vspace{-2.2em}
\end{table*}
\vspace{-0.4em}


\section{Ablations and Analyses}
\label{sec:additional}

\subsection{Does Model Diffing Matter?}
\label{sec:diffing-ablation}

Rotation and visual energy together form the \emph{adapted-feature filter} (Sec.~\ref{sec:adapted}). We test whether it is necessary from two directions: varying the selection rule over a fixed dictionary, and changing the dictionary itself.

\noindent\textbf{Feature selection methods.}\quad Each row of Table~\ref{tab:sel-ablation} is a complete selection rule evaluated end-to-end under the causal protocol of Sec.~\ref{sec:expsetup-ablation}. Dropping the adapted-feature filter gives \emph{larger} VSR drops but degrades general VQA by $24$--$26$\%, so those removals disrupt the model globally rather than isolating spatial computation; the lexical-invariance filter alone does not prevent this, and the adapted-feature filter alone leaves VQA intact but yields almost no task effect. Only the full pipeline is both causally effective and selective ($-12.3$\% VSR, $-0.1$\% VQA). Ablating randomly-selected features from the same layers moves VSR by $-0.5$\% (App.~\ref{app:random-baseline}), so the effect is carried by the selected directions rather than by the intervention itself. Rotation without visual energy selects no features.

\begin{wraptable}{r}{0.52\textwidth}
  \vspace{-1.2em}
  \centering
  \scriptsize
  \setlength{\tabcolsep}{3pt}
  \begin{tabular}{l@{\hskip 0.5em}cccc@{\hskip 0.5em}>{\columncolor{tabAccent}}rr}
    \toprule
    \rowcolor{tabHeader}
    Selection rule & F & $E_v$ & R & L & $\Delta$VSR & $\Delta$VQA \\
    \midrule
    Random features & $\times$ & $\times$ & $\times$ & $\times$ & $\phantom{-}-0.5$ & $\phantom{-}-0.2$ \\
    \midrule
    Firing only & $\checkmark$ & $\times$ & $\times$ & $\times$ & $-15.1$ & $-25.9$ \\
    \quad + vis.\ resp. & $\checkmark$ & $\checkmark$ & $\times$ & $\times$ & $-15.9$ & $-26.3$ \\
    \quad + lexical & $\checkmark$ & $\times$ & $\times$ & $\checkmark$ & $-14.6$ & $-24.4$ \\
    \quad + adapted & $\checkmark$ & $\checkmark$ & $\checkmark$ & $\times$ & $\phantom{-}-1.0$ & $\phantom{-}-0.2$ \\
    \textbf{Full MMDiff} & $\checkmark$ & $\checkmark$ & $\checkmark$ & $\checkmark$ & $\mathbf{-12.3}$ & $\mathbf{-0.1}$ \\
    \bottomrule
  \end{tabular}
  \caption{\textbf{Feature selection method ablation (PaliGemma~2).} Each method selects features from the same warm-started dictionary and is run end-to-end under the paper's causal protocol. F = contrastive firing, $E_v$ = visual energy, R = decoder rotation ($E_v$ and R together form the adapted-feature filter), L = lexical-invariance filter. The first row selects at random, isolating the effect of the intervention itself.}
  \label{tab:sel-ablation}
  \vspace{-1em}
\end{wraptable}

\noindent\textbf{A standard SAE trained on MLLM activations.}\quad We also train the dictionary itself from scratch: a randomly initialised SAE on LLaVA-MORE activations with identical data and hyperparameters, changing only the base-LM warm start. Without it there is no index correspondence to diff against, leaving contrastive firing over its own dictionary as the only selection route. Selection then degenerates: the top $10$ spatial features by odds ratio all lie in a single early layer and fire on $100$\% of VSR samples, since the odds ratio saturates once a feature fires on every task sample. Ablating them leaves VSR unchanged (mean $+0.22$, no feature beyond $\pm 1.4$), against $-10.11$ for MMDiff features on the same model (per-feature values in App.~\ref{app:scratch-sae}). The causally effective set is therefore not recovered by conventional MLLM SAE training alone; it comes from the adapted-feature filter, which requires diffing.

\subsection{Attribution Patching for Task-Specific Heads}
\label{sec:attr-patch}

\paragraph{Method.}
Attribution patching~\cite{nanda2023attribution} is a scalable alternative to activation patching~\cite{zhang2024activationpatching}, which measures causal effects by replacing activations with counterfactuals. Activation patching requires one forward pass per intervention; attribution patching uses a gradient-based linear approximation to estimate interventions with two forward and one backward pass, making it practical to probe attribution scores across layers and heads. We adapt this to identify attention heads driving task-specific SAE features. For a target feature $f$ at layer $L$, we project the layer-$L$ activations onto the SAE decoder vector to define a scalar objective; gradients of this objective with respect to upstream residuals and attention inputs, indicate how strongly each head contributes to $f$. We compare a clean run (original image--text input) against a corrupt run in which layer-$0$ visual token embeddings are replaced by a mean embedding computed over many VQA samples, suppressing visual content while preserving distributional statistics. The two attribution variants are $(\text{corr} - \text{clean}) \cdot \nabla_{\text{clean}}$ (Method~A) and $(\text{clean} - \text{corr}) \cdot \nabla_{\text{corr}}$ (Method~B); we report per-layer and per-head scores averaged over the top-$k$ samples that most strongly activate $f$. See App.~\ref{app:attribution-formalism} for details.

\begin{figure*}[h]
  \centering
  \includegraphics[width=0.92\linewidth]{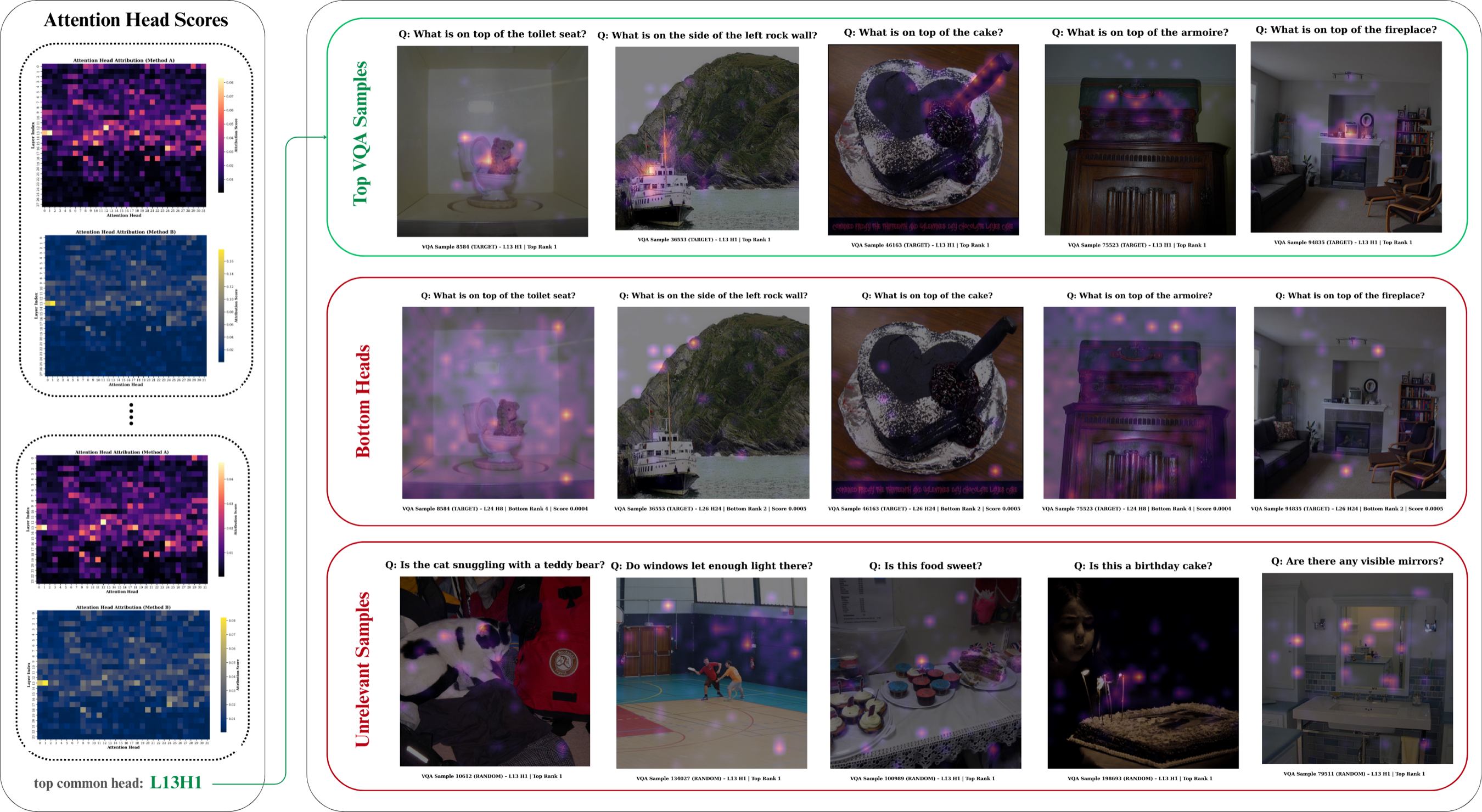}
  \caption{\textbf{Attribution patching across related spatial features.} Top: a recurring top-scoring head (L13H1) localizes to relevant regions in queries about ``on top of'' relations. Middle: bottom-ranked heads on the same samples fail to capture spatial structure. Bottom: unrelated queries confirm that the top head does not spuriously activate.}
  \label{fig:ap-main}
\end{figure*}

\paragraph{Results.}
Across the spatially selective features we examined, layer-wise attribution curves typically peak in middle layers, consistent with the layer distribution of MMDiff-discovered spatial features (Fig.~\ref{fig:layer-agg}). At the head level, both methods highlight a small subset of heads with notably high scores, and the top heads identified are largely consistent across methods (Fig.~\ref{fig:head-agg}). Some of the same heads recur across related spatial relations: in the top row of Fig.~\ref{fig:ap-main}, head L13H1 attends to semantically relevant regions across queries about ``on top of''; the middle row confirms that bottom-ranked heads on the same samples fail to localize meaningfully, and the bottom row confirms that unrelated queries do not trigger spurious activation. The clustering of driving heads near each feature's home layer provides the mechanistic justification for the layer-targeted injections used for steering in Sec.~\ref{sec:spatial}.

\vspace{-0.5em}

\subsection{Auto-Interpretation}
\label{sec:auto-interp}

\begin{figure}[h]
  \centering
  \includegraphics[width=0.95\linewidth]{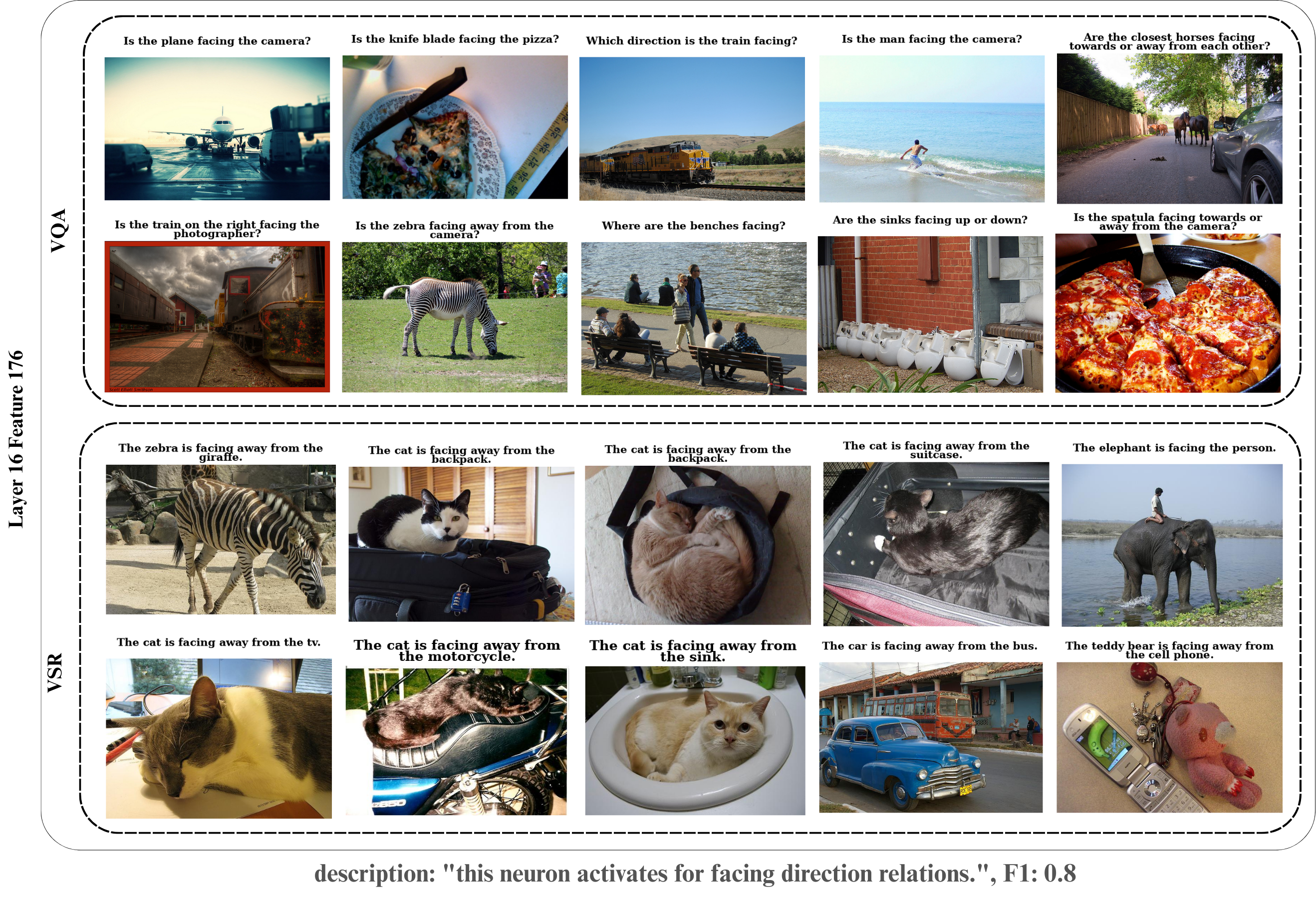}
  \caption{\textbf{Auto-Interp example (Layer 16, Feature 176, MMDiff-Llama).} Top VQA and VSR samples highlight \emph{facing direction}, activating on objects facing toward, away, or relative to others.}
  \label{fig:auto-interp-example}
  \vspace{-0.8em}
\end{figure}

For each MMDiff-discovered feature we collect its top-activating samples from VQAv2 and from VSR~\cite{liu2023vsr}, and pass them to GPT-4o-mini~\cite{gpt-4o-mini} to obtain a short natural-language description and an F1-based confidence score from a held-out classification task. The resulting labels are stored alongside the contrastive-firing metrics from Sec.~\ref{sec:taskspec} and lightly reviewed by hand. Auto-interpretation is used here as a qualitative validation layer rather than as a primary contribution; Figure~\ref{fig:auto-interp-example} shows a representative example, and additional examples are in App. \ref{app:auto-interp}.

\section{Related Work}
\label{sec:related}

\noindent\textbf{Model Diffing.}\quad Model diffing studies how internal representations change across models or training stages. Earlier work largely compared models at the representation level, through function-space visualization, model stitching, or similarity metrics \citep{olah2015visualizing, erhan2010, lenc2015understanding, bansal2021revisiting, kornblith2019similarity, barannikov2021representation}, while later studies also examined convergent neurons and feature-like units \citep{li2015convergent, olah2020zoom}. SAEs provide a feature-level lens, and \citet{kissane2024saes} show base-LM SAE dictionaries largely transfer to fine-tuned models. Stage-wise model diffing \citep{Bricken2024StageWiseModelDiffing} retrains SAEs across checkpoints with aligned feature indices, revealing dynamics including sleeper-agent features \citep{hubinger2024sleeper, minder2025robustly}. Crosscoder-based diffing \citep{Lindsey2024SparseCrosscoders, mishrasharma2025crosscoderinsights, jiralerspong2026difftool} and diff-SAEs \citep{minder2025diff, aranguri2025diffsae} are alternative formulations. We extend feature-level model diffing to the LM$\rightarrow$MLLM transition, using feature changes as discovery signals for downstream control.

\noindent\textbf{Mechanistic Interpretability and SAEs on MLLMs.}\quad Prior MLLM-internals work spans tool-based and causal explanations \citep{stan2024lvlm, basu2024understanding, palit2023towards} and probing- and feature-based analyses \citep{tong2024eyes, gandelsman2023interpreting, chen2023interpreting, schwettmann2023multimodal, jiang2024interpreting, neo2024towards, venhoff2025toolate, khayatan2025analyzing, venhoff2025visual, lin2025survey, zhang2025crossmodal, joseph2024bridging}. SAE-on-MLLM work targets either the vision encoder or the LM backbone \citep{daujotas2024clip, joseph2025steering, pach2025sparse, lim2025patchsae, joseph2025prisma, olson2025probing, cornet2025explaining, kissane2024saes, kulkarni2025cbsae}. These studies analyze multimodal representations, alignment, or steering interfaces, but to our knowledge none combines a base-LM SAE with an MLLM-adapted SAE and uses their difference as a discovery signal.

\noindent\textbf{Activation Steering and Multimodal Safety.}\quad Activation steering modifies the residual stream along a chosen direction \citep{turner2023activation, li2023inference, rimsky2024steering, arditi2024refusal}; CAA \citep{rimsky2024steering} uses mean activation differences at a fixed mid-layer, task/function vectors target specific layers \citep{hendel2023incontext, todd2024function, meng2022locating}, and SAE-based methods steer along feature directions \citep{chalnev2024improving, anthropic2024goldengate, marks2025sparse}. Multimodal extensions inject such directions into MLLMs and VLAs \citep{sivakumar2025steervlm, haon2025vla, grant2026notall, buurmeijer2026observing, qwen2026qwenscope}. Multimodal safety benchmarks \citep{liu2024mmsafetybench, hu2024vlsbench, gong2025figstep, qi2024visual, li2024hades, chao2024jailbreakbench} pair with prompt-level safeguards \citep{wang2024adashield}, latent-space steering \citep{wang2025astra, zeng2025safesteer}, and refusal-direction interventions \citep{obrien2024steering, prakash2025dissecting}. MMDiff-CAA differs by steering with decoder directions of MMDiff-discovered features at their feature-associated layers rather than relying only on a generic single mid-layer direction.


\section{Limitations}
\label{sec:limitations}

We instantiate MMDiff on three backbones (LLaMA-3.1-8B, Gemma-2-2B, and Qwen3-1.7B); safety and OCR are evaluated on PaliGemma~2 only. Applying the full recipe to additional MLLM families, including larger backbones, mixture-of-experts variants, and Qwen-VL or Pixtral-style architectures, is a natural next step. MMDiff~CAA assumes access to an instruction-tuned reference model from which the steering direction can be extracted; in settings where only the base or only the instruction-tuned model is available, the recipe reduces to standard SAE-feature steering. Finally, a minority of safety candidates cause generation collapse rather than refusal when ablated, and currently require a post-hoc filter to exclude.


\section{Conclusion}
\label{sec:conclusion}

We presented MMDiff, a model-diffing pipeline for MLLMs that isolates multimodal-adapted features and uses them as targets for causal ablation, MMDiff~CAA steering, auto-interpretation, and attribution patching. Across three MLLMs (LLaVA-MORE, PaliGemma~2, InternVL3.5-2B) and three domains (spatial reasoning, multimodal safety, OCR), the same recipe supports behavior control through ablation and improved steering through layer-targeted intervention, with each application instantiated by varying only the target distribution. Cross-stage ablation indicates that mid-layer spatial features in PaliGemma~2 are predominantly acquired during multimodal fine-tuning rather than inherited from the base LM. More broadly, our results position multimodal SAEs as feature-level interfaces for auditing, localizing, and controlling multimodal behavior in MLLMs. The pipeline extends naturally to additional MLLM families and to domains including embodied-AI safety, visual mathematical reasoning, and medical-image grounding. We hope MMDiff provides a useful foundation for future work on mechanistic understanding and intervention in multimodal systems, including isolating misaligned features and steering toward safer generations.

\section*{Acknowledgments}
We acknowledge Cosmos Institute and Modal Academics credits grant for providing access to compute resources that assisted in training the Multimodal SAEs used in our work. This work was supported in part by Advanced Micro Devices, Inc. under the AMD University Program’s AI \& HPC Cluster.






\bibliographystyle{plainnat}
\bibliography{references}

\clearpage
\appendix

\begin{center}
{\LARGE \textbf{Appendix: Table of Contents}}
\end{center}
\medskip

\begin{flushleft}
\begin{tabularx}{\linewidth}{@{}lXr@{}}
\textbf{A} & \textbf{Preliminaries} & \pageref{app:preliminaries} \\
 & A.1 \quad Multimodal Large Language Models & \pageref{app:prelim-vlm} \\
 & A.2 \quad Sparse Autoencoders & \pageref{app:prelim-sae} \\
 & A.3 \quad Stage-Wise Model Diffing for MLLMs & \pageref{app:prelim-diffing} \\[4pt]

\textbf{B} & \textbf{Algorithm: The MMDiff Pipeline} & \pageref{app:algorithm} \\
 & B.1 \quad MMDiff~CAA Steering & \pageref{app:algorithm-steer} \\[4pt]

\textbf{C} & \textbf{Multimodal SAE Training} & \pageref{app:sae-training-diagnostics} \\
 & C.1 \quad Training regimes & \pageref{app:training-regimes} \\
 & C.2 \quad Optimization and configurations & \pageref{app:sae-config} \\
 & C.3 \quad Reconstruction quality (MMDiff-Llama) & \pageref{app:recon-llama} \\
 & C.4 \quad Decoder geometry across regimes & \pageref{app:rotations} \\
 & C.5 \quad Seed stability of the learned dictionary & \pageref{app:seed-stability} \\[4pt]

\textbf{D} & \textbf{Adapted Feature Selection Diagnostics} & \pageref{app:adapted-selection-scatter} \\
 & D.1 \quad Joint visual-energy and cosine distribution & \pageref{app:joint-ev-cos} \\
 & D.2 \quad Per-layer adapted-feature statistics & \pageref{app:adapted-perlayer} \\
 & D.3 \quad Threshold sweep for feature selection & \pageref{app:threshold-sweep} \\
 & D.4 \quad Filtering funnel for task-specific feature discovery & \pageref{app:filtering-funnel} \\
 & D.5 \quad Distribution-shift visualizations & \pageref{app:dist-shift} \\
 & D.6 \quad Lexical-invariance prompt banks & \pageref{app:lexical-prompts} \\
 & D.7 \quad Image counterfactuals & \pageref{app:image-counterfactuals} \\
 & D.8 \quad Standard SAE trained directly on MLLM activations & \pageref{app:scratch-sae} \\
 & D.9 \quad Randomly-selected feature ablation & \pageref{app:random-baseline} \\[4pt]

\textbf{E} & \textbf{Steering Decomposition and Feature Correspondence} & \pageref{app:steer-matching} \\
 & E.1 \quad Decomposing the steering gains & \pageref{app:steer-decomp} \\
 & E.2 \quad Feature correspondence across dictionaries & \pageref{app:matching} \\[4pt]

\textbf{F} & \textbf{Auto-Interpretation} & \pageref{app:auto-interp} \\
 & F.1 \quad Auto-Interpretation: Examples & \pageref{app:auto-interp-examples} \\
 & F.2 \quad Auto-Interpretation and Scoring Pipeline & \pageref{app:auto-interp-pipeline} \\[4pt]

\textbf{G} & \textbf{Attribution Patching: Aggregated and Per-Feature Panels} & \pageref{app:attribution} \\
 & G.1 \quad Formalism & \pageref{app:attribution-formalism} \\
 & G.2 \quad Bottom-Ranked Heads as a Control & \pageref{app:bottom-heads} \\[4pt]

\textbf{H} & \textbf{OCR Feature Examples} & \pageref{app:ocr} \\
\end{tabularx}
\end{flushleft}

\noindent\rule{\linewidth}{0.4pt}
\bigskip

\section{Preliminaries}
\label{app:preliminaries}

This appendix reviews the main concepts underlying MMDiff. We first introduce the multimodal large language model (MLLM) setting and the multimodal fine-tuning pipeline used in modern multimodal large language models (MLLMs). We then review Sparse Autoencoders (SAEs) as interpretable feature dictionaries over transformer residual streams, including the sparsity mechanisms used in the SAE suites employed here. Finally, we introduce stage-wise model diffing and explain how aligned SAE dictionaries enable feature-level analysis of multimodal adaptation.

\subsection{Multimodal Large Language Models}
\label{app:prelim-vlm}

We use \emph{multimodal large language model} (MLLM) throughout this paper to refer to a pretrained language model extended with a visual encoder and projector. Our experiments use MLLMs whose language backbone is a text-pretrained LM (LLaMA, Gemma) for which a base-LM SAE suite is publicly available; pairing each MLLM with a matching base-LM SAE dictionary on the same backbone is what makes stage-wise model diffing tractable.

\paragraph{Architecture.}
An MLLM consists of three components: a visual encoder $f_V$, a pretrained language model $f_{\mathrm{LM}}$, and a trainable projector $P$. The visual encoder, typically a Vision Transformer \citep{radford2021learning, zhai2023sigmoid}, extracts patch embeddings
\[
V = f_V(x) = [v_1, \ldots, v_{N_V}],
\]
where $x$ is an input image and $N_V$ is the number of visual tokens (which depends on the image resolution and patch size of $f_V$). The projector maps $V$ into the LM token space:
\[
\tilde{V} = P(V) = [\tilde{v}_1, \ldots, \tilde{v}_{N_V}], \qquad \tilde{v}_i \in \mathbb{R}^{d},
\]
where $d$ is the hidden dimension of $f_{\mathrm{LM}}$. The projected image tokens are concatenated with tokenized text embeddings $T = [t_1, \ldots, t_{N_T}]$ to form the multimodal input
\[
X = [\tilde{v}_1, \ldots, \tilde{v}_{N_V}, t_1, \ldots, t_{N_T}].
\]
Visual tokens come first; this ordering matters for our analyses, since it lets us cleanly mask token spans by modality during SAE training and during downstream interventions.

\paragraph{Models studied.}
We instantiate MMDiff on three MLLMs of different backbone families.

\begin{itemize}
    \item \textbf{LLaVA-MORE} \citep{cocchi2025llava} extends the LLaVA framework \citep{liu2023visual, liu2024improved} by integrating recent language models with diverse visual backbones. We use the variant combining the CLIP ViT-Large-Patch14--336 encoder \citep{radford2021learning} with a LLaMA-3.1-8B language backbone \citep{grattafiori2024llama}.
    \item \textbf{PaliGemma~2} \citep{steiner2024paligemma2} combines a SigLIP-So400m vision encoder \citep{zhai2023sigmoid} with the Gemma-2-2B language backbone \citep{riviere2024gemma2}.
    \item \textbf{InternVL3.5-2B} \citep{wang2025internvl35advancingopensourcemultimodal} pairs an InternViT vision encoder with a Qwen3-1.7B language backbone \citep{yang2025qwen3technicalreport}. We diff against a Qwen-Scope Top-$K$ SAE for that backbone \citep{qwen2026qwenscope} (width $32{,}768$, $k=50$).
\end{itemize}

The three MLLMs differ along several axes that matter for evaluating the generality of MMDiff: backbone family (LLaMA vs.\ Gemma vs.\ Qwen), vision-encoder objective (CLIP contrastive vs.\ SigLIP pairwise), hidden size, depth, and the family of base-LM SAE suite available for each backbone (LLaMA-Scope TopK SAEs \citep{he2024llama} for LLaVA-MORE, Gemma-Scope JumpReLU SAEs \citep{lieberum2024gemmascope} for PaliGemma~2, and Qwen-Scope Top-$K$ SAEs \citep{qwen2026qwenscope} for the Qwen3 backbone of InternVL3.5-2B). Where a property is shared across the models we report it as a property of MMDiff; where it is specific to one MLLM we say so explicitly.

\subsection{Sparse Autoencoders}
\label{app:prelim-sae}

\paragraph{Motivation.}
Internal representations of large language models exhibit \emph{superposition}: more features are encoded than there are neuron dimensions, with many features sharing the same residual-stream coordinates \citep{elhage2022superposition}. As a consequence, individual neurons are typically polysemantic, and per-neuron analyses confound multiple functional roles. Sparse Autoencoders (SAEs) attempt to undo superposition by learning an overcomplete dictionary of feature directions in which each input activation is approximated by a small number of active features \citep{bricken2023monosemanticity, cunningham2023sparse}. The dictionary directions are not constrained to align with neurons, which lets them recover finer-grained, often more interpretable units of computation.

\paragraph{Vanilla SAE.}
Given an activation $x \in \mathbb{R}^{D}$ taken from a transformer's residual stream, a vanilla SAE consists of an encoder and a decoder,
\[
h(x) = \mathrm{ReLU}(W_{\mathrm{enc}}\, x + b_{\mathrm{enc}}),
\qquad
\hat{x} = W_{\mathrm{dec}}\, h(x) + b_{\mathrm{dec}},
\]
with $W_{\mathrm{enc}} \in \mathbb{R}^{F \times D}$, $b_{\mathrm{enc}} \in \mathbb{R}^{F}$, $W_{\mathrm{dec}} \in \mathbb{R}^{D \times F}$, and $b_{\mathrm{dec}} \in \mathbb{R}^{D}$. The dictionary size $F$ is typically chosen larger than the input dimension $D$ (overcomplete). Each decoder column
\[
v_f \;=\; (W_{\mathrm{dec}})_{:,f} \;\in\; \mathbb{R}^{D}
\]
defines feature $f$'s direction in residual-stream space, and the corresponding encoder row $(W_{\mathrm{enc}})_{f,:}$ acts as a detector that determines when $f$ is present in the input. We refer to $h_f(x)$ as feature $f$'s activation strength on $x$.

Training minimizes a reconstruction term plus a sparsity penalty,
\[
\mathcal{L}(x) \;=\; \|x - \hat{x}\|_{2}^{2} \;+\; \lambda \sum_{f=1}^{F} |h_f(x)|,
\]
where $\lambda$ trades off reconstruction quality against $L_1$ sparsity.

\paragraph{TopK SAEs.}
TopK SAEs \citep{gao2024scaling} replace the $L_1$ penalty with a hard top-$k$ operator on the encoder pre-activations:
\[
h(x) \;=\; \mathrm{TopK}_k\!\bigl(W_{\mathrm{enc}}\, x + b_{\mathrm{enc}}\bigr),
\]
which keeps the $k$ largest pre-activations and zeros the rest. This yields an exact-$k$ sparsity guarantee per token: $\|h(x)\|_0 = k$ for every $x$. Since sparsity is enforced by the architecture rather than by a tunable penalty, TopK SAEs decouple the rate of feature firing from the reconstruction objective. The LLaMA-Scope suite \citep{he2024llama} releases TopK SAEs for LLaMA-3.1-8B and the Qwen-Scope suite \citep{qwen2026qwenscope} for Qwen3-1.7B, both at fixed values of $k$; we use $k = 50$ in MMDiff-Llama and MMDiff-Qwen.

\paragraph{JumpReLU SAEs.}
JumpReLU SAEs \citep{rajamanoharan2024jumprelu} use a learned per-feature threshold $\theta_f$ and fire feature $f$ only when its pre-activation $z_f(x) = (W_{\mathrm{enc}})_{f,:}\, x + (b_{\mathrm{enc}})_f$ exceeds $\theta_f$:
\[
h_f(x) \;=\;
\begin{cases}
z_f(x) & \text{if } z_f(x) > \theta_f, \\
0 & \text{otherwise.}
\end{cases}
\]
The thresholds $\theta_f$ are learned jointly with the encoder/decoder weights using a straight-through estimator, targeting an average sparsity $\ell_0$ (the expected number of active features per token). Unlike TopK, JumpReLU does not enforce fixed per-token sparsity; instead, sparsity fluctuates around the target. For MMDiff-Gemma, we use the Gemma-Scope JumpReLU SAE with target $\ell_0 = 50$ \citep{lieberum2024gemmascope}.

\subsection{Stage-Wise Model Diffing for MLLMs}
\label{app:prelim-diffing}

\paragraph{Setup.}
Stage-wise model diffing \citep{Bricken2024StageWiseModelDiffing} extends SAE-based interpretability across training stages by re-training dictionaries on activations from successive checkpoints of the \emph{same} architecture, while keeping feature indices aligned across stages. Aligned indices mean that feature $f$ in the stage-$A$ SAE and feature $f$ in the stage-$B$ SAE are intended to refer to the same conceptual unit, so a feature's evolution can be tracked by quantities such as the cosine similarity between its decoder directions, $\cos\!\bigl(v_f^{(A)}, v_f^{(B)}\bigr)$, or by changes in its activation pattern on a fixed dataset. This per-feature alignment lets one ask, for each direction $f$, whether it is preserved (high cosine, similar firing), rotated (low cosine, similar role), repurposed (low cosine, different firing), or newly emergent.

\paragraph{Comparison with crosscoders.}
An alternative to stage-wise diffing is crosscoder-based model diffing \citep{Lindsey2024SparseCrosscoders}, which trains a single SAE-like model that simultaneously reconstructs activations from multiple checkpoints, sharing a feature dictionary across them. Crosscoders give a single global decomposition and are convenient when the goal is to locate features that are systematically shared or unique across many models. Stage-wise diffing instead trains one dictionary per stage and aligns features post hoc, which provides finer per-feature resolution: features can rotate or specialize in ways that are visible at the dictionary level but would be averaged away by a single shared decoder. Recent analyses \citep{mishrasharma2025crosscoderinsights} report that crosscoders can have lower sensitivity for sparse adapted features (those that fire infrequently and account for a small fraction of activation variance), which is precisely the regime we operate in, since single spatial relations and per-category safety triggers are rare on a generic VQA distribution.

\paragraph{Why warm-starting from the base-LM SAE works.}
\citet{kissane2024saes} show that SAE dictionaries trained on a base LM largely transfer to its fine-tuned counterparts: most features remain aligned and a relatively small fraction is meaningfully reshaped. This empirical finding has two consequences for MMDiff. First, it justifies initializing the MLLM-adapted SAE from the base-LM SAE rather than retraining from scratch, since the warm start preserves the monosemantic features already learned for the language backbone and only the multimodally affected subset needs to be tracked. Second, it explains why the adapted feature set is small enough to be useful as a discovery signal: if every feature were rotated by adaptation, the diff would collapse back to the full dictionary and lose its specificity. In our experiments (Sec.~\ref{sec:adapted}), the adapted set typically contains around $5\%$ of features, which is consistent with the largely-transfer picture.

\paragraph{Limitations.} Stage-wise diffing assumes the base LM and adapted model share an architecture and vocabulary, which holds for the LM,$\rightarrow$,MLLM transition studied here (the language backbone is unchanged; only a vision encoder, projector, and multimodal fine-tuning are added). It also assumes adaptation induces feature-level changes rather than wholesale residual-space rotations. Our diagnostics (App.~\ref{app:sae-training-diagnostics}) support this most strongly for text-only SAEs, which preserve alignment with the base-LM dictionary and are therefore used for diffing. In regimes where these assumptions weaken (e.g., early-layer image-only SAEs), we observe larger decoder rotations and reduced feature alignment.

\section{Algorithm: The MMDiff Pipeline}
\label{app:algorithm}

Algorithm~\ref{alg:mmdiff} summarizes the full MMDiff pipeline as referenced in the main paper Sec.~\ref{sec:method}. Stage~1 trains a multimodal SAE warm-started from the base-LM dictionary; Stage~2 selects features that have rotated under multimodal training and prefer visual input; Stage~3 isolates a task-specific subset by contrasting per-token firing under a target distribution against a generic VQA baseline, with a Fisher-exact selectivity test (Benjamini--Hochberg correction) and a lexical-invariance filter. Figure~\ref{fig:pipeline} illustrates the MMDiff pipeline, and Figure~\ref{fig:main_demo} demonstrates it in action.

\begin{algorithm}[h]
\caption{MMDiff: Multimodal Model Diffing for Feature Discovery}
\label{alg:mmdiff}
\begin{algorithmic}[1]
\Require base-LM SAE $\mathcal{S}_{\mathrm{base}}$; multimodal model $\mathcal{M}_{\mathrm{vlm}}$; VQAv2 mix $\mathcal{D}_{\mathrm{base}}$; target distribution $\mathcal{D}_{\mathrm{tgt}}$; thresholds $\varepsilon, q, \tau_{\mathrm{OR}}, \tau_{\Delta p}$
\Ensure task-specific feature set $\mathcal{T}$
\Statex \textbf{\emph{Stage 1: Adapt SAE to MLLM activations}}
\State $\mathcal{S}_{\mathrm{vlm}} \leftarrow \textsc{AdaptSAE}(\mathcal{S}_{\mathrm{base}},\, \mathcal{M}_{\mathrm{vlm}},\, \mathcal{D}_{\mathrm{base}})$ \Comment{warm-start, fine-tune}
\State validate per-layer FVU; confirm text-only regime converges
\Statex \textbf{\emph{Stage 2: Identify adapted features}}
\For{each feature $f$ in $\mathcal{S}_{\mathrm{vlm}}$}
  \State $c_f \leftarrow \cos\!\bigl(W_{\mathrm{dec}}^{\mathrm{base}}[f],\, W_{\mathrm{dec}}^{\mathrm{vlm}}[f]\bigr)$ \Comment{decoder geometry}
  \State $E_v(f) \leftarrow \mathbb{E}_{\mathrm{vis}}[\,h_f(x)^2\,]$ \Comment{visual energy}
\EndFor
\State $\mathcal{A} \leftarrow \{f : E_v(f) > \varepsilon \,\wedge\, c_f \in \text{bottom-}q \text{ quantile}\}$
\Statex \textbf{\emph{Stage 3: Discover task-specific features}}
\For{each $f \in \mathcal{A}$}
  \State $p_f^{\mathrm{base}} \leftarrow$ \Call{PerTokenFiringRate}{$f, \mathcal{D}_{\mathrm{base}}$}
  \State $p_f^{\mathrm{tgt}} \leftarrow$ \Call{PerTokenFiringRate}{$f, \mathcal{D}_{\mathrm{tgt}}$}
  \State $\mathrm{OR}_f \leftarrow $ \Call{FisherExactTest}{$p_f^{\mathrm{base}}, p_f^{\mathrm{tgt}}$} \Comment{Benjamini--Hochberg correction}
\EndFor
\State $\mathcal{C} \leftarrow \{f \in \mathcal{A} : \mathrm{OR}_f \geq \tau_{\mathrm{OR}} \,\wedge\, \Delta p_f \geq \tau_{\Delta p}\}$ \Comment{statistical candidates}
\State $\mathcal{L} \leftarrow \{f \in \mathcal{C} : f \text{ fires on neutral lexical-invariance prompts}\}$ \Comment{lexical filter}
\State $\mathcal{T} \leftarrow \mathcal{C} \cap \mathcal{L}$
\State \Return $\mathcal{T}$
\end{algorithmic}
\end{algorithm}

\subsection{MMDiff~CAA Steering}
\label{app:algorithm-steer}

Algorithm~\ref{alg:mmdiff-caa} gives the steering recipe referenced in Sec.~\ref{sec:expsetup-steering}: extract a feature direction $v_f$ from $\mathcal{M}_{\mathrm{vlm}}$, combine it with the canonical CAA mean-activation contrast injected across a backbone layer set $\mathcal{L}$, and apply the intervention to $\mathcal{M}_{\mathrm{base}}$. The OCR variant differs only in the contrast cache (GT vs.\ distorted-GT pairs) and the injection token (decision position rather than each generated token).

\begin{algorithm}[h]
\caption{MMDiff~CAA Steering}
\label{alg:mmdiff-caa}
\begin{algorithmic}[1]
\Require feature $f$ at layer $\ell_f$ with decoder direction $v_f$ from $\mathcal{S}_{\mathrm{vlm}}$; backbone-CAA layer set $\mathcal{L}$; positive/negative prompt sets $P^+, P^-$ from $\mathcal{D}_{\mathrm{tgt}}$; intervention scales $\alpha, \gamma_f$; target model $\mathcal{M}_{\mathrm{base}}$
\Ensure steered $\mathcal{M}_{\mathrm{base}}$ residual stream
\For{each layer $\ell \in \mathcal{L}$}
  \State $\bar{h}^+_{\ell} \leftarrow \mathbb{E}_{x \in P^+}[\,h_{\ell}(x;\, \mathcal{M}_{\mathrm{vlm}})\,]$
  \State $\bar{h}^-_{\ell} \leftarrow \mathbb{E}_{x \in P^-}[\,h_{\ell}(x;\, \mathcal{M}_{\mathrm{vlm}})\,]$
  \State $d_{\ell} \leftarrow (\bar{h}^+_{\ell} - \bar{h}^-_{\ell}) / \|\bar{h}^+_{\ell} - \bar{h}^-_{\ell}\|$ \Comment{unit-normalize}
\EndFor
\State At inference on $\mathcal{M}_{\mathrm{base}}$, for each text token position $t$:
\For{each layer $\ell \in \mathcal{L}$}
  \If{$\ell = \ell_f$}
    \State $h_{\ell, t} \leftarrow h_{\ell, t} + \alpha\, d_{\ell} + \gamma_f\, v_f$ \Comment{backbone CAA + feature amplification}
  \Else
    \State $h_{\ell, t} \leftarrow h_{\ell, t} + \alpha\, d_{\ell}$ \Comment{backbone CAA only}
  \EndIf
\EndFor
\end{algorithmic}
\end{algorithm}

\section{Multimodal SAE Training}
\label{app:sae-training-diagnostics}

\subsection{Training regimes}
\label{app:training-regimes}

For each MLLM, we train SAEs on cached hidden states from 50k VQAv2 image--question pairs. The full input sequence contains projected visual tokens followed by text tokens, allowing token-type masks to determine which positions contribute to the reconstruction loss. We compare three masked training regimes: \emph{full-sequence}, where all tokens contribute; \emph{image-only}, where only projected visual-token positions contribute; and \emph{text-only}, where only non-visual token positions contribute. In all three masked regimes, the SAE is warm-started from the matching base-LM SAE suite, receives hidden states from the same MLLM forward pass, and is trained under identical optimizer settings and schedules; only the loss mask differs.

\paragraph{Random-initialization control.}
As a strict initialization ablation we also train a fourth variant (\emph{random}) whose SAE encoder, decoder, and bias parameters are sampled from the SAE-suite's default initialization scheme rather than warm-started from the base-LM SAE checkpoint. Random-init SAEs use the \emph{full-sequence} loss (no token-type masking), the same Adam optimizer settings, the same learning rate, the same width-based LR scaling (which is constant across layers in our setup since each layer's SAE shares the same dictionary width), the same number of cached training tokens, and the same per-chunk training schedule as the warm-started variants. Unlike the three masking regimes, the random-init control does not separate modality-specific contributions; its purpose is to verify that adaptation benefits from a pretrained base-LM dictionary rather than from any sufficiently parameterized sparse code. The consistently higher FVU of the random-init variant in Fig.~\ref{fig:fvu-curves} and Fig.~\ref{fig:fvu-layerwise} confirms this.

\paragraph{Why text-only masking for model diffing.}
MMDiff asks which \emph{language-backbone} features are repurposed by multimodal training, not which features reconstruct projected visual tokens most accurately. For stage-wise diffing to be meaningful, the adapted SAE must remain aligned with the base-LM dictionary so that feature-wise decoder comparisons remain interpretable. Text-only masking best preserves this alignment: text-token activations remain close to the original LM basis while still reflecting visual context through cross-modal attention, whereas full-sequence and image-only training rotate the dictionary toward projector-induced visual activations and weaken feature correspondence. This interpretation is consistent with the lower FVU and higher decoder cosine alignment of the text-only regime in Figs.~\ref{fig:fvu-curves}--\ref{fig:decoder-cos-trends}.

\subsection{Optimization and configurations}
\label{app:sae-config}

Training uses Adam ($\beta_1=0$, $\beta_2=0.999$, learning rate $7\times10^{-5}$ with $1{,}000$-step linear warmup) on cached activations from $50{,}000$ VQAv2 image--question pairs, sharded into $1{,}000$-sample chunks. For LLaVA-MORE we adapt LLaMA-Scope Top-$K$ SAEs with $k=50$ at training batch size $32$; for PaliGemma~2 we adapt Gemma-Scope JumpReLU SAEs with target $\ell_0=50$, bandwidth $0.001$ for the straight-through threshold gradient, and sparsity-penalty coefficient $1.0$ ramped over $2{,}000$ steps, at training batch size $8$ (memory-bound by the $16{,}384$-feature dictionary). All SAEs are trained with one job per layer in parallel on $8\times$ A100 80GB. Reconstruction is evaluated on a held-out split via fraction of variance unexplained (FVU), $\mathrm{FVU} = \frac{\mathbb{E}[\|x-\hat{x}\|_2^2]}{\mathbb{E}[\|x-\mathbb{E}[x]\|_2^2]}$, which measures normalized reconstruction error (lower is better); achieved sparsity and architectural settings are summarized in Table~\ref{tab:mmdiff-saes}.

\begin{table}[h]
  \centering
  \footnotesize
  \setlength{\tabcolsep}{3pt}
  \renewcommand{\arraystretch}{0.92}
  \begin{tabular}{lllllrrrr}
    \toprule
    \textbf{MMDiff SAE} & \textbf{MLLM} & \textbf{Backbone} & \textbf{Vision enc.} & \textbf{SAE} & \textbf{L} & \textbf{$d$} & \textbf{Width} & \textbf{Sparsity} \\
    \midrule
    MMDiff-Llama & LLaVA-MORE & LLaMA-3.1-8B & CLIP ViT-L/14-336 & TopK & 32 & 4096 & 32K & $k{=}50$ \\
    MMDiff-Gemma & PaliGemma~2 & Gemma-2-2B & SigLIP-So400m & JumpReLU & 26 & 2304 & 16K & $\ell_0{=}50$ \\
    \bottomrule
  \end{tabular}
  \caption{\textbf{MMDiff SAE configurations.} Each SAE is warm-started from the matching base-LM suite and adapted on cached VQAv2 activations across all backbone layers.}
  \label{tab:mmdiff-saes}
\end{table}

\subsection{Reconstruction quality (MMDiff-Llama)}
\label{app:recon-llama}

\begin{figure}[h]
  \centering
  \begin{minipage}[c]{0.57\linewidth}
    \vspace{0pt}
    \centering
    \includegraphics[width=\linewidth,clip,trim=0 0 0 8]{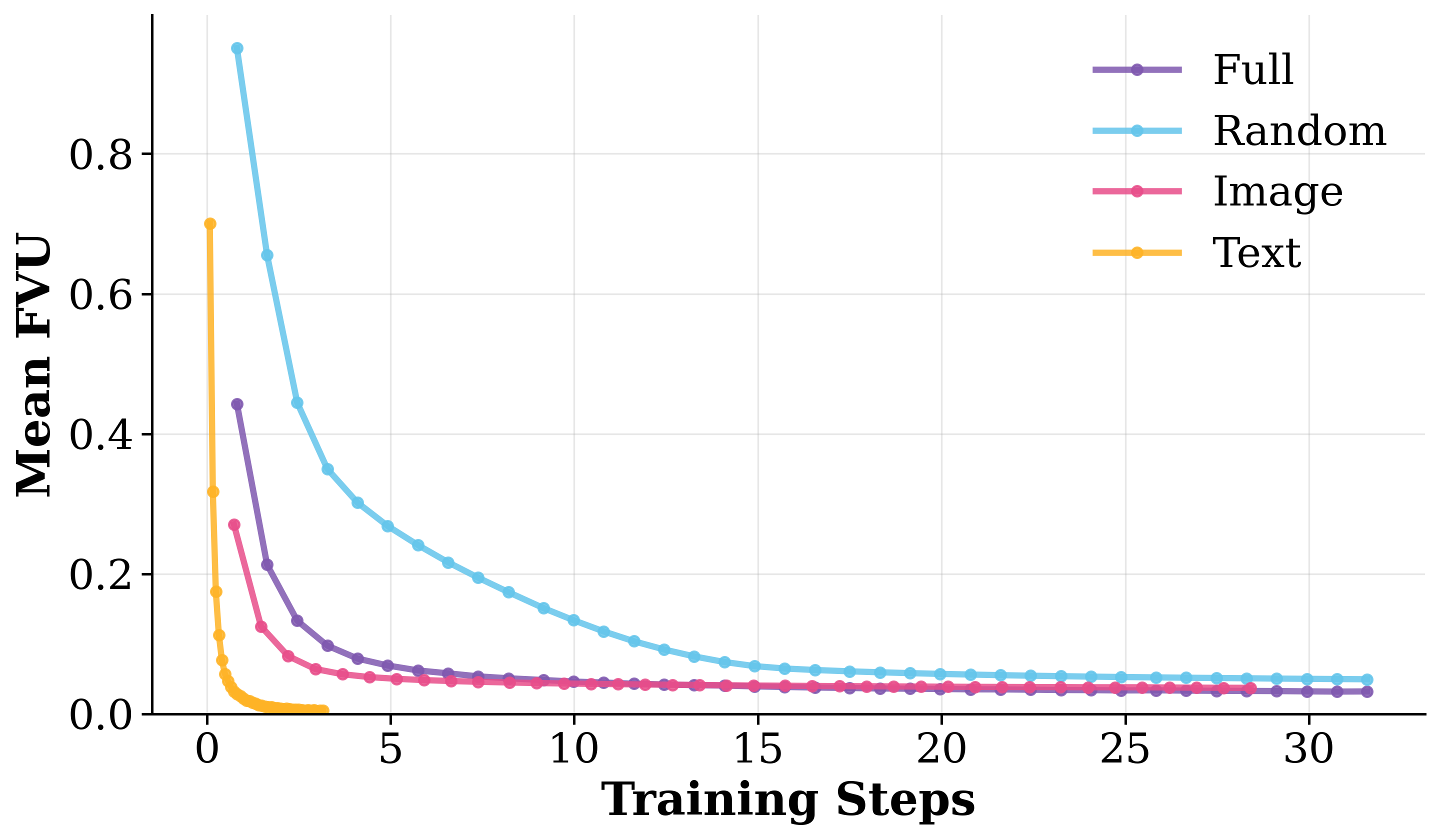}
  \end{minipage}\hfill
  \begin{minipage}[c]{0.42\linewidth}
    \vspace{0pt}
    \centering
    \scriptsize
    \setlength{\tabcolsep}{3pt}
    \renewcommand{\arraystretch}{0.92}
    \resizebox{\linewidth}{!}{%
      \begin{tabular}{lrrrr}
        \toprule
        \textbf{Metric} & \textbf{Full} & \textbf{Random} & \textbf{Image} & \textbf{Text} \\
        \midrule
        Mean        & 0.032 & 0.050 & 0.037 & \textbf{0.005} \\
        Std         & 0.028 & 0.041 & 0.027 & \textbf{0.009} \\
        Min         & 0.013 & 0.020 & 0.017 & \textbf{0.000} \\
        Max         & 0.123 & 0.198 & 0.123 & \textbf{0.037} \\
        Tokens (M)  & 31.6  & 31.6  & 28.4  & 3.2 \\
        \bottomrule
      \end{tabular}
    }
  \end{minipage}
  \caption{\textbf{SAE adaptation (MMDiff-Llama).} Left: mean FVU across layers; right: per-regime FVU summary. Text-only achieves lowest reconstruction; random init worst.}
  \label{fig:fvu-curves}
\end{figure}

\begin{figure}[h]
  \centering
  \includegraphics[width=\linewidth]{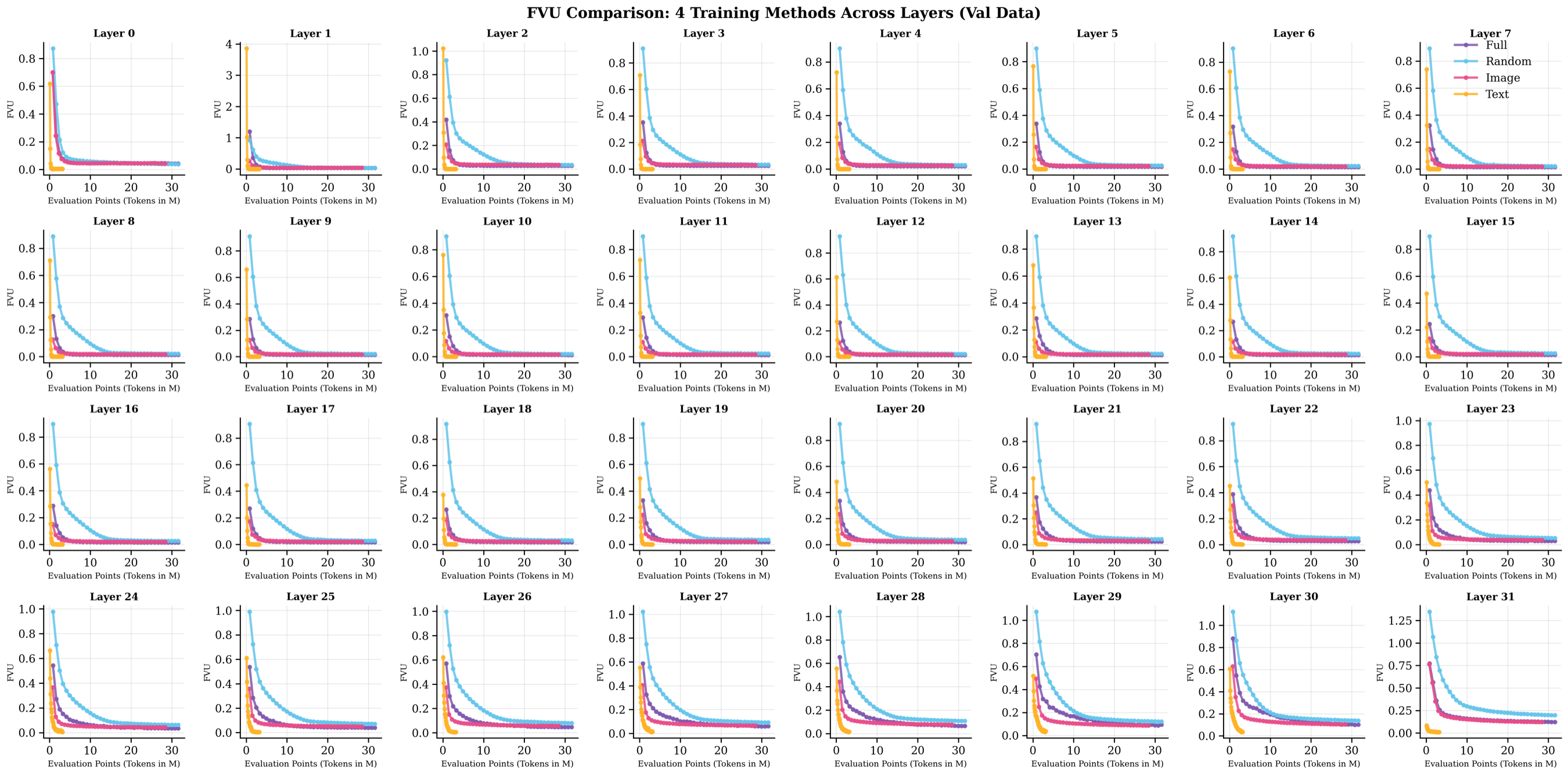}
  \caption{\textbf{Per-layer FVU across regimes (MMDiff-Llama).} Text-only SAEs converge to near-zero error rapidly; image and full-sequence regimes plateau higher.}
  \label{fig:fvu-layerwise}
\end{figure}

Figure~\ref{fig:fvu-curves} summarizes aggregated FVU trends for MMDiff-Llama, and Figure~\ref{fig:fvu-layerwise} reports the corresponding per-layer trajectories. Text-only SAEs converge rapidly and achieve the lowest reconstruction error across layers. Although text-only training uses fewer tokens than full-sequence or image-only training, it converges to lower FVU because those activations remain closer to the warm-started base-LM distribution. Image-only and full-sequence SAEs converge more slowly and plateau at higher FVU, consistent with a distributional mismatch between projected image tokens and the pretrained language-model activation basis. Random initialization performs worst, confirming that the pretrained base-LM dictionary provides a useful starting point for multimodal adaptation. 

\subsection{Decoder geometry across regimes}
\label{app:rotations}

\begin{figure}[h]
  \centering
  \includegraphics[width=0.8\linewidth]{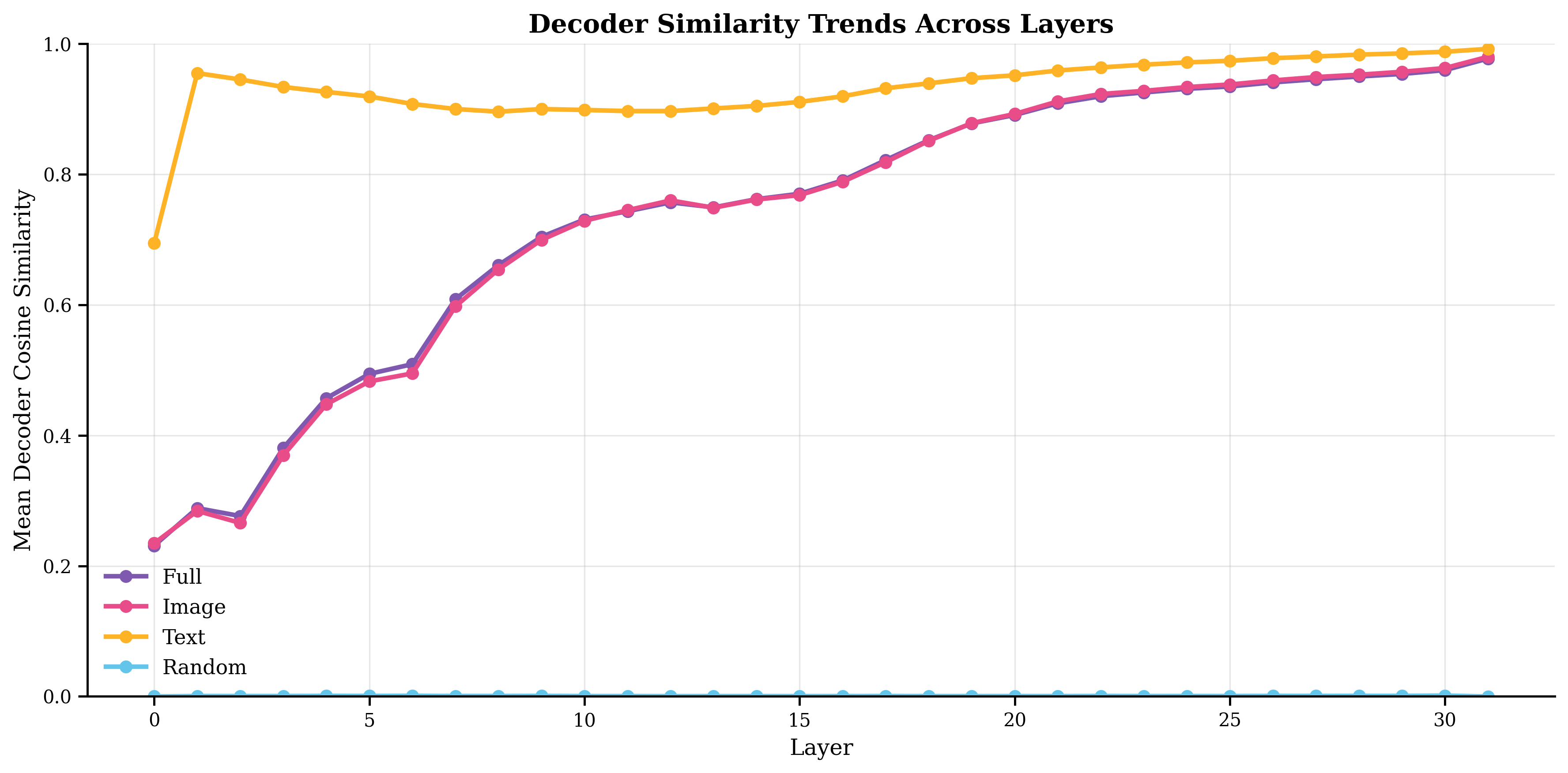}
  \caption{\textbf{Decoder cosine similarity vs.\ layer (MMDiff-Llama).} Text-only SAEs remain aligned with the base-LM dictionary across layers; image-only and full-sequence SAEs diverge in shallow layers and only re-align deeper. Random initialization stays decorrelated.}
  \label{fig:decoder-cos-trends}
\end{figure}

Stage-wise diffing assumes that adaptation produces feature-level changes rather than an arbitrary wholesale rotation of the representation. To quantify how SAE feature geometry shifts across training regimes, we track cosine similarity between decoder directions from SAEs trained on different input types. Figure~\ref{fig:decoder-cos-trends} shows that text-only SAEs remain closely aligned with the base-LM dictionary across layers, while image-only and full-sequence SAEs diverge in early layers before realigning deeper in the model. Randomly initialized SAEs stay largely uncorrelated, confirming the stability of the observed trends. We therefore use text-only SAEs as the main dictionary for model diffing, where feature identities are stable enough to compare base-LM and MLLM-adapted directions.

\subsection{Seed stability of the learned dictionary}
\label{app:seed-stability}

Since rotation is measured against same-index base features, we check that the multimodal dictionary is reproducible across training seeds rather than an artifact of one run. We retrained the PaliGemma~2 SAEs for all eight layers hosting our top spatial features with a different seed, varying only the seed and the data ordering, and matched the two runs feature by feature (Table~\ref{tab:seed-stability}). The mean same-index decoder cosine is $0.93$ with $84$\% of features at or above $0.9$, and the cross-seed relocation rate is $0.44$\% ($\approx$$72$ of $16{,}384$ features per layer), so a feature's best match stays at its own index in over $99.5$\% of cases. The top $10$ features we report are individually more stable than the dictionary average, with a mean same-index cosine of $0.98$.

\begin{table}[h]
  \centering
  \small
  \setlength{\abovecaptionskip}{7pt}
  \begin{tabular}{c>{\columncolor{tabAccent}}rrrrr}
    \toprule
    \rowcolor{tabHeader}
    Layer & Mean cos & Median cos & $\ge 0.9$ & Relocated & Count \\
    \midrule
    4  & $0.936$ & $0.961$ & $84.6$\% & $0.32$\% & $52$ \\
    6  & $0.937$ & $0.962$ & $84.3$\% & $0.34$\% & $56$ \\
    9  & $0.931$ & $0.960$ & $83.5$\% & $0.60$\% & $98$ \\
    11 & $0.933$ & $0.959$ & $83.6$\% & $0.43$\% & $71$ \\
    12 & $0.928$ & $0.955$ & $82.7$\% & $0.50$\% & $82$ \\
    13 & $0.931$ & $0.956$ & $83.9$\% & $0.41$\% & $67$ \\
    14 & $0.930$ & $0.955$ & $83.7$\% & $0.40$\% & $65$ \\
    15 & $0.931$ & $0.956$ & $84.1$\% & $0.54$\% & $89$ \\
    \bottomrule
  \end{tabular}
  \caption{\textbf{Cross-seed dictionary stability (PaliGemma~2).} Same-index decoder cosine between two SAE training runs differing only in seed and data order, per layer. ``Relocated'' is the fraction of features whose best cross-seed match is not at their own index, out of $16{,}384$ features per layer; ``Count'' gives the corresponding number of features.}
  \label{tab:seed-stability}
\end{table}

\section{Adapted Feature Selection Diagnostics}
\label{app:adapted-selection-scatter}

\subsection{Joint visual-energy and cosine distribution}
\label{app:joint-ev-cos}

\begin{figure*}[t]
  \centering
  \includegraphics[width=0.95\textwidth]{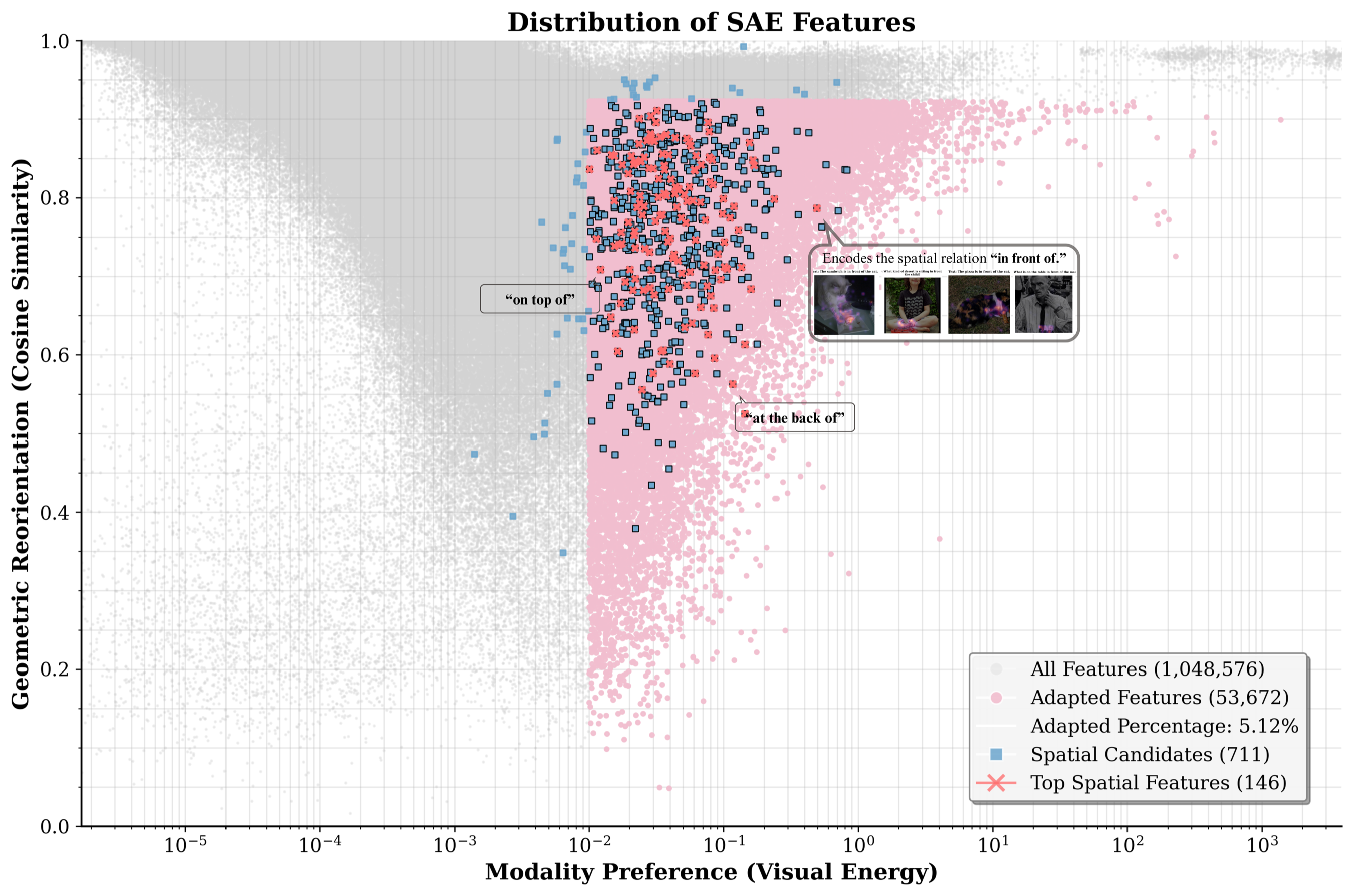}
  \caption{\textbf{Distribution of SAE features by visual energy and cosine similarity (MMDiff-Llama).} All features are shown in gray; adapted features are highlighted in pink. Task-specific candidates (here: spatial) are marked with blue squares, and the subset used for downstream analysis is shown as red crosses.}
  \label{fig:global-scatter}
\end{figure*}

Figure~\ref{fig:global-scatter} visualizes the joint distribution over $E_v$ and $c_f$ used in Sec.~\ref{sec:adapted} to define the adapted set $\mathcal{A}$. Adapted features form a compact pink envelope; spatial candidates and the subset used for downstream analysis sit inside it.

\begin{figure*}[t]
  \centering
  \includegraphics[width=0.7\textwidth]{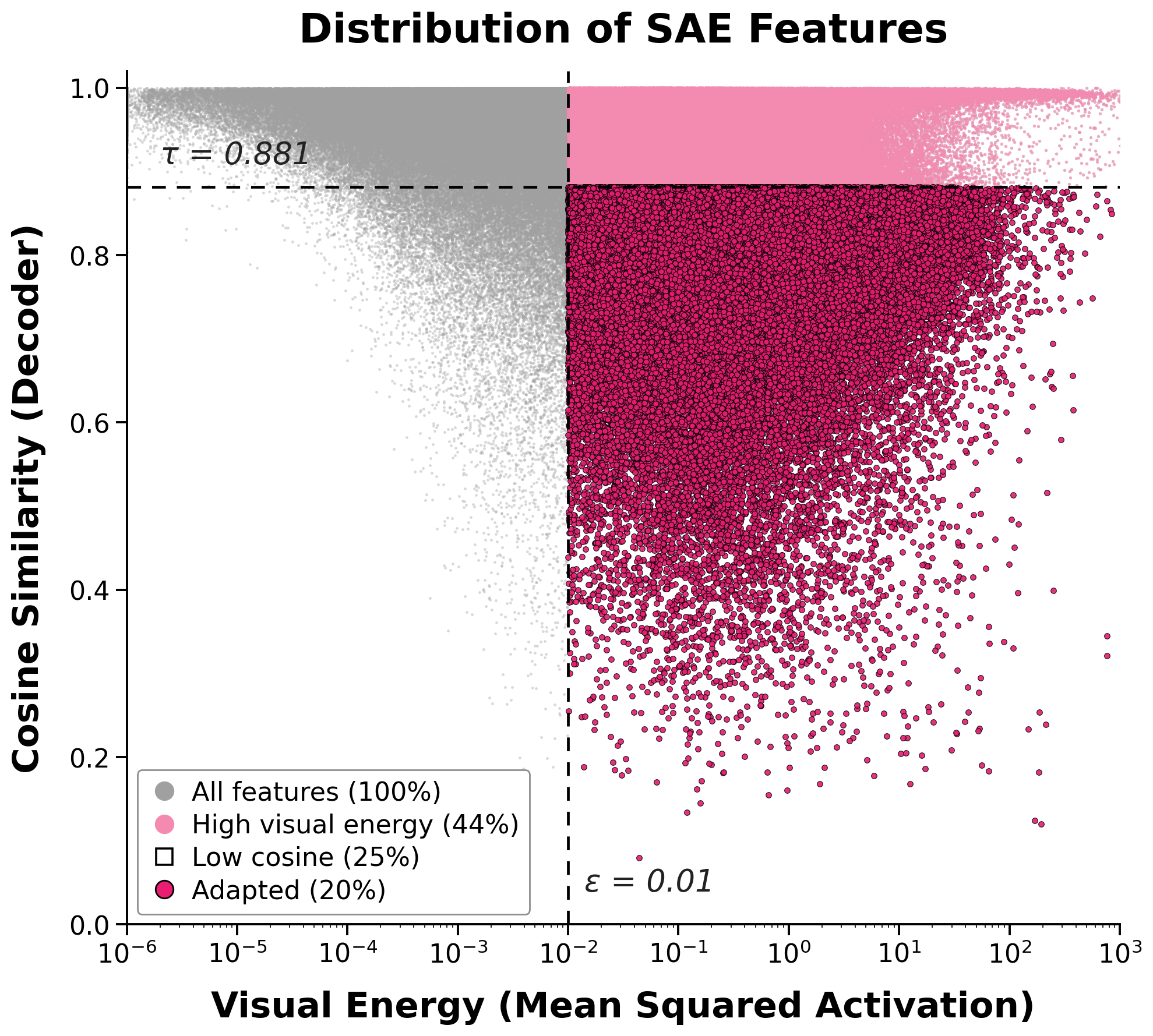}
  \caption{\textbf{Distribution of SAE features by visual energy and cosine similarity (MMDiff-Gemma).} All features in gray; high-visual-energy features ($E_v > \epsilon = 0.01$) in light pink; the adapted set $\mathcal{A}$ (high $E_v$ \emph{and} bottom-$25\%$ cosine, $\tau = 0.881$) in dark pink. Adapted features comprise $\sim$$20\%$ of MMDiff-Gemma's dictionary, larger than the $\sim$$5\%$ adapted set on MMDiff-Llama (Fig.~\ref{fig:global-scatter}).}
  \label{fig:global-scatter-pg2}
\end{figure*}

\subsection{Per-layer adapted-feature statistics}
\label{app:adapted-perlayer}

\begin{figure}[h]
  \centering
  \begin{subfigure}[t]{0.48\linewidth}
    \centering
    \includegraphics[width=\linewidth]{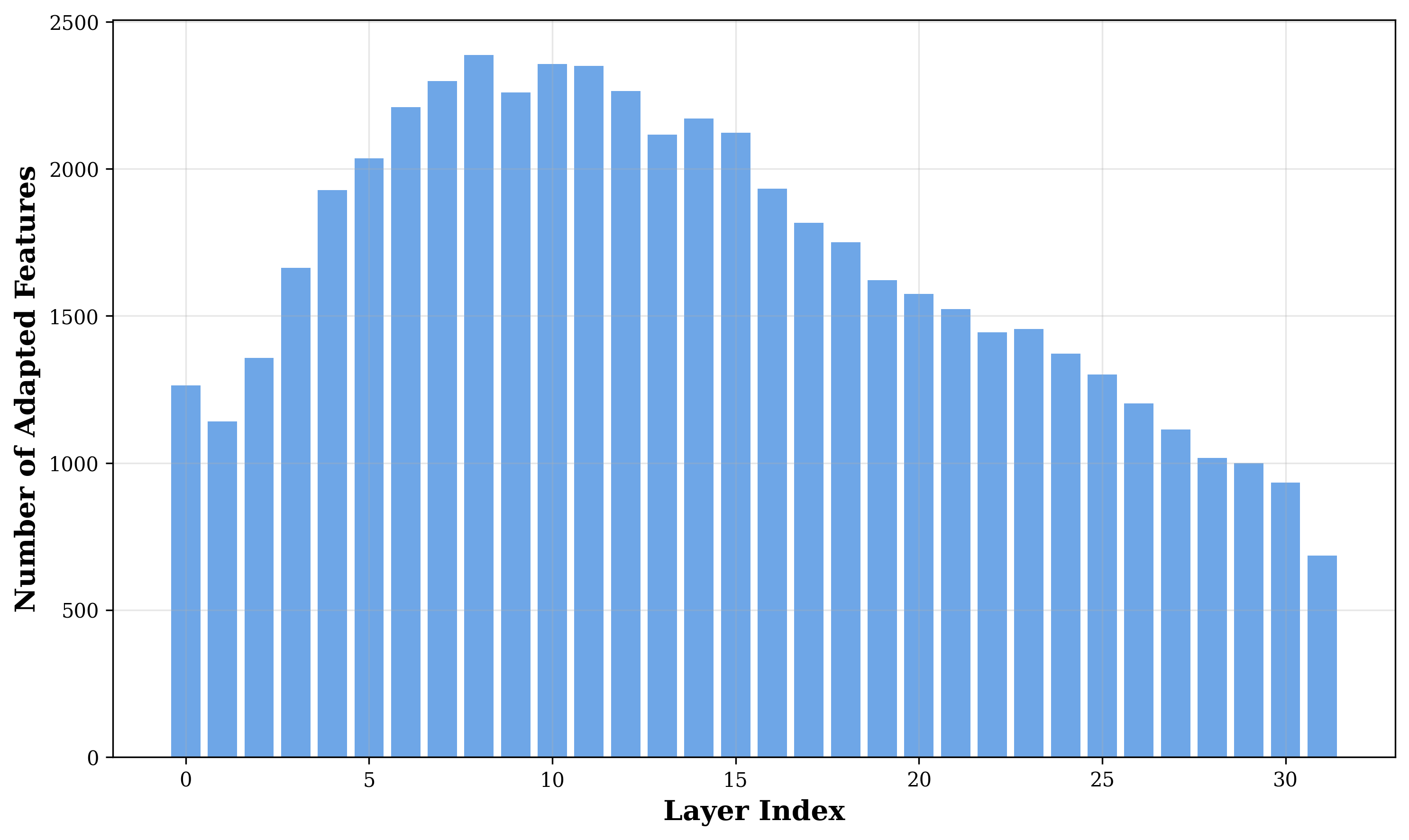}
    \caption{Adapted features per layer. Most concentrate in mid layers, tapering in deeper blocks.}
    \label{fig:suspects-per-layer}
  \end{subfigure}\hfill
  \begin{subfigure}[t]{0.48\linewidth}
    \centering
    \includegraphics[width=\linewidth]{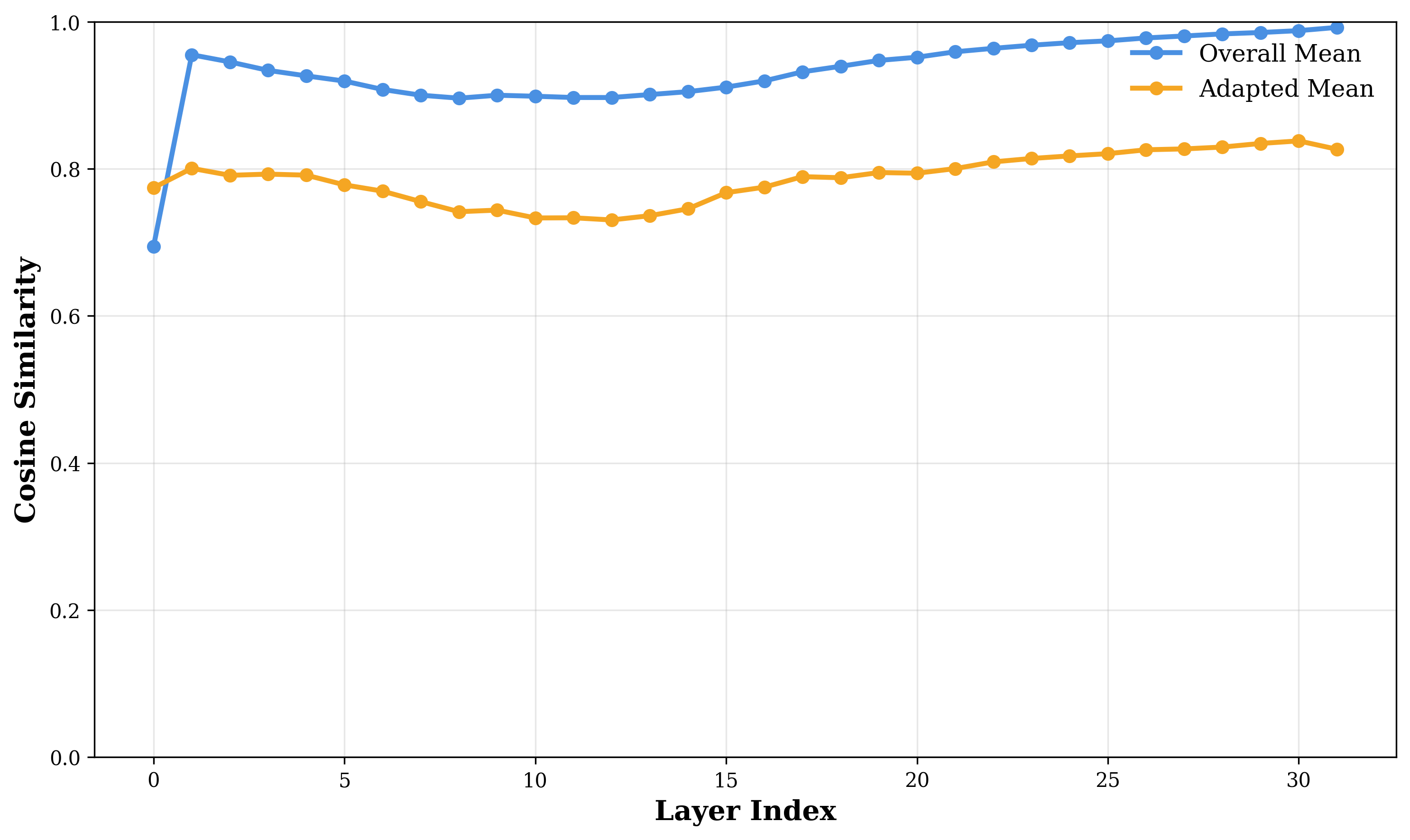}
    \caption{Decoder cosine by layer. Adapted features remain less aligned to the base dictionary than the overall pool.}
    \label{fig:cosine-overall-vs-suspect}
  \end{subfigure}
  \caption{\textbf{Per-layer adapted-feature stats (MMDiff-Llama).} Counts and mean cosine for the adapted set $\mathcal{A}$ (Sec.~\ref{sec:adapted}).}
\end{figure}

\subsection{Threshold sweep for feature selection}
\label{app:threshold-sweep}

\begin{figure}[h]
  \centering
  \includegraphics[width=\linewidth]{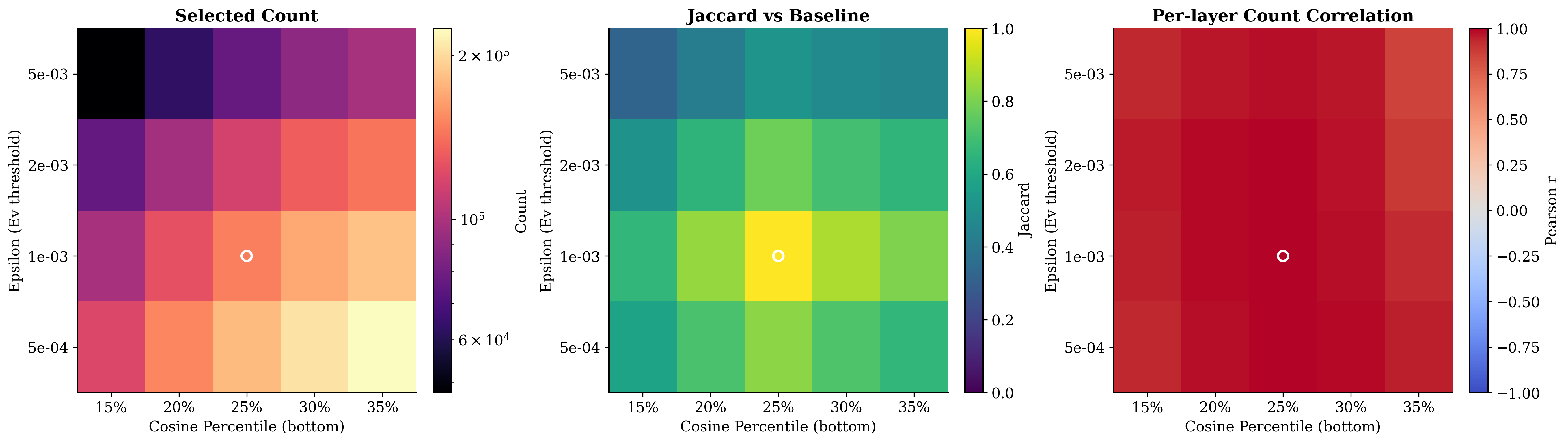}
  \caption{\textbf{Threshold sweep.} Adapted-set size as a function of $\epsilon$ and $p_{\cos}$.}
  \label{fig:threshold-sweep}
\end{figure}

Figure~\ref{fig:threshold-sweep} shows how the size of the adapted set varies with the visual-energy cutoff $\epsilon$ and the cosine percentile $p_{\cos}$. The selection used in Sec.~\ref{sec:adapted} ($p_{\cos}=25\%$) is highlighted; downstream results are stable across reasonable choices.

\subsection{Filtering funnel for task-specific feature discovery}
\label{app:filtering-funnel}

For each feature $f$, the odds ratio in Sec.~\ref{sec:taskspec} is computed from the firing/non-firing $\times$ target/baseline contingency table:
\[
\mathrm{OR}_f
=
\frac{
n^{\mathrm{tgt}}_{\mathrm{fire}}
\cdot
n^{\mathrm{base}}_{\mathrm{nonfire}}
}{
n^{\mathrm{base}}_{\mathrm{fire}}
\cdot
n^{\mathrm{tgt}}_{\mathrm{nonfire}}
},
\]
where $n_{\mathrm{fire}}$ and $n_{\mathrm{nonfire}}$ denote firing and non-firing token counts for feature $f$ in the target and baseline distributions.

Table~\ref{tab:filtering-funnel} summarizes the per-stage filtering counts referenced in Sec.~\ref{sec:taskspec} across MMDiff models and target distributions. The funnel shrinks the full SAE dictionary (size $L \cdot F_{\text{width}}$) to a compact, task-specific feature set in three stages: (i)~the adapted set $\mathcal{A}$ defined by the visual-energy and decoder-cosine criteria of Sec.~\ref{sec:adapted}; (ii)~distribution-shift candidates within $\mathcal{A}$ that pass the Fisher exact test with $\mathrm{OR}_f \ge 3$ and $\Delta p_f \ge 0.05$ (Sec.~\ref{sec:taskspec}, step~1); and (iii)~the final task-specific set after lexical-invariance filtering (Sec.~\ref{sec:taskspec}, step~2). The same three-stage funnel is applied uniformly across spatial, safety, and OCR target distributions; only $\mathcal{D}_{\text{tgt}}$ and the neutral-prompt bank change between applications.

\begin{table}[h]
  \centering
  \footnotesize
  \setlength{\tabcolsep}{3pt}
  \renewcommand{\arraystretch}{0.92}
  \begin{tabular}{llrrr}
    \toprule
    Model & Target & All features & $\mathcal{A}$ (adapted) & Discovered \\
    \midrule
    MMDiff-Llama & Spatial         & $\sim$$1{,}024$K & $\sim$$51$K (5\%) & $711$ \\
    MMDiff-Gemma & Spatial         & $\sim$$416$K & $\sim$$85$K ($\sim$$20$\%) & $\sim$$1{,}400$ \\
    MMDiff-Gemma & Safety  & $\sim$$416$K & $\sim$$85$K ($\sim$$20$\%) & $1{,}061$ \\
    MMDiff-Gemma & OCR             & $\sim$$416$K & $\sim$$85$K ($\sim$$20$\%) & $1{,}070$ \\
    MMDiff-Qwen  & Spatial         & $\sim$$918$K & $\sim$$123$K ($\sim$$13$\%) & $\sim$$2{,}800^{\dagger}$ \\
    \bottomrule
  \end{tabular}
  \caption{\textbf{Filtering funnel across MMDiff models and target distributions.} Counts at each stage of the task-specific discovery pipeline (Sec.~\ref{sec:taskspec}). Total feature counts = $L \cdot F_{\text{width}}$ (Tab.~\ref{tab:mmdiff-saes}). \emph{Discovered} = features after contrastive-firing screen and lexical-invariance filter; for safety this is the candidate sweep size used in Sec.~\ref{sec:safety}. $^{\dagger}$For MMDiff-Qwen the contrastive-firing stage yields $28{,}955$ candidates within $\mathcal{A}$; the lexical-invariance filter was run on a $400$-feature sample of these, with a pass rate that is stable across the odds-ratio range. The reported count extrapolates that rate to the full candidate set.}
  \label{tab:filtering-funnel}
\end{table}

\subsection{Distribution-shift visualizations}
\label{app:dist-shift}

\begin{figure}[h]
  \centering
  \includegraphics[width=0.5\linewidth]{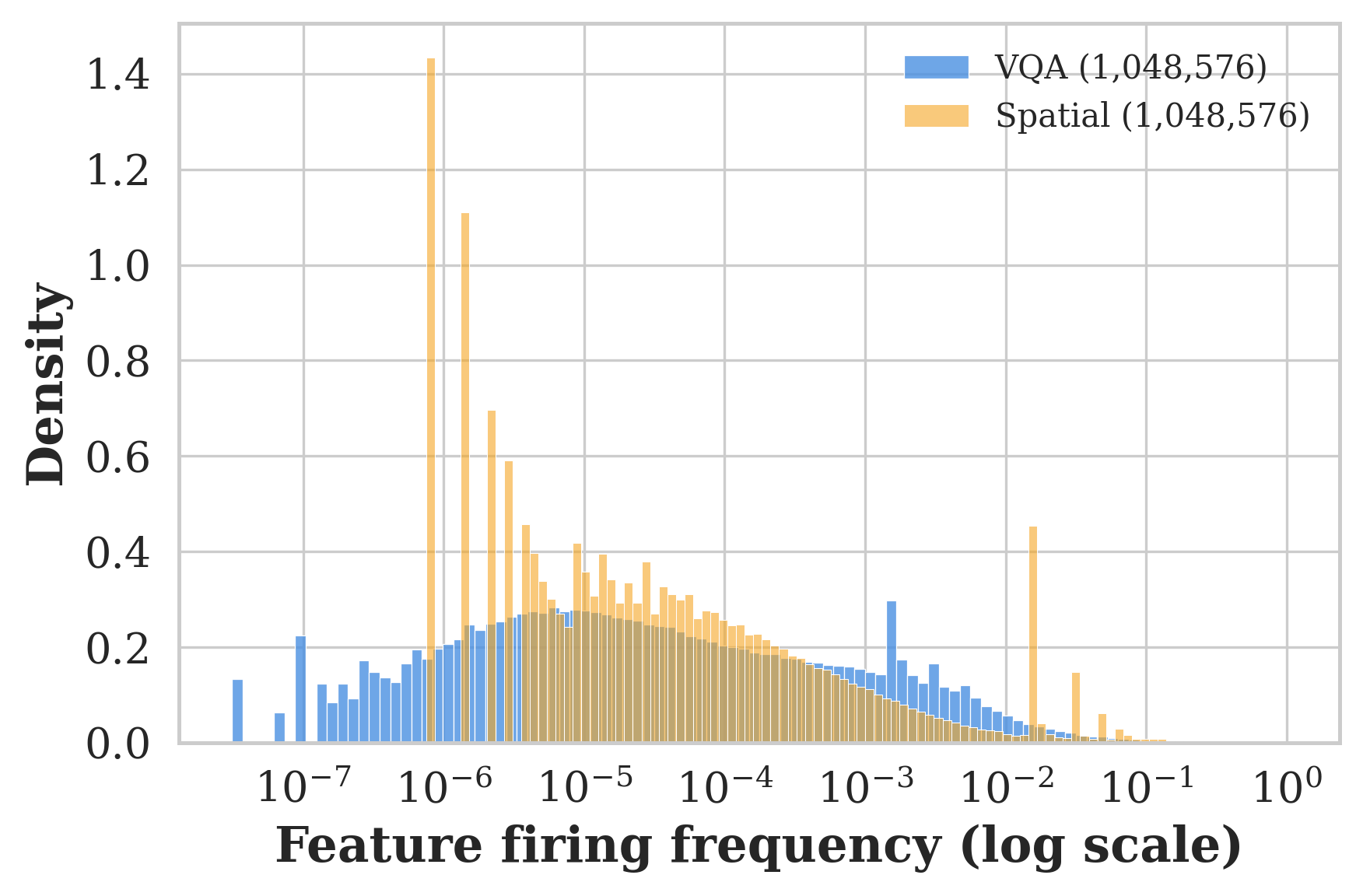}
  \caption{\textbf{Firing-frequency distributions for $\mathcal{D}_{\text{base}}$ and $\mathcal{D}_{\text{sp}}$.}}
  \label{fig:firing-hist}
\end{figure}

Figure~\ref{fig:firing-hist} shows the per-feature firing-frequency distributions for the spatial split $\mathcal{D}_{\text{sp}}$ and the baseline $\mathcal{D}_{\text{base}}$ used in Sec.~\ref{sec:taskspec}. Analogous histograms for the safety target distribution and (when results land) for the OCR target distribution are reported in supplementary panels.

\subsection{Lexical-invariance prompt banks}
\label{app:lexical-prompts}

The lexical-invariance filter (Sec.~\ref{sec:taskspec}) replaces the original question on each top-activating sample with a small bank of generic prompts and keeps a feature only if it continues to fire on those generic prompts. We use three target-distribution-specific banks, listed verbatim from our pipeline:

\paragraph{Spatial.}
\begin{enumerate}
  \item \emph{Describe how the items are arranged.}
  \item \emph{Comment on the overall layout and organization of the scene.}
  \item \emph{Summarize the structure in terms of grouping or separation.}
  \item \emph{Explain the relative positioning of objects without naming directions.}
  \item \emph{Describe patterns of arrangement, such as order or symmetry.}
\end{enumerate}
The spatial bank avoids any explicit spatial vocabulary (\emph{left/right/above/below}, etc.) so that surviving features cannot be firing on the spatial \emph{lexicon} of the prompt; they must be firing on the spatial \emph{content} of the image.

\paragraph{OCR and Multimodal Safety.}
\begin{enumerate}
  \item \emph{Describe this image.}
  \item \emph{What do you see in this picture?}
  \item \emph{Summarize the contents of the image.}
  \item \emph{Describe the objects and scene in this image.}
  \item \emph{What is happening in this image?}
\end{enumerate}
For OCR and safety, the bank is generic image-description prompts (no OCR-specific cue like ``what does the sign say'', no safety-specific cue). Surviving features must fire on visual content alone.

A feature passes the lexical filter if it remains active above the activation cutoff $\eta = 0.01$ on at least one of the bank's prompts (per top-$k$ activating sample, $k=5$).

\subsection{Image counterfactuals}
\label{app:image-counterfactuals}

$E_v$ alone does not separate image-driven from text-driven activation. To disentangle the two we hold the text fixed and replace the image, on every feature reported in the paper and $300$ evaluation samples per feature. \emph{Shuffled} substitutes another image drawn from the same evaluation set, \emph{blank} a solid grey image of matched dimensions, and \emph{noise} an image whose every pixel is drawn uniformly at random. Table~\ref{tab:image-cf} reports the change in each feature's mean activation relative to the correct image, averaged within each domain.

Every domain loses activation under every corruption, so these features depend on image content rather than on the prompt alone. Blanking the image is the most disruptive intervention throughout, and OCR features are the most image-dependent, losing $36.8$\% of their activation when the image carries no text to read. Substitution is the mildest intervention in all three domains, which is expected: a replacement image drawn from the same distribution still contains some of the property the feature tracks.

\begin{table}[h]
  \centering
  \small
  \setlength{\abovecaptionskip}{7pt}
  \begin{tabular}{l>{\columncolor{tabAccent}}rrr}
    \toprule
    \rowcolor{tabHeader}
    Task features & Shuffled & Blank & Noise \\
    \midrule
    Spatial & $-7.2$\% & $-8.3$\% & $-13.2$\% \\
    Safety & $-4.2$\% & $-21.8$\% & $-8.6$\% \\
    OCR & $-16.6$\% & $\mathbf{-36.8}$\% & $-28.7$\% \\
    \bottomrule
  \end{tabular}
  \caption{\textbf{Image counterfactuals (PaliGemma~2).} Change in mean feature activation relative to the correct image, with the text held fixed ($300$ samples per feature). A value of $0$\% would mean the image made no difference.}
  \label{tab:image-cf}
\end{table}

\subsection{Standard SAE trained directly on MLLM activations}
\label{app:scratch-sae}

Sec.~\ref{sec:diffing-ablation} reports that a randomly initialised SAE trained directly on MLLM activations yields no causally effective features. Table~\ref{tab:scratch-sae} gives the per-feature values for both dictionaries on \textbf{LLaVA-MORE}, under the identical three-point all-layers projection protocol. For the from-scratch dictionary the contrastive odds ratio saturates: every candidate in the top group fires on all VSR samples, so the ranking cannot separate task-specific features from always-on ones, and the ten reported here are drawn from that tied group.

\begin{table}[h]
  \centering
  \small
  \setlength{\abovecaptionskip}{7pt}
  \begin{tabular}{l>{\columncolor{tabAccent}}rr@{\hskip 2em}l>{\columncolor{tabAccent}}rr}
    \toprule
    \rowcolor{tabHeader}
    Standard SAE & $\Delta$VSR & $\Delta$VQA & MMDiff & $\Delta$VSR & $\Delta$VQA \\
    \midrule
    L5\_F7871 & $-1.10$ & $+0.20$ & L7\_F15870  & $-15.54$ & $-0.10$ \\
    L5\_F2220 & $-0.57$ & $+0.10$ & L11\_F27061 & $-13.30$ & $-0.40$ \\
    L5\_F1591 & $-0.12$ & $\phantom{-}0.00$ & L9\_F15404  & $-11.19$ & $-0.80$ \\
    L5\_F2267 & $\phantom{-}0.00$ & $\phantom{-}0.00$ & L7\_F6986   & $-10.87$ & $-0.50$ \\
    L5\_F8323 & $\phantom{-}0.00$ & $+0.20$ & L12\_F23874 & $-10.24$ & $-0.40$ \\
    L5\_F2079 & $+0.38$ & $+0.20$ & L14\_F17873 & $-10.00$ & $-0.30$ \\
    L5\_F8517 & $+0.38$ & $\phantom{-}0.00$ & L18\_F29948 & $\phantom{-}-7.98$ & $-0.30$ \\
    L5\_F4652 & $+0.87$ & $-0.20$ & L10\_F5121  & $\phantom{-}-7.92$ & $-0.10$ \\
    L5\_F3277 & $+1.00$ & $+0.30$ & L11\_F24089 & $\phantom{-}-7.68$ & $-0.60$ \\
    L5\_F4537 & $+1.38$ & $-0.10$ & L12\_F13305 & $\phantom{-}-6.38$ & $-0.70$ \\
    \midrule
    \textbf{Mean} & $\mathbf{+0.22}$ & $\mathbf{+0.07}$ & \textbf{Mean} & $\mathbf{-10.11}$ & $\mathbf{-0.42}$ \\
    \bottomrule
  \end{tabular}
  \caption{\textbf{From-scratch SAE vs MMDiff on LLaVA-MORE.} Left: top spatial features selected by contrastive firing over a randomly initialised dictionary trained directly on MLLM activations. Right: MMDiff features on the same model. Both ablated under the identical protocol.}
  \label{tab:scratch-sae}
\end{table}

\subsection{Randomly-selected feature ablation}
\label{app:random-baseline}

The selection ablation in Sec.~\ref{sec:diffing-ablation} reports a randomly-selected feature baseline to establish that the causal effects come from the selected directions rather than from the projection intervention itself. Table~\ref{tab:random-baseline} gives the per-feature values. We sample one random feature per layer from the same eight layers that host our top spatial features, ablate it under the identical three-point all-layers protocol, and evaluate on the same relation subsets. All ten have odds ratio $1.0$ by construction, confirming they carry no contrastive-firing signal.

\begin{table}[h]
  \centering
  \small
  \setlength{\abovecaptionskip}{7pt}
  \begin{tabular}{ccr>{\columncolor{tabAccent}}rr}
    \toprule
    \rowcolor{tabHeader}
    Layer & Feature & OR & $\Delta$VSR & $\Delta$VQA \\
    \midrule
    4  & 5043  & $1.0$ & $\phantom{-}0.00$ & $-0.20$ \\
    6  & 671   & $1.0$ & $-0.31$ & $-0.10$ \\
    9  & 4420  & $1.0$ & $\phantom{-}0.00$ & $\phantom{-}0.00$ \\
    9  & 13936 & $1.0$ & $+1.04$ & $-0.20$ \\
    11 & 1232  & $1.0$ & $-0.54$ & $-0.60$ \\
    11 & 8374  & $1.0$ & $-1.56$ & $-0.70$ \\
    12 & 13324 & $1.0$ & $-1.96$ & $\phantom{-}0.00$ \\
    13 & 270   & $1.0$ & $-0.28$ & $-0.10$ \\
    14 & 10435 & $1.0$ & $\phantom{-}0.00$ & $+0.20$ \\
    15 & 2871  & $1.0$ & $-0.99$ & $-0.40$ \\
    \midrule
    \multicolumn{3}{l}{\textbf{Mean}} & $\mathbf{-0.46}$ & $\mathbf{-0.21}$ \\
    \bottomrule
  \end{tabular}
  \caption{\textbf{Randomly-selected feature ablation (PaliGemma~2).} One random feature per layer, drawn from the same layers as the top spatial features and ablated under the identical protocol. OR $= 1.0$ confirms no contrastive-firing selection.}
  \label{tab:random-baseline}
\end{table}

\section{Steering Decomposition and Feature Correspondence}
\label{app:steer-matching}

\subsection{Decomposing the steering gains}
\label{app:steer-decomp}

MMDiff~CAA combines two mechanisms, multi-layer CAA at the discovered feature layers and injection of the feature's decoder direction, so we evaluate each in isolation on the same top $10$ spatial features (Table~\ref{tab:steer-decomp}). SAE feature steering improves VSR by $2.63$ with a single feature and $3.66$ with all ten, well below either CAA variant, so the direction is not sufficient on its own. Single-layer CAA gives $8.96$. Multi-layer CAA adds $1.82$, and injecting the feature's decoder direction adds a further $1.81$, giving $12.59$. Features with stronger isolated directions produce larger steering effects, on both VSR and OCR.

\subsection{Feature correspondence across dictionaries}
\label{app:matching}

Index alignment between the base-LM and multimodal dictionaries underpins the rotation measure, so we verify that features do not permute between the two dictionaries. For every feature in the full PaliGemma~2 dictionary ($26$ layers $\times$ $16{,}384$ features) we compute $c_{\text{same}}$, the cosine to its same-index base feature, and $c_{\max}$, the cosine to its best match anywhere in the base dictionary, and classify accordingly (Table~\ref{tab:matching}). Relocation is $0.00$\% at every layer, so warm-started features do not permute and index alignment is a valid correspondence; splitting and merging would surface as relocation or emergence, and both are near absent. The features we select are the rotated-in-place tail: $9$ of our top $10$ spatial features fall in the bottom quartile of their layer's rotation distribution. The correspondence is also stable across training seeds, with a cross-seed relocation rate of $0.44$\% and a mean same-index decoder cosine of $0.93$ (App.~\ref{app:seed-stability}).

\begin{table}[h]
  \centering
  \small
  \setlength{\abovecaptionskip}{7pt}
  \begin{minipage}[t]{0.53\linewidth}
    \centering
    \begin{tabular}{l>{\columncolor{tabAccent}}r}
      \toprule
      \rowcolor{tabHeader}
      Steering variant & $\Delta$VSR \\
      \midrule
      Single SAE feature direction & $+2.63$ \\
      All ten feature directions & $+3.66$ \\
      Single-layer CAA (vanilla) & $+8.96$ \\
      Multi-layer CAA & $+10.78$ \\
      \textbf{MMDiff~CAA} & $\mathbf{+12.59}$ \\
      \bottomrule
    \end{tabular}
    \captionof{table}{\textbf{Steering decomposition (PaliGemma~2 base).} Each component of MMDiff~CAA in isolation on the top $10$ spatial features.}
    \label{tab:steer-decomp}
  \end{minipage}
  \hfill
  \begin{minipage}[t]{0.44\linewidth}
    \centering
    \begin{tabular}{l>{\columncolor{tabAccent}}rr}
      \toprule
      \rowcolor{tabHeader}
      Feature type & Prop. & Count \\
      \midrule
      Preserved & $96.41$\% & $410{,}709$ \\
      Rotated in place & $3.36$\% & $14{,}295$ \\
      Relocated & $0.00$\% & $0$ \\
      Emergent & $0.23$\% & $971$ \\
      \bottomrule
    \end{tabular}
    \captionof{table}{\textbf{Explicit matching} over the full PaliGemma~2 dictionary.}
    \label{tab:matching}
  \end{minipage}
\end{table}

\section{Auto-Interpretation}
\label{app:auto-interp}

\subsection{Auto-Interpretation: Examples}
\label{app:auto-interp-examples}

\begin{figure}[h]
  \centering
  \includegraphics[width=\linewidth]{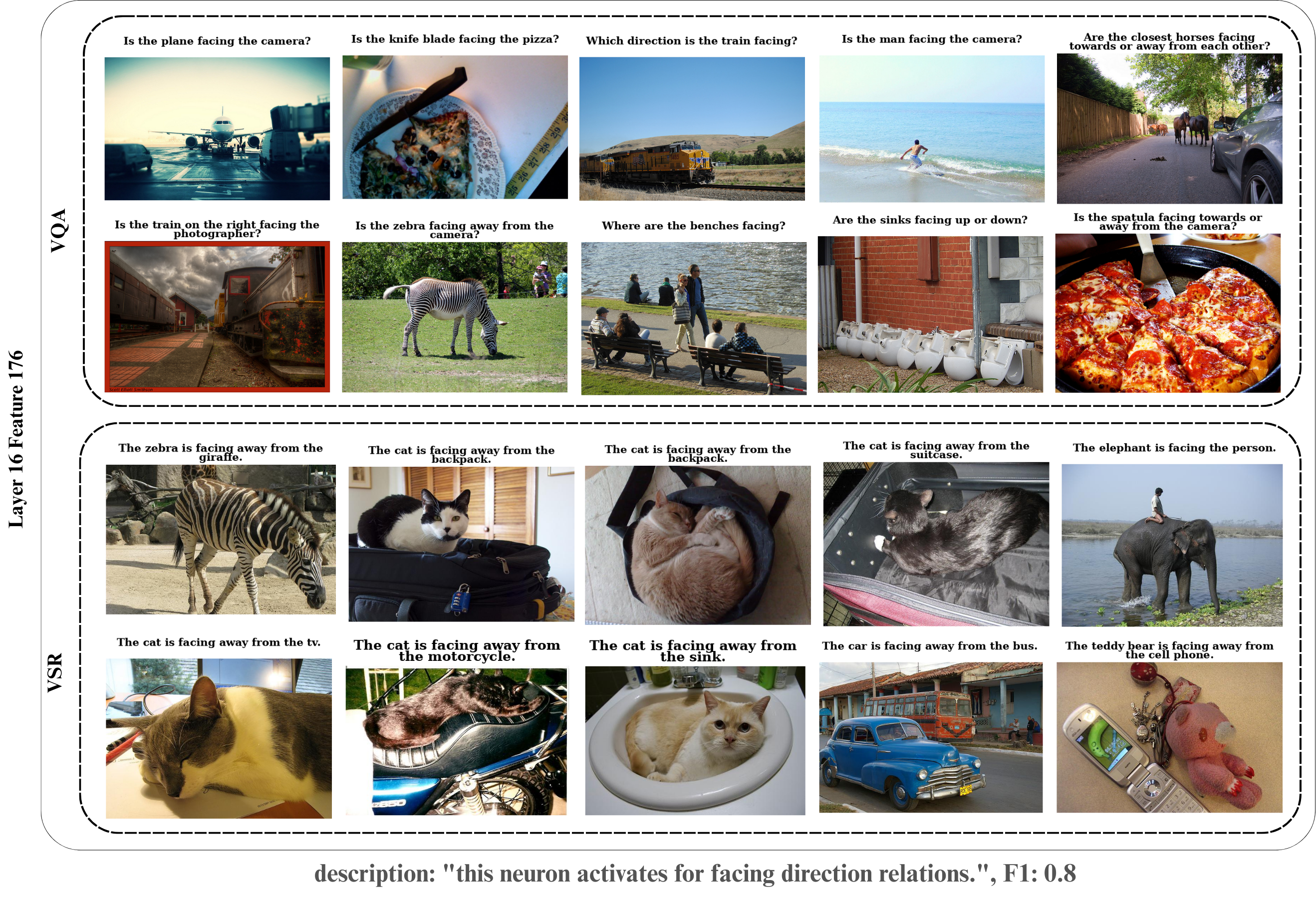}
  \caption{\textbf{Auto-Interp example (L16/F176, MMDiff-Llama).} Top VQA + VSR samples highlight \emph{facing direction}.}
  \label{fig:auto-interp-example-appendix}
\end{figure}

For each MMDiff-discovered feature we collect its top-activating samples from VQAv2 and VSR~\citep{liu2023vsr} and pass them to GPT-4o-mini~\citep{gpt-4o-mini}, which proposes a one-sentence description and an F1-based confidence score from a held-out classification task. Outputs are stored alongside the contrastive-firing metrics from Sec.~\ref{sec:taskspec} and lightly reviewed by hand; auto-interpretation is used as a qualitative validation layer rather than as a primary contribution. Figure~\ref{fig:auto-interp-example-appendix} shows a representative example (Layer 16, Feature 176, MMDiff-Llama: features sensitive to \emph{facing direction}). Two further examples follow; in both cases, the top-activating samples agree across VQA and VSR, and the interpretations are consistent and monosemantic.

\begin{figure*}[tbp]
  \centering
  \begin{subfigure}[t]{0.9\linewidth}
    \centering
    \includegraphics[width=\linewidth]{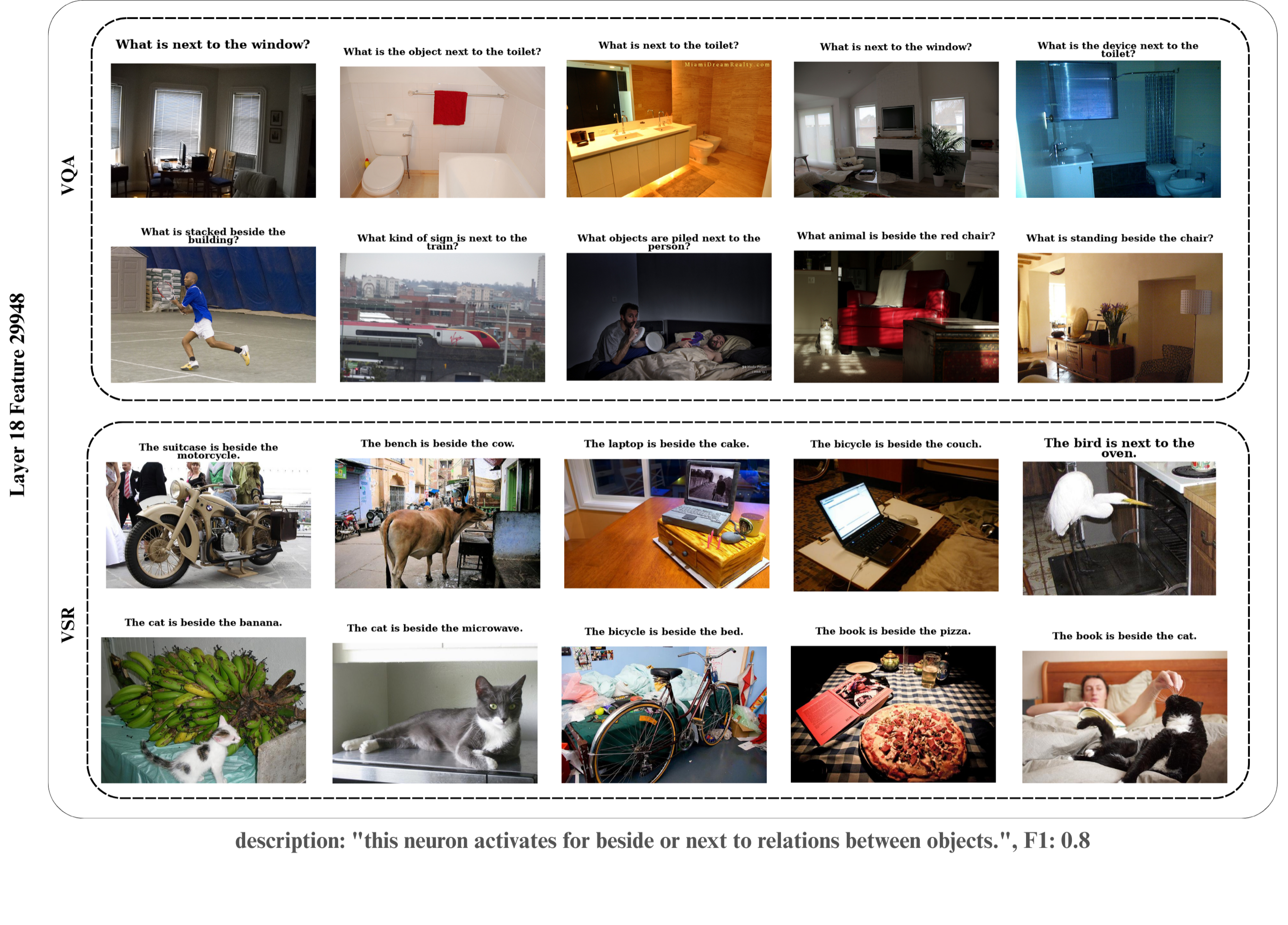}
    \label{fig:auto-interp-f29948}
  \end{subfigure}

  \begin{subfigure}[t]{0.9\linewidth}
    \centering
    \includegraphics[width=\linewidth]{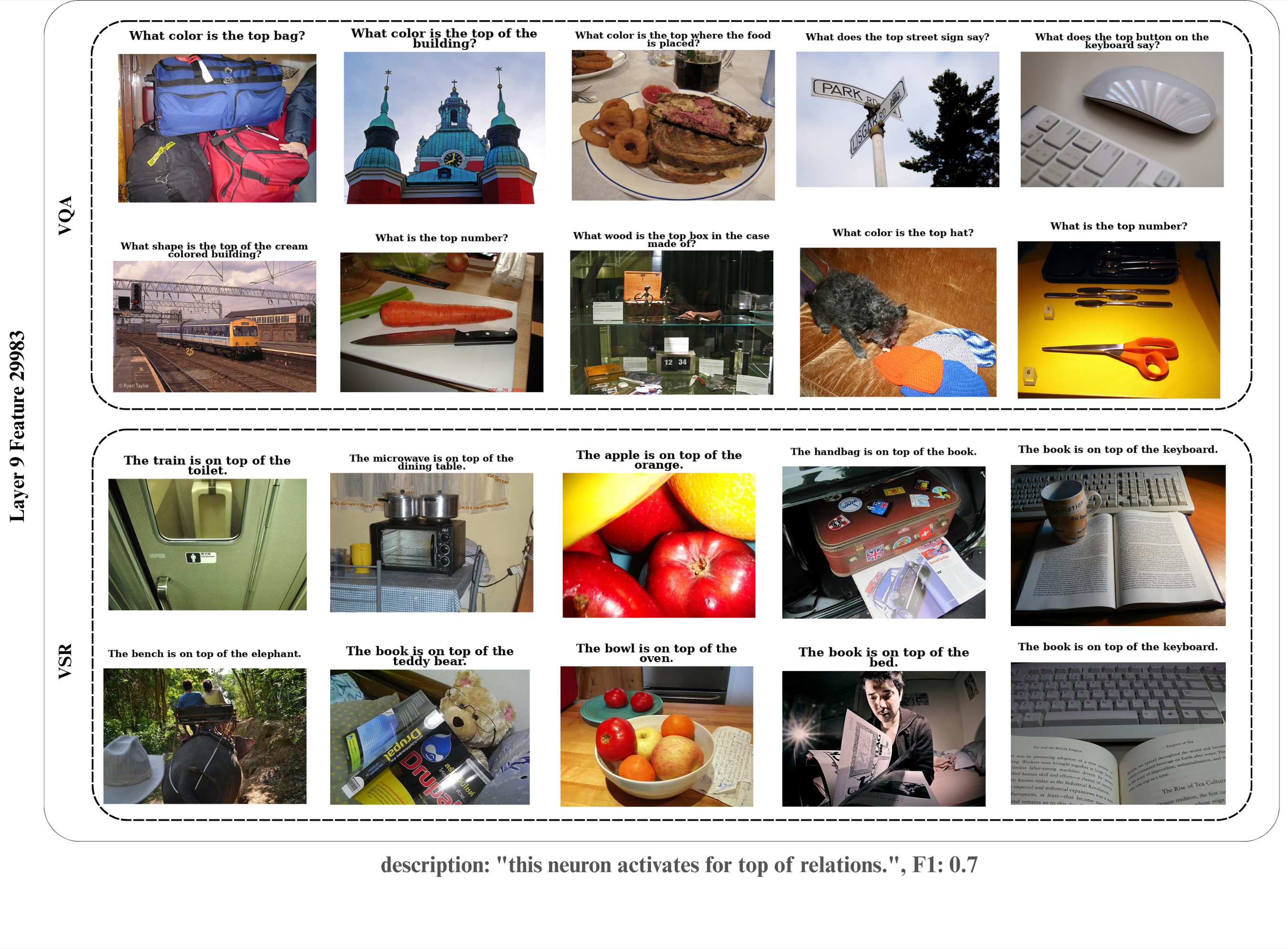}
    \label{fig:auto-interp-f29983}
  \end{subfigure}
  \caption{\textbf{Additional Auto-Interp examples.} Two adapted features; top VQA + VSR samples show consistent spatial relations.}
  \label{fig:appendix-auto-interp}
\end{figure*}

\subsection{Auto-Interpretation and Scoring Pipeline}
\label{app:auto-interp-pipeline}

We evaluate interpretability using an automated feature-description pipeline with two variants: \emph{RAW} (image+text) and \emph{OVERLAY} (image+text+top-head heatmaps). For each feature $f$:
\begin{enumerate}
  \item Select up to $k{=}5$ top-activating samples (deduped across VQA / VQA-spatial / VSR).
  \item Call the API once to generate a single concise description.
  \item Validate using held-out positive samples and random VQA negatives (two short rounds).
  \item Compute F1 as a lightweight proxy for description confidence.
\end{enumerate}
Outputs are stored per feature as JSON (\texttt{description}, examples, classification results). Adding overlays improves interpretability, with early results showing a typical gain of about $+0.2$ F1.

\paragraph{Prompt A: Description (RAW / OVERLAY).}
\textbf{System.} You are analyzing individual neurons using their top-activating samples (image$+$text; OVERLAY also includes attention heatmaps).\quad
\textbf{Task.} Produce one short, lower-case sentence completing: ``this neuron activates for \ldots''.\quad
\textbf{Guidelines.} Base it on consistent patterns supported by image ($+$overlays) and text; be specific; no hedging.\quad
\textbf{Return.} \texttt{\{"description": "one concise sentence"\}}.

\paragraph{Prompt B: Validation (F1).}
\textbf{System.} You are validating a neuron description against short examples (image$+$text; OVERLAY adds heatmaps).\quad
\textbf{Task.} For each sample, output $1$ if it reasonably matches the description; else $0$.\quad
\textbf{Return.} \texttt{\{"classifications": [0/1, \dots]\}}.

\section{Attribution Patching: Aggregated and Per-Feature Panels}
\label{app:attribution}

Attribution patching~\citep{nanda2023attribution} is an efficient alternative to activation patching~\citep{zhang2024activationpatching}, replacing repeated forward passes by a gradient-based linear approximation that estimates interventions with two forward and one backward pass. We adapt it to identify which attention heads drive a task-specific SAE feature $f$ at layer $L$: we read out the SAE decoder direction $v_f$ at layer $L$ to define a scalar objective, and use gradients with respect to upstream residuals and attention inputs to score each head's contribution. The clean run uses original image--text input; the corrupt run replaces layer-$0$ visual token embeddings with a mean embedding over many VQA samples. The two variants are $(\text{corr} - \text{clean}) \cdot \nabla_{\text{clean}}$ (Method~A) and $(\text{clean} - \text{corr}) \cdot \nabla_{\text{corr}}$ (Method~B); we report per-layer and per-head scores averaged over the top-$k$ samples activating $f$.

\begin{figure*}[!t]
  \centering
  \includegraphics[width=\linewidth]{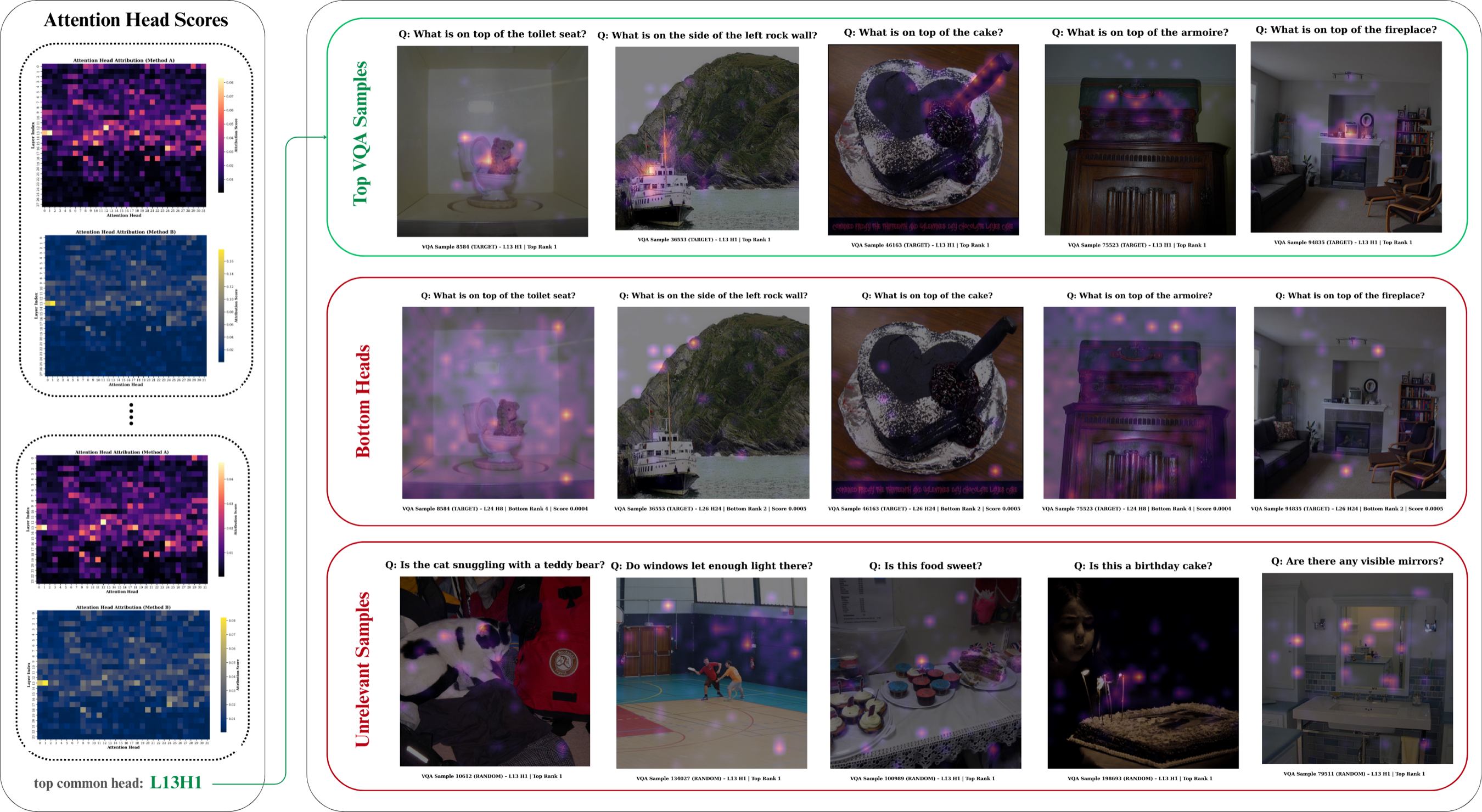}
  \caption{\textbf{Attribution patching on related spatial features.} Top: top head L13H1 localizes to relevant regions for ``on top of''. Middle: bottom-ranked heads fail to localize. Bottom: unrelated queries do not trigger spurious activation.}
  \label{fig:ap-main-appendix}
\end{figure*}

Across the spatial features we examined, layer-wise attribution curves peak in middle layers, consistent with the layer distribution of MMDiff-discovered spatial features (Fig.~\ref{fig:layer-agg}). At the head level, both methods highlight a small subset of heads, and the top heads are largely consistent across variants (Fig.~\ref{fig:head-agg}). Some heads recur across related spatial relations: in the top row of Fig.~\ref{fig:ap-main-appendix}, head L13H1 attends to semantically relevant regions across queries about ``on top of''; the middle row confirms that bottom-ranked heads on the same samples fail to localize, and the bottom row confirms that unrelated queries do not trigger spurious activation. This clustering near $\ell_f$ is consistent with the layer-targeted injection site used by MMDiff~CAA in Sec.~\ref{sec:spatial}.

\subsection{Formalism}
\label{app:attribution-formalism}

This subsection makes the attribution-patching procedure explicit: the scalar objective, the corruption operator, and the per-layer and per-head score formulas, including the absolute-value choice raised in the reviews.

\paragraph{Notation.}
For an input $x$, let $h^{(\ell)}_t(x) \in \mathbb{R}^{d}$ denote the residual-stream output of transformer layer $\ell$ at token position $t$, and let $z^{(\ell)}_t(x) \in \mathbb{R}^{d}$ denote the input to the layer-$\ell$ output projection $W_O^{(\ell)}$, i.e.\ the concatenation of all head outputs at position $t$. We write $z^{(\ell, h)}_t(x) \in \mathbb{R}^{d_h}$ for the slice of $z^{(\ell)}_t(x)$ corresponding to head $h$. Token positions split as $\mathcal{T}(x) = \mathcal{T}_{\text{img}}(x) \sqcup \mathcal{T}_{\text{post}}(x)$, where $\mathcal{T}_{\text{img}}(x)$ is the contiguous span of projected visual tokens and $\mathcal{T}_{\text{post}}(x)$ is the post-image text span at which the model produces its answer.

\paragraph{Scalar objective.}
Given a target SAE feature $f$ with decoder direction $v_f \in \mathbb{R}^{d}$ from the layer-$\ell$ adapted SAE, we define a scalar objective by reading out $v_f$ at a chosen layer $L$ (in our experiments $L = \ell$ unless stated otherwise):
\begin{equation}
\label{eq:attr-objective}
\mathcal{L}(f \mid x) \;=\; \sum_{t \,\in\, \mathcal{T}_{\text{post}}(x)} \big\langle h^{(L)}_t(x),\ v_f \big\rangle .
\end{equation}
Note that $\mathcal{L}$ uses the inner product with the decoder direction, \emph{not} the SAE reconstruction $\hat h^{(L)}_t = \sum_{f'} h_{f'}(x)\, v_{f'}$ nor the gated activation $h_f(x)$. Using the inner product makes $\mathcal{L}$ a linear functional of the residual stream, so gradients flow regardless of whether $f$ is selected by the TopK or JumpReLU sparsity gate, and the linearization in Eqs.~\eqref{eq:method-A-layer}--\eqref{eq:method-B-head} below is well-defined.

\paragraph{Corruption operator.}
Let $\bar{e}_{\text{img}} \in \mathbb{R}^{d}$ be the mean layer-$0$ input over visual-token positions, computed across $N$ reference VQAv2 samples drawn outside the target distribution (we use $N = 256$):
\begin{equation}
\label{eq:mean-emb}
\bar{e}_{\text{img}} \;=\; \frac{1}{\sum_i |\mathcal{T}_{\text{img}}(x_i)|}\, \sum_{i=1}^{N}\, \sum_{t \,\in\, \mathcal{T}_{\text{img}}(x_i)} h^{(0)}_t(x_i).
\end{equation}
The clean run uses the original $x$; the corrupt run replaces layer-$0$ inputs at every visual-token position with $\bar{e}_{\text{img}}$, leaving all other positions and all later layers' computations unchanged:
\begin{equation}
\label{eq:corrupt}
h^{(0)}_t(x^{\text{corr}}) \;=\;
\begin{cases}
\bar{e}_{\text{img}} & \text{if } t \in \mathcal{T}_{\text{img}}(x), \\
h^{(0)}_t(x) & \text{otherwise.}
\end{cases}
\end{equation}
This preserves the layer-$0$ distributional statistics of visual-token positions while suppressing the image-specific signal that drives the target feature, so $h^{(\ell)}_t(x^{\text{clean}}) - h^{(\ell)}_t(x^{\text{corr}})$ at any later layer $\ell$ is the residual-stream change attributable to the visual content of $x$.

\paragraph{Per-layer attribution.}
For each layer $\ell < L$ we evaluate the first-order approximation of the change in $\mathcal{L}(f \mid x)$ when $h^{(\ell)}_t$ is patched, using the gradient of $\mathcal{L}$ at one of the two endpoints. Method~A linearizes around the clean run and Method~B linearizes around the corrupt run:
\begin{align}
s^{(A)}_{\ell}(f \mid x) \;&=\; \frac{1}{|\mathcal{T}_{\text{post}}(x)|} \sum_{t \in \mathcal{T}_{\text{post}}(x)} \!\Big| \big\langle h^{(\ell)}_t(x^{\text{corr}}) - h^{(\ell)}_t(x^{\text{clean}}),\ \nabla_{h^{(\ell)}_t} \mathcal{L}(f \mid x^{\text{clean}}) \big\rangle \!\Big|, \label{eq:method-A-layer} \\
s^{(B)}_{\ell}(f \mid x) \;&=\; \frac{1}{|\mathcal{T}_{\text{post}}(x)|} \sum_{t \in \mathcal{T}_{\text{post}}(x)} \!\Big| \big\langle h^{(\ell)}_t(x^{\text{clean}}) - h^{(\ell)}_t(x^{\text{corr}}),\ \nabla_{h^{(\ell)}_t} \mathcal{L}(f \mid x^{\text{corr}}) \big\rangle \!\Big|. \label{eq:method-B-layer}
\end{align}
The per-token absolute value is taken \emph{before} averaging, so $s^{(A)}_{\ell}$ and $s^{(B)}_{\ell}$ measure attribution magnitude rather than signed effect: a high score means the layer's residual contribution is causally aligned with $f$ in either direction.

\paragraph{Per-head attribution.}
The per-head version replaces the residual-stream activation $h^{(\ell)}_t$ in Eqs.~\eqref{eq:method-A-layer}--\eqref{eq:method-B-layer} with the layer-$\ell$ output-projection input restricted to head $h$, $z^{(\ell, h)}_t$:
\begin{align}
s^{(A)}_{\ell, h}(f \mid x) \;&=\; \frac{1}{|\mathcal{T}_{\text{post}}(x)|} \sum_{t \in \mathcal{T}_{\text{post}}(x)} \!\Big| \big\langle z^{(\ell, h)}_t(x^{\text{corr}}) - z^{(\ell, h)}_t(x^{\text{clean}}),\ \nabla_{z^{(\ell, h)}_t} \mathcal{L}(f \mid x^{\text{clean}}) \big\rangle \!\Big|, \label{eq:method-A-head} \\
s^{(B)}_{\ell, h}(f \mid x) \;&=\; \frac{1}{|\mathcal{T}_{\text{post}}(x)|} \sum_{t \in \mathcal{T}_{\text{post}}(x)} \!\Big| \big\langle z^{(\ell, h)}_t(x^{\text{clean}}) - z^{(\ell, h)}_t(x^{\text{corr}}),\ \nabla_{z^{(\ell, h)}_t} \mathcal{L}(f \mid x^{\text{corr}}) \big\rangle \!\Big|. \label{eq:method-B-head}
\end{align}
We attribute to $z^{(\ell, h)}$ rather than to the post-output-projection contribution because this isolates the per-head signal before mixing through $W_O^{(\ell)}$.

\paragraph{Aggregation across samples.}
Both per-layer and per-head scores are averaged over the top-$k$ samples that most strongly activate $f$ on the target distribution (we use $k = 100$ unless stated otherwise). With $\mathcal{X}_f^{\text{top}}$ the top-$k$ activating set and $\bullet \in \{A, B\}$,
\begin{equation}
\label{eq:attr-agg}
S^{(\bullet)}_{\ell}(f) = \frac{1}{|\mathcal{X}_f^{\text{top}}|} \sum_{x \in \mathcal{X}_f^{\text{top}}} s^{(\bullet)}_{\ell}(f \mid x), \qquad
S^{(\bullet)}_{\ell, h}(f) = \frac{1}{|\mathcal{X}_f^{\text{top}}|} \sum_{x \in \mathcal{X}_f^{\text{top}}} s^{(\bullet)}_{\ell, h}(f \mid x).
\end{equation}
Top-driving heads are obtained by ranking $S^{(\bullet)}_{\ell, h}(f)$ in decreasing order; because the per-token absolute value is already taken inside the per-sample score (Eqs.~\eqref{eq:method-A-head}--\eqref{eq:method-B-head}), this ranking is by attribution magnitude regardless of the sign of the underlying contrast.

\begin{figure*}[t]
    \centering
    \includegraphics[width=0.42\textwidth]{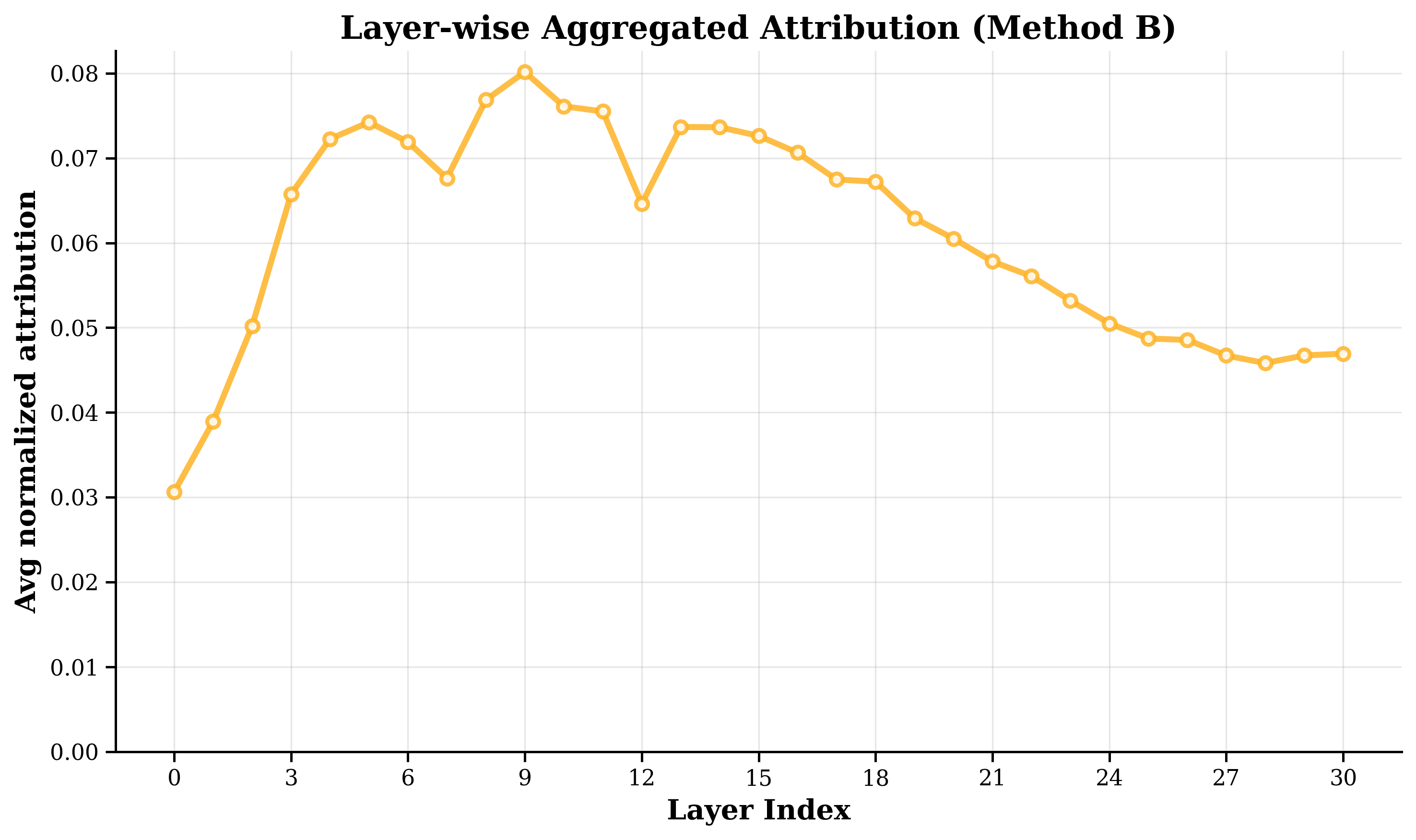}
    \hspace{1em}\includegraphics[width=0.42\textwidth]{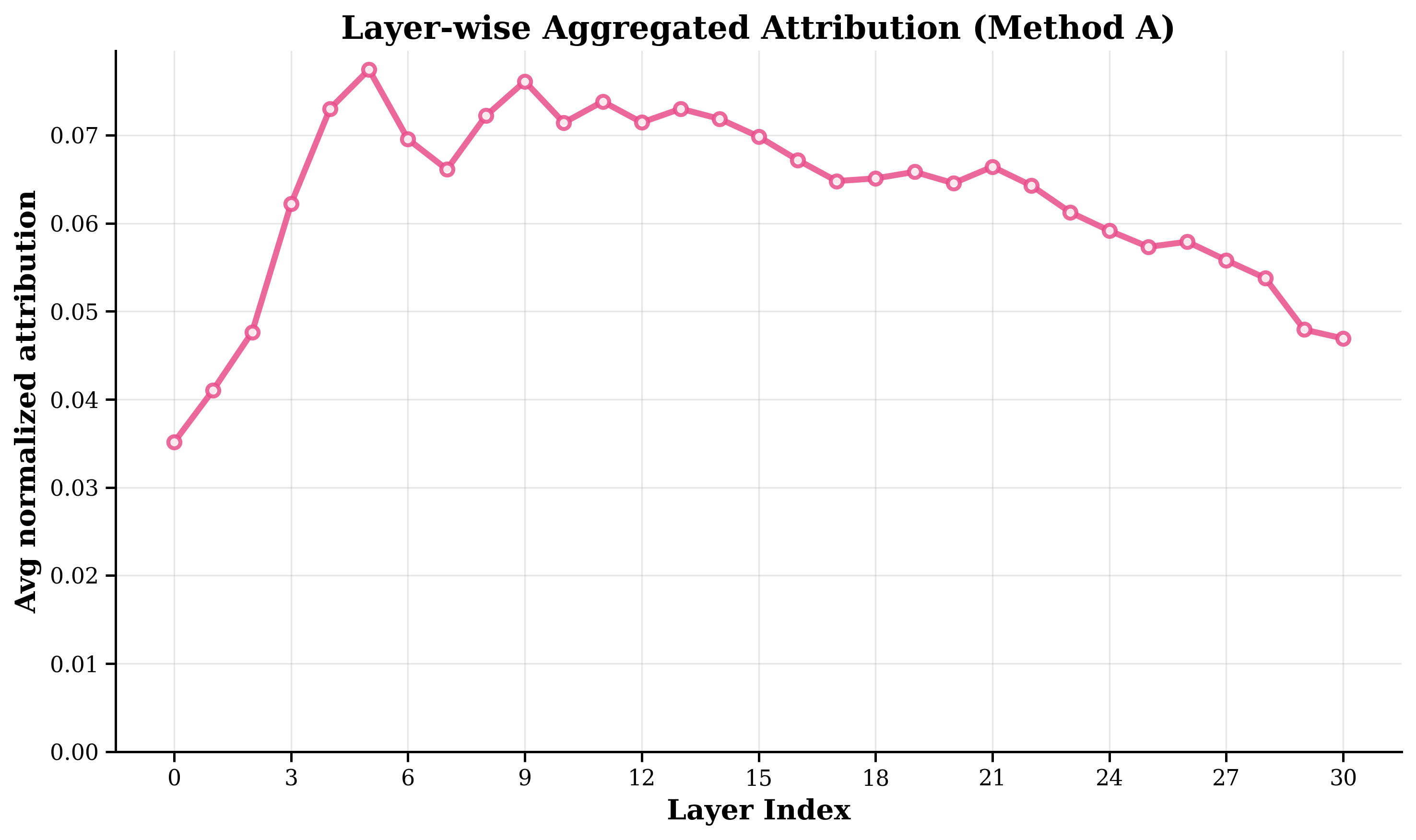}
    \caption{Layer-wise aggregated attribution curves for Method B (left) and Method A (right). Both peak around middle layers, consistent with the emergence of spatial features.}
    \label{fig:layer-agg}
\end{figure*}

\begin{figure*}[t]
    \centering
    \includegraphics[width=0.5\textwidth]{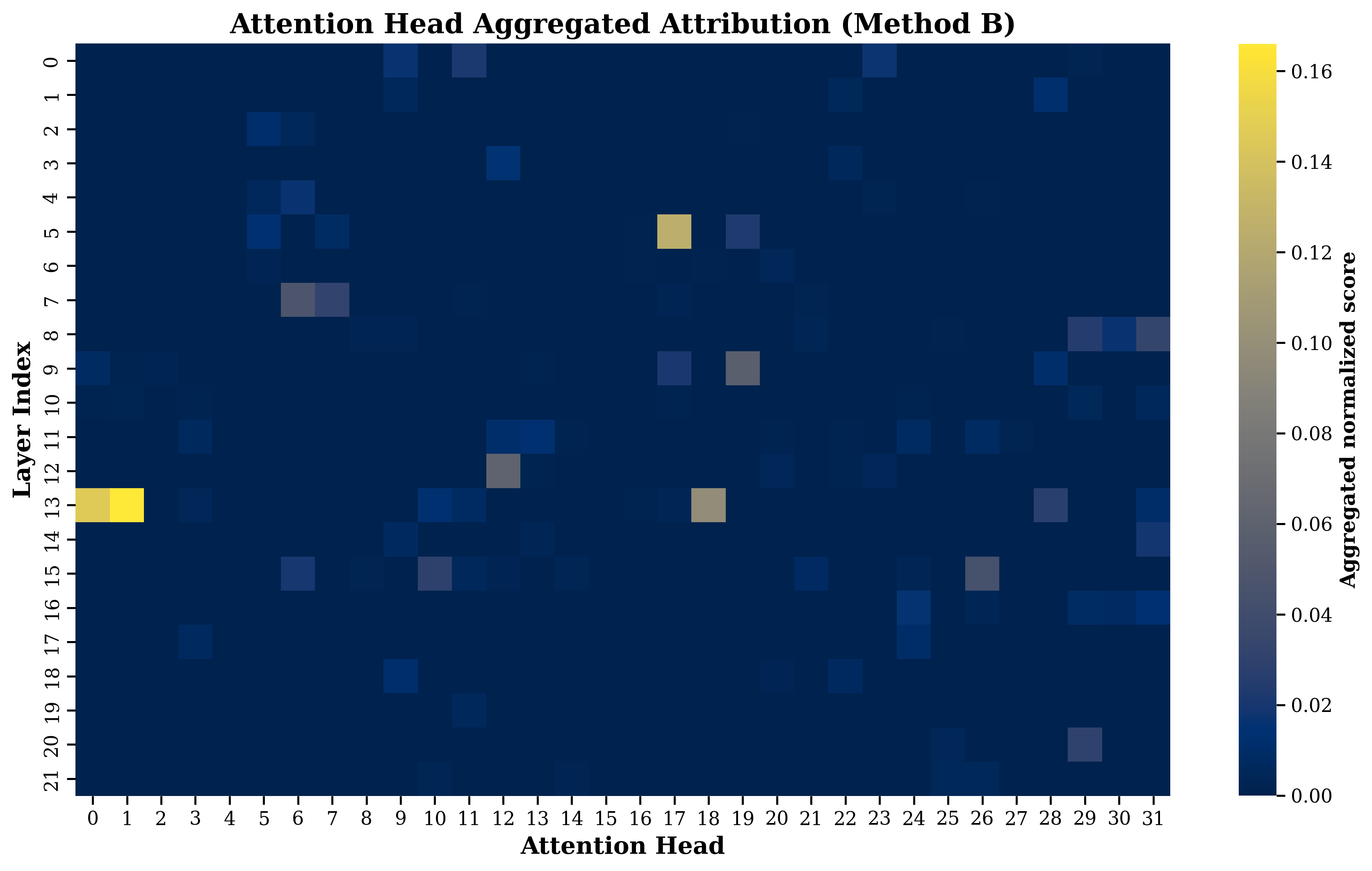}
    \includegraphics[width=0.45\textwidth]{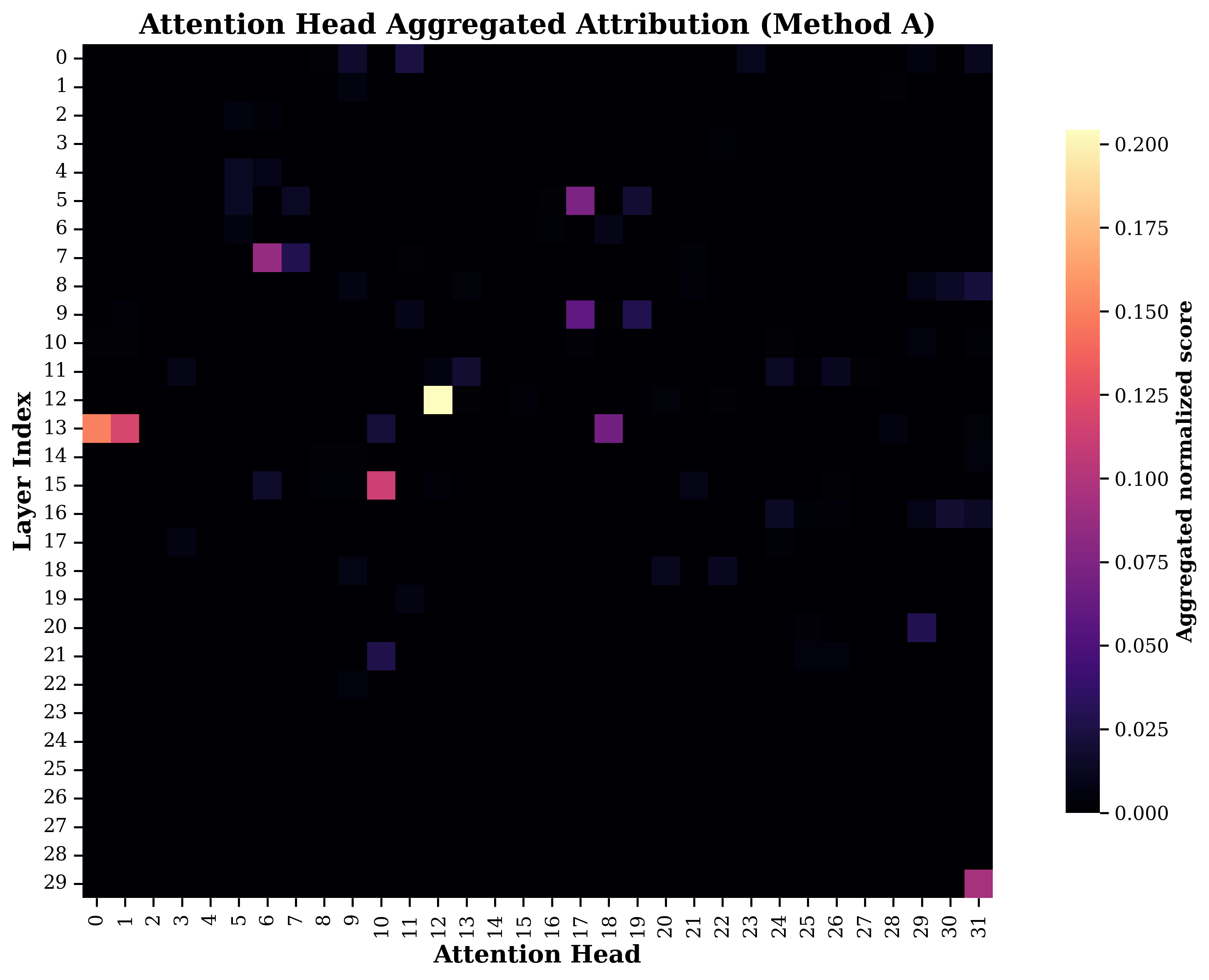}
    \caption{Attention head aggregated attribution maps for Method B (left) and Method A (right). Both highlight a similar set of specialized heads with high attribution scores.}
    \label{fig:head-agg}
\end{figure*}

\begin{figure*}[t]
  \centering
  \begin{subfigure}[t]{0.98\textwidth}
    \centering
    \includegraphics[width=\linewidth]{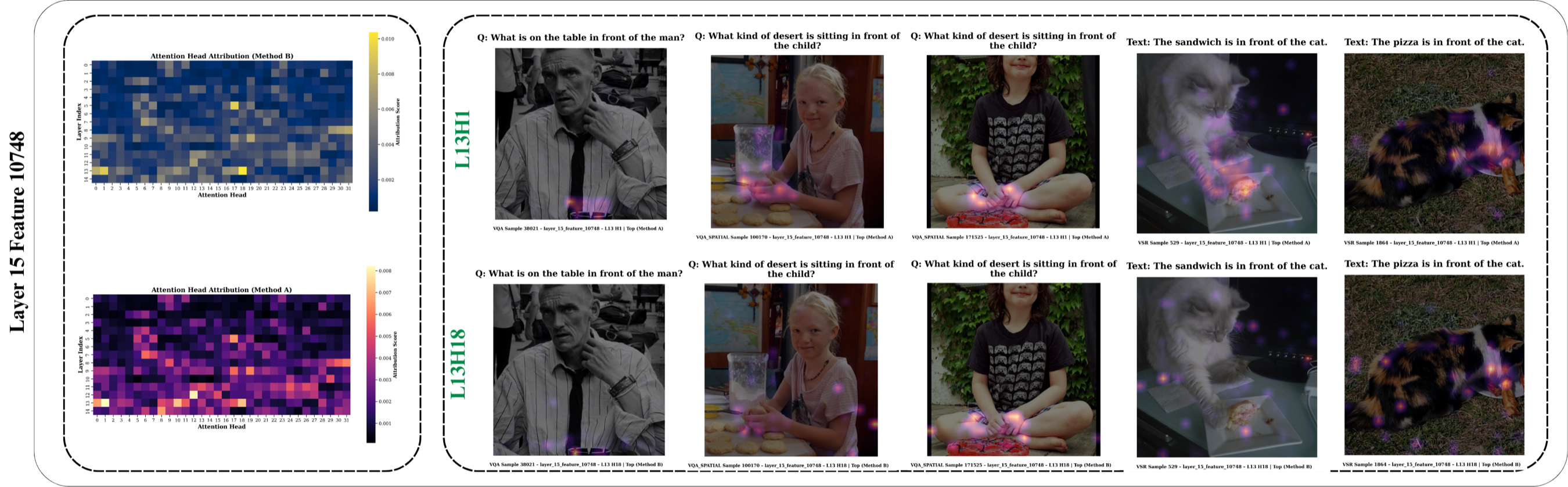}
    \caption{\textbf{Layer 15, Feature 10748.} VSR Relation: ``in front of.''\;
    Top heads (Method A): \texttt{L13H1}, \texttt{L12H12}, \texttt{L13H18}.\;
    Top heads (Method B): \texttt{L13H18}, \texttt{L5H17}, \texttt{L13H1}.\;
    \emph{Overlap}: \texttt{L13H1}, \texttt{L13H18}.
    Attention overlays are shown on the top-activating samples across VSR and VQA.}
    \label{fig:ap-single-l15f10748}
  \end{subfigure}

  \begin{subfigure}[t]{0.98\textwidth}
    \centering
    \includegraphics[width=\linewidth]{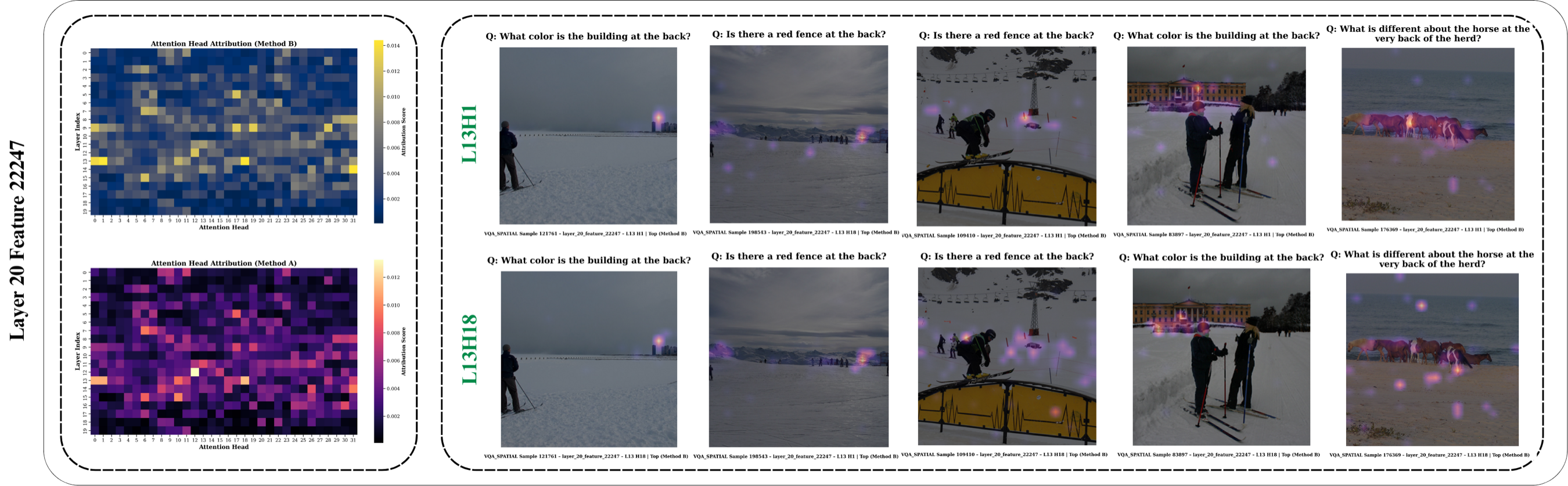}
    \caption{\textbf{Layer 20, Feature 22247.} VSR Relation: ``at the back of.''\;
    Top heads (Method A): \texttt{L12H12}, \texttt{L13H18}, \texttt{L13H1}.\;
    Top heads (Method B): \texttt{L13H1}, \texttt{L13H18}, \texttt{L14H31}.\;
    \emph{Overlap}: \texttt{L13H1}, \texttt{L13H18}.
    Attention overlays are shown on the top-activating samples across VSR and VQA.}
    \label{fig:ap-single-l20f22247}
  \end{subfigure}

  \begin{subfigure}[t]{0.98\textwidth}
    \centering
    \includegraphics[width=\linewidth]{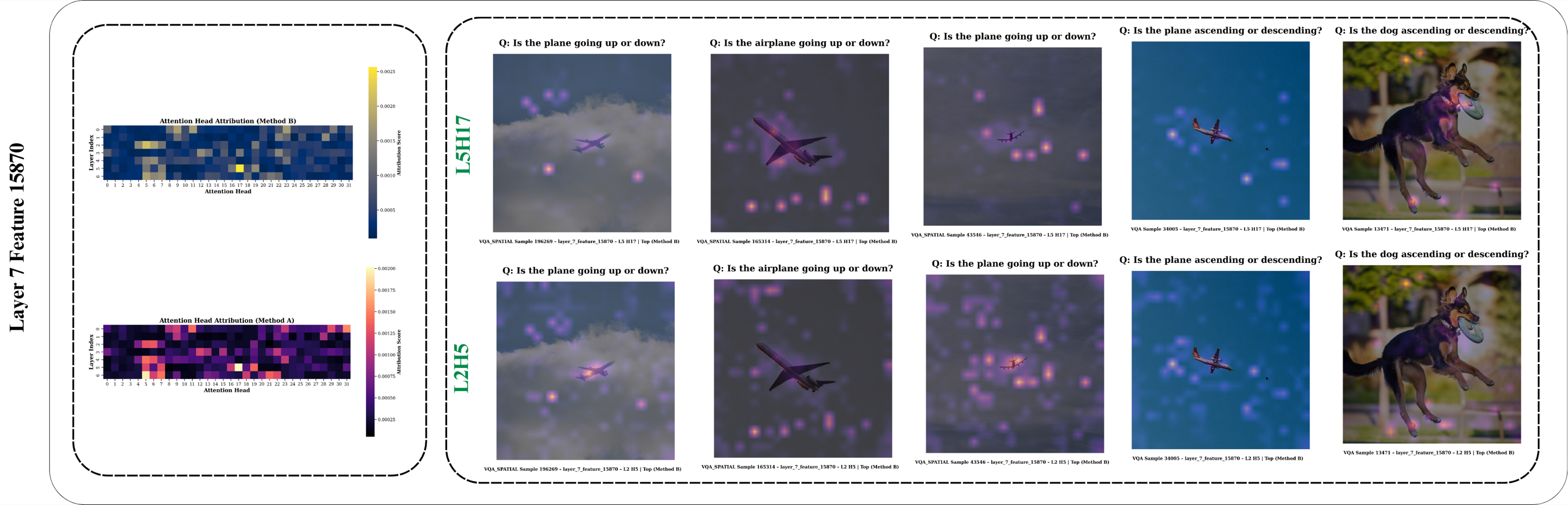}
    \caption{\textbf{Layer 7, Feature 15870.} VSR Relation: ``above.''\;
    Top heads (Method A): \texttt{L5H17}, \texttt{L6H5}, \texttt{L0H31}.\;
    Top heads (Method B): \texttt{L5H17}, \texttt{L2H5}, \texttt{L2H6}.\;
    \emph{Overlap}: \texttt{L5H17}.
    Attention overlays are shown on the top-activating samples across VSR and VQA.}
    \label{fig:ap-single-l7f15870}
  \end{subfigure}

  \caption{\textbf{Attribution patching on individual spatial features.}
  Each subfigure displays aggregated head/layer attribution maps (left) and attention overlays (right) using the strongest heads on the feature's top-activating samples across both VSR and VQA.}
  \label{fig:ap-single-all}
\end{figure*}

\begin{figure*}[t]
    \centering
    \includegraphics[width=\linewidth]{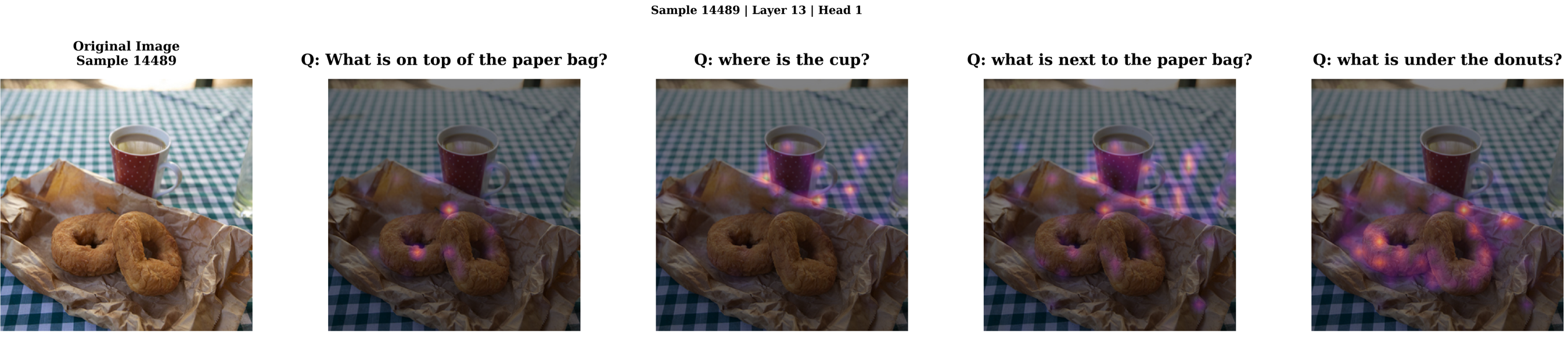}
    \vspace{0.5em}
    \includegraphics[width=\linewidth]{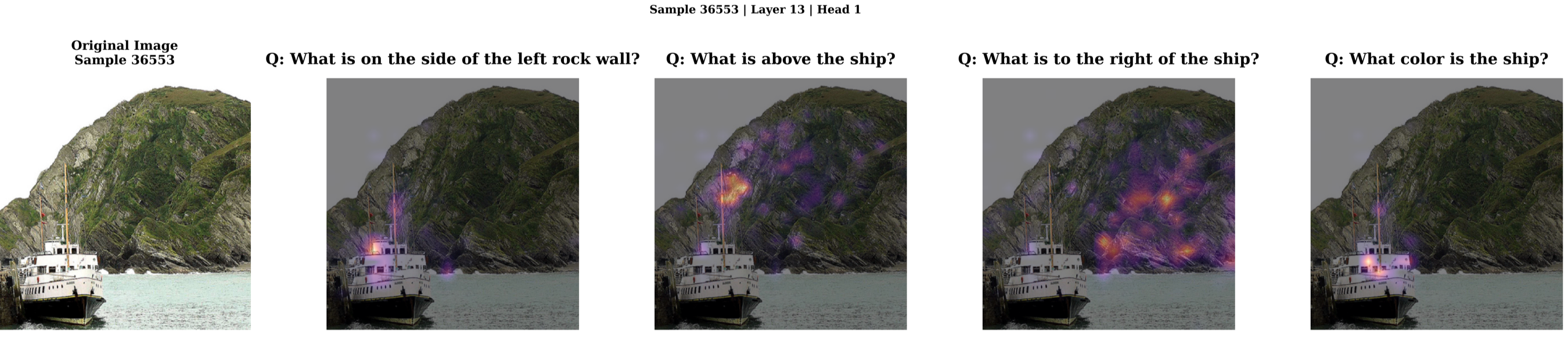}
    \vspace{0.5em}
    \includegraphics[width=\linewidth]{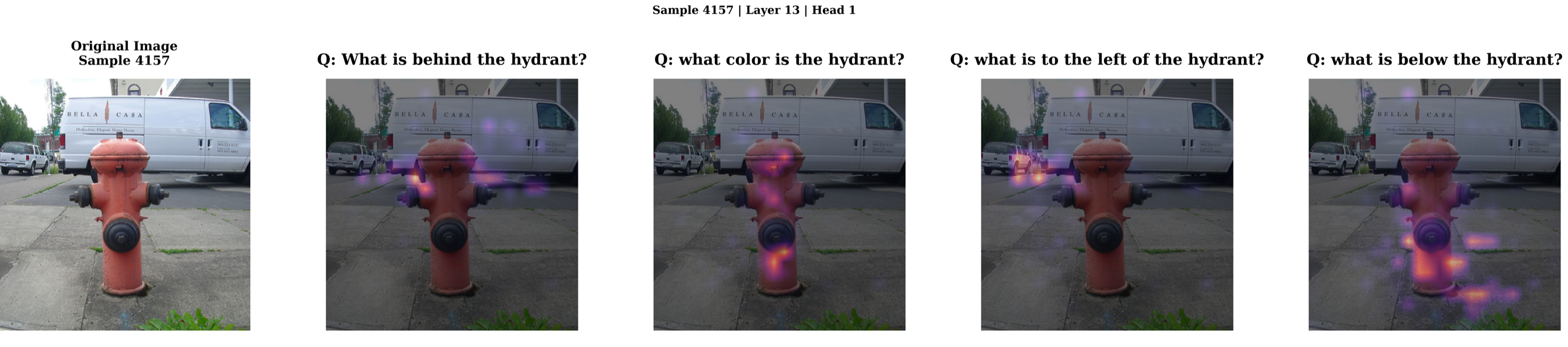}
  \caption{\textbf{Attention head visualizations across queries.}
Each row shows one image with attention overlays from a single high-attribution head across multiple spatial and non-spatial custom queries. The same heads consistently focus on semantically relevant regions.}
  \label{fig:semantic-heads}
\end{figure*}

Figure~\ref{fig:layer-agg} aggregates per-layer attribution scores across the spatial features in Sec.~\ref{sec:spatial}, showing the mid-layer concentration referenced in the main text. Figure~\ref{fig:head-agg} aggregates per-head scores. Figure~\ref{fig:ap-single-all} provides per-feature panels showing the top-scoring heads and their attention maps for individual features. Figure~\ref{fig:semantic-heads} examines how the recurring top heads behave under custom semantic prompts.

\subsection{Bottom-Ranked Heads as a Control}
\label{app:bottom-heads}

As a control, we visualize overlays from the \emph{bottom-ranked} heads (per method, per feature). Across VSR and VQA top-activating samples, these heads generally fail to localize semantically relevant regions, in contrast to the top-ranked heads in Fig.~\ref{fig:ap-single-all}.

\begin{figure*}[!t]
  \centering
  \begin{subfigure}[t]{0.98\textwidth}
    \centering
    \includegraphics[width=\linewidth]{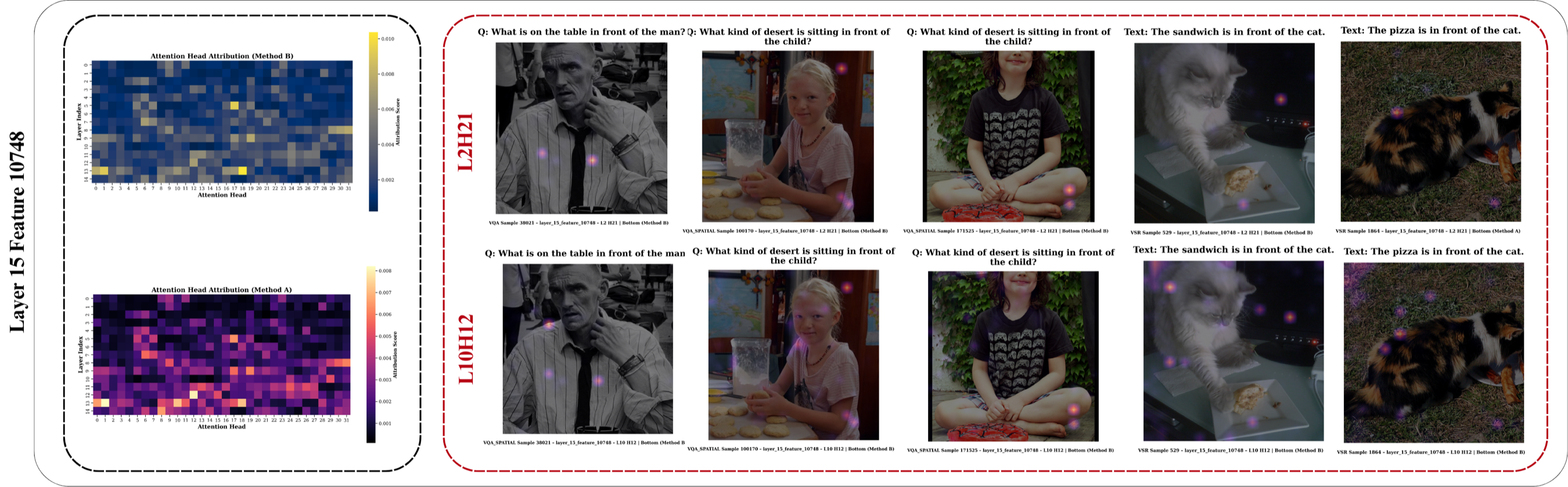}
    \caption{\textbf{Layer 15, Feature 10748.} VSR Relation: ``in front of.''}
    \label{fig:ap-neg-l15f10748}
  \end{subfigure}

  \vspace{1em}

  \begin{subfigure}[t]{0.98\textwidth}
    \centering
    \includegraphics[width=\linewidth]{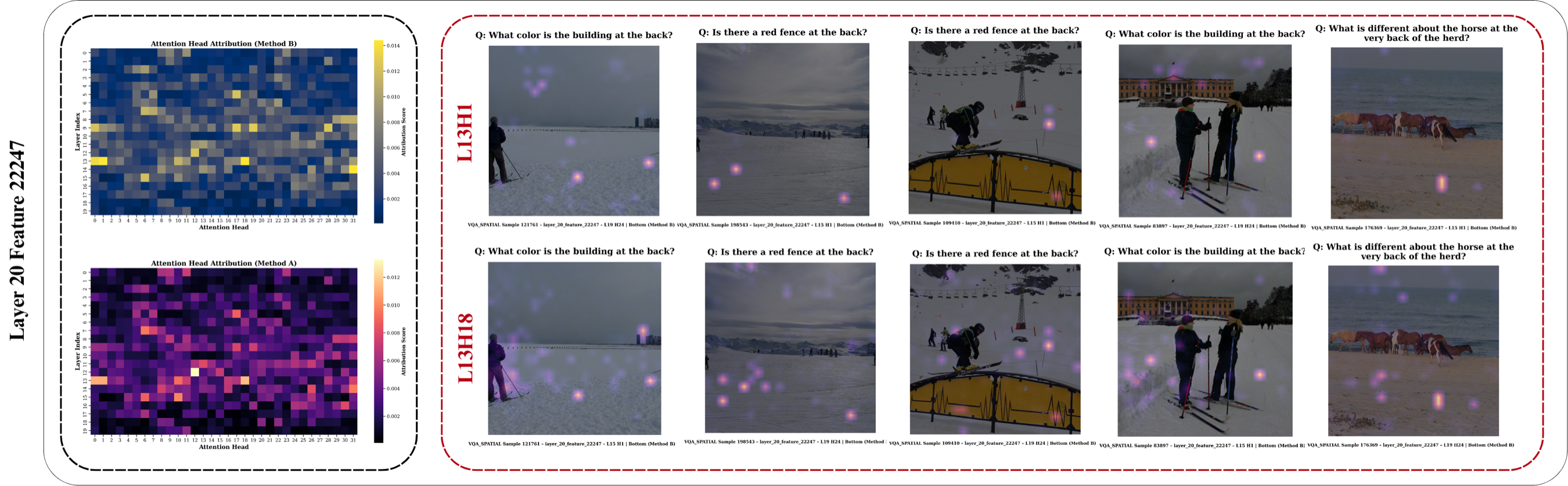}
    \caption{\textbf{Layer 20, Feature 22247.} VSR Relation: ``at the back of.''}
    \label{fig:ap-neg-l20f22247}
  \end{subfigure}

  \vspace{1em}

  \begin{subfigure}[t]{0.98\textwidth}
    \centering
    \includegraphics[width=\linewidth]{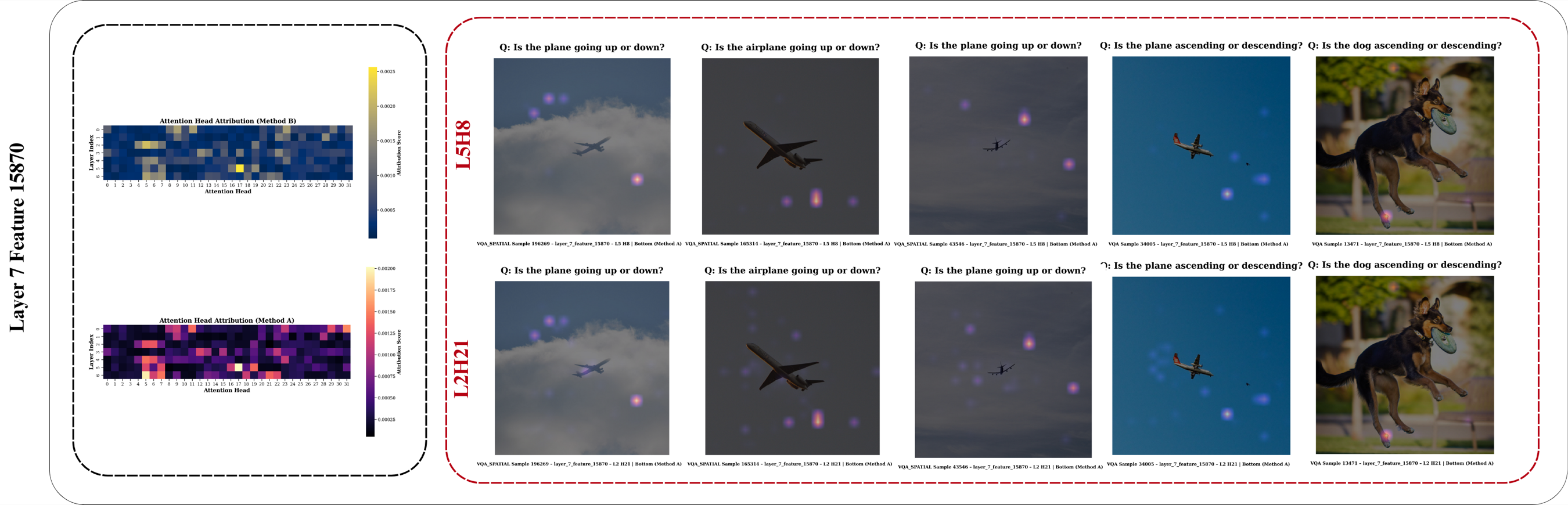}
    \caption{\textbf{Layer 7, Feature 15870.} VSR Relation: ``above.''}
    \label{fig:ap-neg-l7f15870}
  \end{subfigure}

  \caption{\textbf{Bottom-ranked heads yield weak localization.} For each feature, we show overlays from the lowest-scoring heads under Methods A and B on the feature's top-activating samples across VSR and VQA. In contrast to Fig.~\ref{fig:ap-single-all}, these heads produce diffuse or irrelevant attention.}
  \label{fig:ap-single-neg-all}
\end{figure*}

\section{OCR Feature Examples}
\label{app:ocr}

\begin{figure*}[!t]
  \centering
  \includegraphics[width=0.9\linewidth]{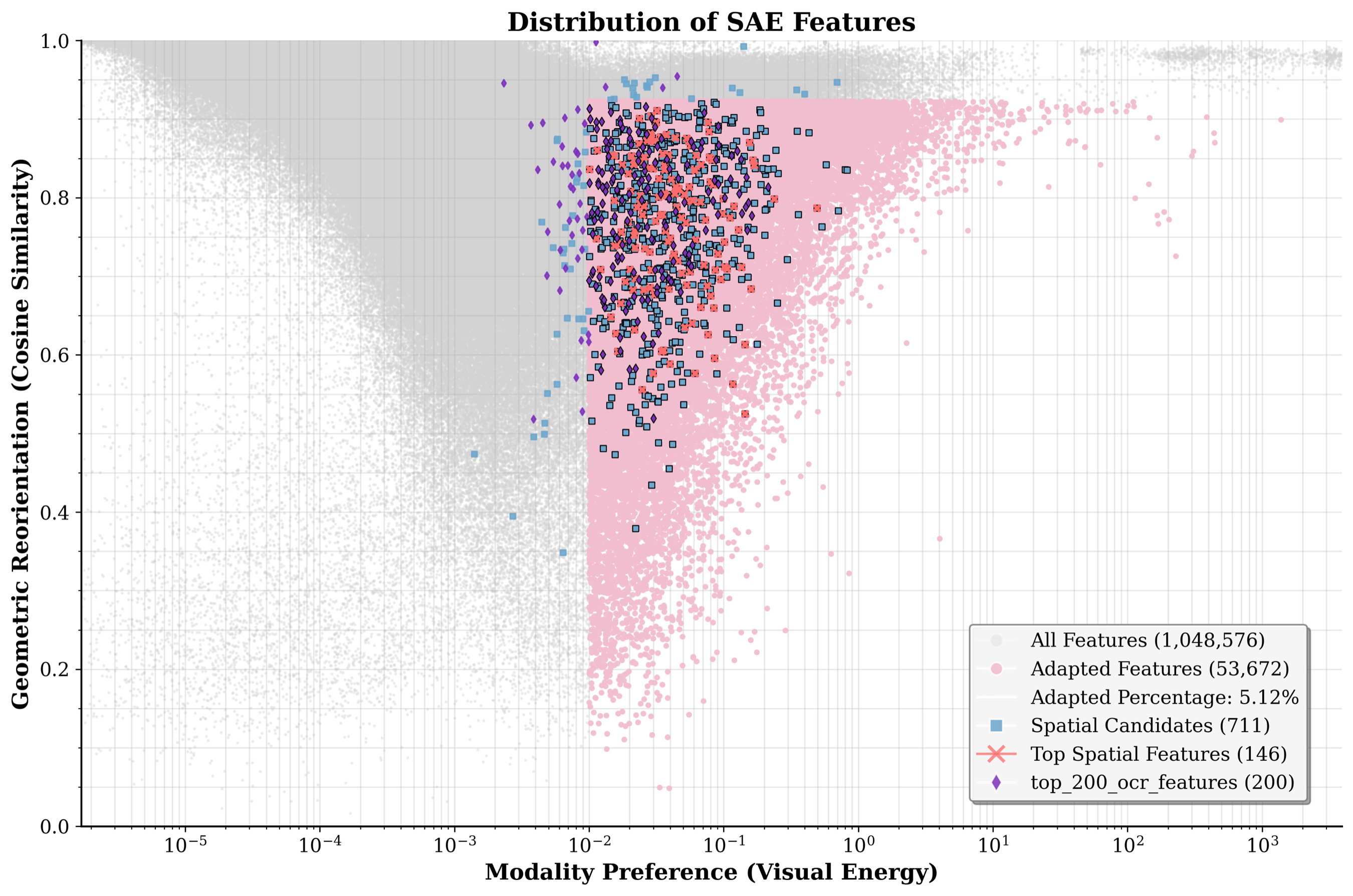}
  \caption{\textbf{Distribution of OCR features.}
  Top OCR candidates (purple) cluster among adapted units (pink), paralleling the
  spatial subset (blue).}
  \label{fig:ocr-scatter}
\end{figure*}

We also apply our distribution-shift procedure to OCR-style prompts (e.g.,
``What does the sign say?''). Fig.~\ref{fig:ocr-scatter} shows that OCR-selective
features cluster within the same adapted region as the spatial subset,
indicating that multimodal fine-tuning concentrates visually grounded capabilities
into a compact envelope of feature space.

\begin{figure*}[!t]
  \centering
  \includegraphics[width=0.95\textwidth]{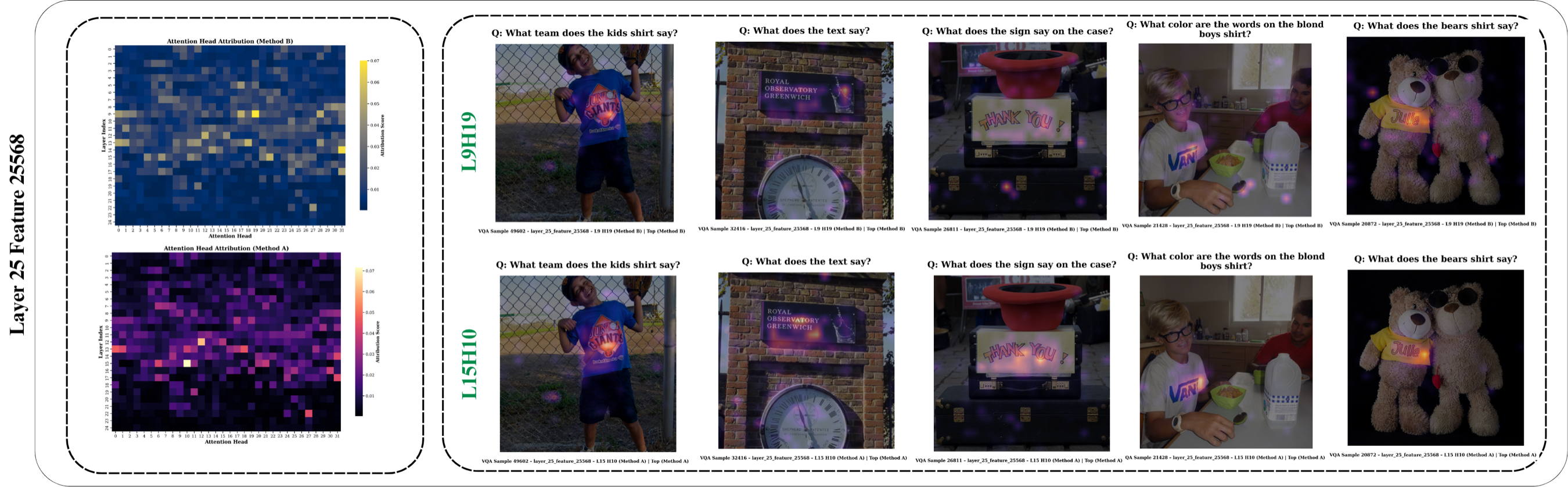}
  \caption{\textbf{Layer 25, Feature 25568.}
  Activates on storefront and clothing text; top heads align to characters.}
  \label{fig:ocr-extra}
\end{figure*}

Qualitative examples confirm that these features reliably activate on embedded
text and that associated heads localise to glyph regions (Fig.
\ref{fig:ocr-extra}), consistent with image-grounded text processing.

Qualitative panels show top-activating samples and overlay maps for OCR-selective features identified in Sec.~\ref{sec:ocr}. The main-paper ablation and steering tables for OCR are Tables~\ref{tab:abl-ocr} and~\ref{tab:steer-ocr}; this appendix is reserved for per-feature qualitative panels.

\clearpage


\end{document}